\pdfoutput=1 
\documentclass[
12pt, 
oneside, 
english, 
singlespacing, 
headsepline, 
]{MastersDoctoralThesis} 

\usepackage[utf8]{inputenc} 
\usepackage[T1]{fontenc} 
\usepackage[utf8]{inputenc}
\usepackage{subfig}
\usepackage{caption}
\usepackage{amsmath}
\usepackage{amssymb}
\usepackage{mathrsfs}
\usepackage{graphicx}
\usepackage{caption}
\usepackage{subfig}
\usepackage{ntheorem}
\usepackage{algorithm}
\usepackage[noend]{algpseudocode}
\usepackage[sfdefault=cmbr,OMLmathsans]{isomath}  
\usepackage{upgreek}  
\usepackage{tensor}  
\usepackage{mathtools}
\usepackage{tikz}

\newcommand{\ignore}[1]{}

\newcommand{\vek}[1]{\mathchoice{\displaystyle\boldsymbol{#1}}
{\textstyle\boldsymbol{#1}}{\scriptstyle\boldsymbol{#1}}
{\scriptscriptstyle\boldsymbol{#1}}}

\def\realnr{\mathbb{R}}

\usepackage{mathpazo} 

\usepackage[style=alphabetic]{biblatex} 

\usepackage[autostyle=true]{csquotes} 

\thesistitle{Classification based on invisible features and thereby finding the effect of tuberculosis vaccine on COVID-19} 
\supervisor{Prof. Hermann \textsc{Matthies}} 
\examiner{} 
\degree{\href{https://www.tu-braunschweig.de/cse}{Computational Science in Engineering}} 
\author{Nihal \textsc{Acharya Adde}} 
\addresses{} 

\subject{Computational Science in Engineering} 
\university{\href{https://www.tu-braunschweig.de}{TU Braunschweig}} 
\department{\href{https://www.tu-braunschweig.de/wire}{Institute for Scientific Computing \\ Technische Universität Braunschweig \\ Mühlenpfordtstrasse 23, 38106 Braunschweig}} 
\group{\href{http://researchgroup.university.com}{Research Group Name}} 
\faculty{\href{http://faculty.university.com}{Faculty Name}} 

\AtBeginDocument{
\hypersetup{pdftitle=\ttitle} 
\hypersetup{pdfauthor=\authorname} 
\hypersetup{pdfkeywords=\keywordnames} 
}

\begin{document}

\frontmatter 

\pagestyle{plain} 


\begin{titlepage}

\setlength{\unitlength}{1mm}
  \begin{picture}(00,00)(+14,+7)
	\put(53,5.5){\color[cmyk]{0.1,1,0.8,0}\linethickness{0.5mm}\line(130,0){130}}
	\put(0,0){\includegraphics[scale=0.9]{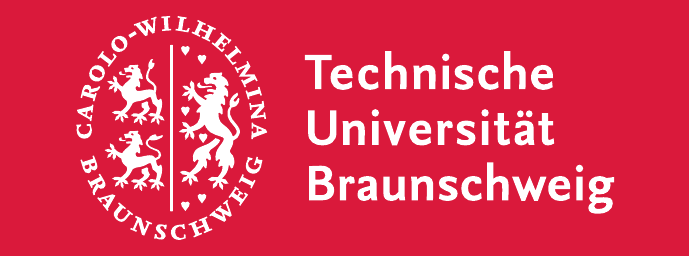}}
	\put(140,6){\includegraphics[width=4cm, height=2cm]{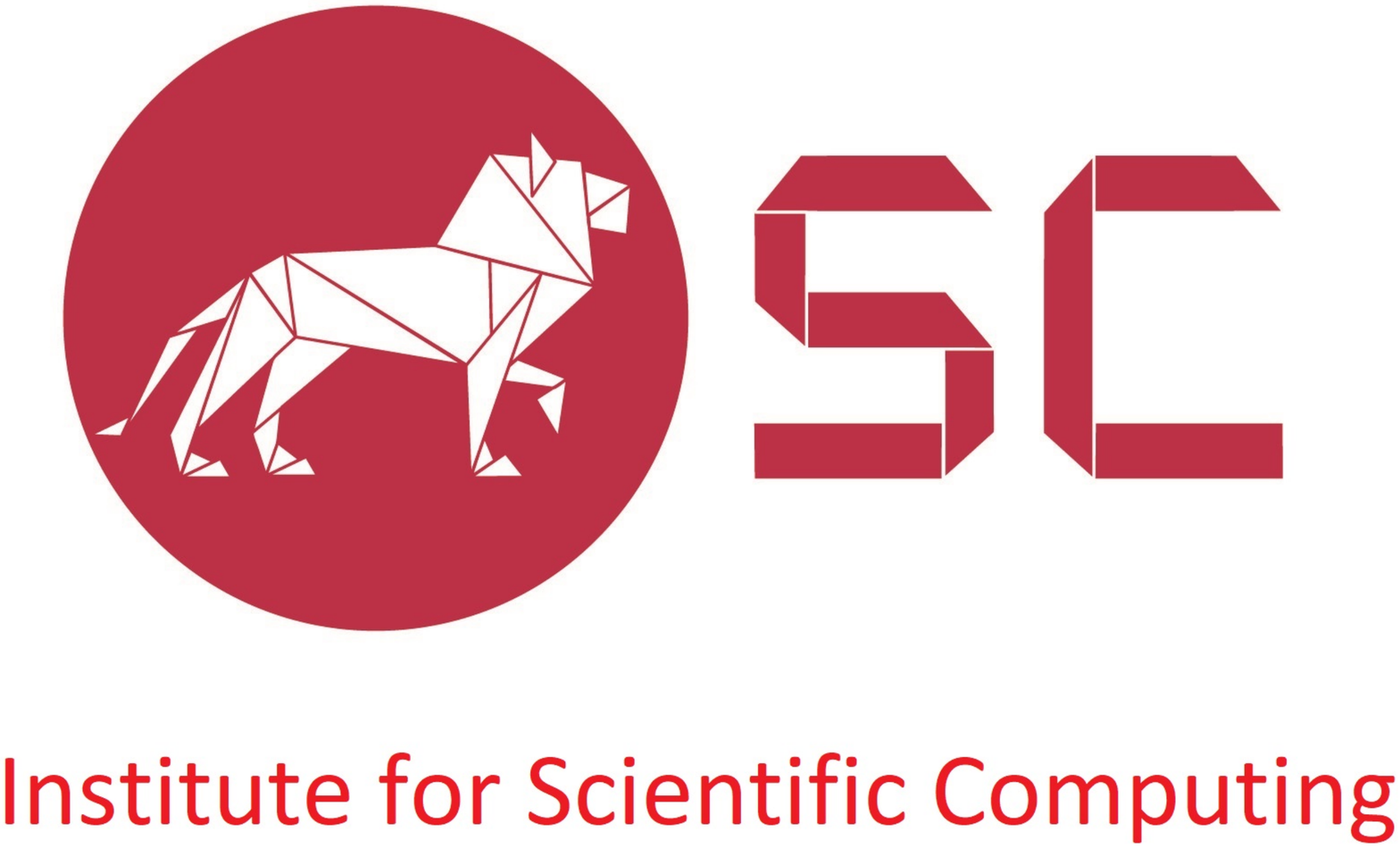}}
  \end{picture}

\begin{center}

\vspace*{.06\textheight}
{\scshape\LARGE \univname\par}\vspace{1.5cm} 
\textsc{\Large Studienarbeit}\\[0.5cm] 

\HRule \\[0.4cm] 
{\huge \bfseries \ttitle\par}\vspace{0.4cm} 
\HRule \\[1.5cm] 
 
\begin{minipage}[t]{0.45\textwidth}
\begin{flushleft} \large
\emph{Author:}\\
\href{mailto:n.acharya-adde@tu-braunschweig.de}{\authorname} 
\emph{Matrikelnummer:}\\
\href{mailto:n.acharya-adde@tu-braunschweig.de}{4941635}
\end{flushleft}
\end{minipage}
\begin{minipage}[t]{0.45\textwidth}
\begin{flushright} \large
\emph{Examiner:} \\
{\href{mailto:h.matthies@tu-braunschweig.de}{\supname}}
 \\
\emph{Supervisor:} \\
\href{mailto:thilo.moshagen@hs-wismar.de}{Dr. Thilo \textsc{Moshagen}}
\end{flushright}
\end{minipage}\\[1cm]

\vfill 

\large \textit{A student project submitted in fulfillment of the requirements\\ for the degree of \degreename}\\[0.3cm] 
\textit{at }\\[0.4cm]
\deptname\\[1cm] 
 
\vfill

{\large October 12, 2020} 
\end{center}
\end{titlepage}

\begin{declaration}
\vspace{0.5cm}
\addchaptertocentry{\authorshipname} 
\noindent Ich versichere, dass ich die Arbeit ohne fremde Hilfe und ohne Benutzung anderer als der
angegebenen Quellen angefertigt habe, und dass die Arbeit in gleicher oder ähnlicher Form noch
keiner anderen Prüfungsbehörde vorgelegen hat und von dieser als Teil einer Prüfungsleistung
angenommen wurde. Alle Ausführungen, die wörtlich oder sinngemäß übernommen wurden,
sind als solche gekennzeichnet.

\vspace{1cm}

\noindent Braunschweig, den

 \vspace{2cm}
 
\noindent Nihal Acharya Adde

\end{declaration}

\cleardoublepage

\begin{abstract}
\addchaptertocentry{\abstractname} 
In the case of clustered data, an artificial neural network with logcosh loss function learns the bigger cluster rather than the mean of the two. Even more so, the ANN when used for regression of a set-valued function, will learn a value close to one of the choices, in other words, it learns one branch of the set-valued function with high accuracy.  This work suggests a method that uses artificial neural networks with logcosh loss to find the branches of set-valued mappings in parameter-outcome sample sets and classifies the samples according to those branches. The method not only classifies the data based on these branches but also provides an accurate prediction for the majority cluster. The method successfully classifies the data based on an invisible feature.  A neural network was successfully established to predict the total number of cases, the logarithmic total number of cases, deaths, active cases and other relevant data of the coronavirus for each German district from a number of input variables. As it has been speculated that the Tuberculosis vaccine provides protection against the virus and since East Germany was vaccinated before reunification, an attempt was made to classify the Eastern and Western German districts by considering the vaccine information as an invisible feature.

\vspace{1cm}
\end{abstract}




\tableofcontents 

\listoffigures 



\begin{abbreviations}{ll} 

\textbf{ANN} & \textbf{A}rtifical \textbf{N}eural \textbf{N}etwork\\
\textbf{CNN} & \textbf{C}onnvolutional \textbf{N}eural \textbf{N}etwork\\
\textbf{DNN} & \textbf{D}eep \textbf{N}eural \textbf{N}etwork\\
\textbf{ELU} & \textbf{E}xponential \textbf{L}inear \textbf{U}nit\\
\textbf{ReLU} & \textbf{Re}ctified \textbf{L}inear \textbf{U}nit\\
\textbf{SGD} & \textbf{S}tochastic \textbf{G}radient \textbf{D}escent\\
\textbf{COVID-19} & \textbf{CO}rona \textbf{VI}rus \textbf{D}isease - 19\\
\textbf{TB} & \textbf{T}uberculosis\\
\textbf{BCG} & \textbf{B}acillus \textbf{C}almette \textbf{G}uérin\\
\textbf{MSE} & \textbf{M}ean \textbf{S}quare \textbf{E}rror\\
\textbf{MAE} & \textbf{M}ean \textbf{A}bsolute \textbf{E}rror\\

\end{abbreviations}












\mainmatter 

\pagestyle{thesis} 



\chapter{Introduction} 

\label{Chapter1} 

Acquiring sensible data for building an efficient model is always a challenging task. A simple question that arises is how significant are the chosen inputs for predictive calculations. Having more data to train a network is a blessing but the most important factor is to have a relevant input feature set that represents the model precisely. Therefore, selection of the right parameters to represent the model becomes one of the most important tasks in deep learning.
\newline
\par
Deep learning approaches have been increasingly used in the field of computer vision, robotics, translations, speech recognition, autonomous vehicles, etc. Deep learning represents the evolution of machine learning. It learns through artificial neural networks that act similar to the human brain and allows the machine to analyze data in a structure very much as humans do. Chapter \ref{Chapter2} discusses the theory behind the deep learning approach followed to represent our model. It discusses in detail the network framework, training of the feed-forward network, behaviour of the loss functions and the importance of the selection of good input features to train the model.
\newline
\par
Many deep learning models are trained with the available data which sometimes doesn't give good results despite using the best-suited algorithm for the model. For most real-world problems, there is a possibility that an input parameter is not considered for training the network because it is potentially unknown which becomes one of the most important features to correctly represent the model. With this work, we aim to check the behaviour of the network when an important feature is ignored in setting up the model. Before getting into the real world example, the research aims to demonstrate the behaviour of the network on a simple 1-dimensional and 2-dimensional toy problems. Chapter \ref{Chapter3} attempts to classify a multi-valued data set. A simple fully connected neural network with logcosh loss function is used to train the model and then tested with a fraction of our data set to check the behaviour of the network. When predicted on our multi-valued data set, the network tends to learn the bigger cluster of data accurately than the mean of the two. This theory is then used to validate if the tuberculosis vaccine protects against the fatal coronavirus by considering the vaccine information as an invisible feature.
\newline
\par
As we know by a fact that Eastern Germany was compulsorily vaccinated for tuberculosis before German unification, it gives clear classification criteria for Eastern and Western Germany based on the vaccine information. However, the network only classifies the data if the vaccine has indeed provided immunity to the people against the disease. In our work, the network is trained with different relevant input features without considering the vaccine information to predict the total number of cases, logarithm of total cumulative cases, active cases, deaths and various other targets. The research aims to check if the network shows different predictions for the western and the eastern districts which would then suggest that the vaccine information is a candidate for an invisible feature and hence provides immunity to the people against the virus. Chapter \ref{Chapter4} discusses in detail the preparation of the data set and the different strategies followed during the research. As using highly relevant data is important to correctly represent our model, high emphasis is given for feature selection. Chapter \ref{section : Result} finally discusses the results of the different strategies used and concludes based on the behaviour of the network.

\chapter{Deep Learning Approach} 
\label{Chapter2} 
Deep learning (also known as deep structured learning) is part of a broader family of machine learning methods based on artificial neural networks with representation learning. Learning can be supervised, semi-supervised or unsupervised.  \parencite{lecun2015deeplearning}. Deep learning architectures such as deep neural networks (DNN), recurrent neural network, convolutions neural network (CNN) have been increasingly used in the field of computer vision, speech recognition, machine vision, medical image analysis, social network filtering, games etc.,  where they have produced results comparable to and in some cases surpassing human expert performance. The most commonly used is the supervised learning, which has the task of learning a function that maps an input to an output based on example input-output pairs.  Supervised learning is where the input variables $x$ and the output variables $y$ are available and one has to use an algorithm to learn the mapping function from the input to the output. The goal is to approximate the mapping function so well that the new input data $x$ can be used to predict the output variables $y$ for that data. However, unsupervised learning is where only the input data $x$ is available without the corresponding output variables. The goal of unsupervised learning is to model the underlying structure or distribution in the data in order to learn more about the data. It allows the model to work on its own to discover patterns and information that was previously undetected. It mainly deals with the unlabelled data. In our research supervised learning is used since we need to find the mapping between the input and output variables via regression.
\newline
\par
Artificial neural networks (ANNs), usually simply called neural networks (NNs), are computing systems vaguely inspired by the biological neural networks that constitute animal brains. ANNs have gained widespread recognition as an effective machine learning algorithm by outperforming many algorithms such as Support Vector Machines in various relevant applications such as pattern recognition \parencite{SVM}. A deep neural network (DNN) is an artificial neural network (ANN) with multiple layers between the input and output layers \parencite{MAL-006}. The DNN finds the correct mathematical manipulation to turn the input into the output, whether it be a linear relationship or a non-linear relationship. A neural network is an architecture that comprises units named neurons. These architectures usually consist of three different layers: the input layer which contains the input feature vector; the output layer that consists of the neural network response; and the layer in between that contains the neurons that connect to both the input and output. An example of a neural network is a Feed-forward neural network which allows signals to travel from input to output.
\newline
\par
In this chapter, we will discuss the basics of deep feedforward networks and get into details on the training of neural networks. Then we discuss the different loss functions and the importance of feature selection for an efficient network.

\section{Deep Feedforward Networks}
\emph{Deep feedforward networks}, also called \emph{feedforward neural networks}, are classic deep learning models \parencite{Goodfellow-et-al-2016}. Usually, a feedforward network is trained to approximate some function $f^*$. As an example, we can consider a function $y=f^*(x)$ which classifies an input $x$ into a category $y$. A feedforward network will try to mitigate this function based on some parameters $\theta$. A feedforward network defines a mapping $y = f(x; \theta)$ using the parameters $\theta$ learned during training process. Via training, the feedforward networks learn the values of the parameters $\theta$ such that the network best approximates the function. The name feedforward suggests the forward flow of information i.e. from an input $x$ to some intermediate calculations within $f$ to get the final output $y$ \parencite{Goodfellow-et-al-2016}. In feedforward networks, these outputs $y$ are not fed again into the network. Being inspired by neuroscience, these are called neural, and they are referred to as networks, as they comprise many different functions \parencite{Goodfellow-et-al-2016}.
\newline
\par
A model usually consists of multiple functions that are combined together. For example, considering a function $f(x)$ composed of three different chain functions $f^{(1)}$, $f^{(2)}$ and $f^{(3)}$, such that $f(x)=f^{(3)}(f^{(2)}(f^{(1)}(x)))$. Neural networks are typically comprised of such chained structures \parencite{Goodfellow-et-al-2016} . The  $f^{(1)}$, $f^{(2)}$ and $f^{(3)}$ in this are the first layer, the second layer, and the third layer of the neural network respectively. The overall length of this chained structure, i.e. the total number of layers, defines the depth of the network. The layer is a generic term used to denote the collection of neurons that operate together at a specific depth in a neural network. Neurons are the building blocks of neural networks. A neuron takes one or more inputs and produces an output. In a neural network, the final layer is called the output layer, which in this case would be the third layer i.e.  $f^{(3)}$.
\newline
\par
\begin{figure}[ht]
    \centering
    \includegraphics[width=14cm, height=4cm]{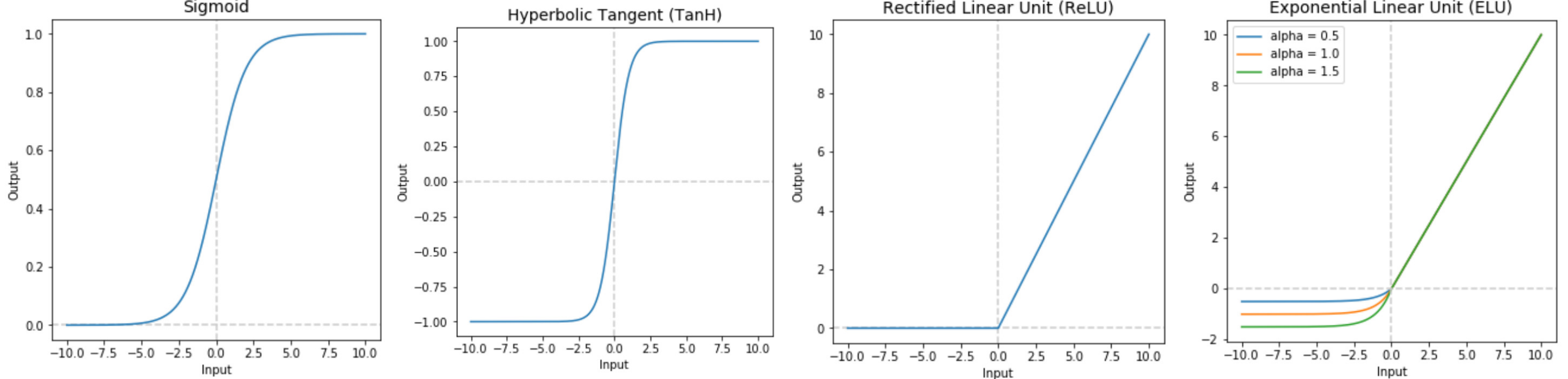}
    \caption{Plots of different activation functions: Sigmoid, Tamh, RelU, ELU (left to right) \parencite{activationpics}}
    \label{fig:activation}
\end{figure}

As stated above, the aim of training a network is to find parameters $\theta$, so that the network $f(x; \theta)$ best approximates the function $f^*(x)$. For this, we use various training points. These training points are together called training data. In the training data, each example $x$ is labelled with a value $y \approx f^*(x)$. In other words, training examples are the known data points i.e. we know what our desired model $f$ is supposed to output given the training data. The functionality of other layers is not directly governed by the labels of the training data. The results from the output layer are forced to match the label. As the training data does not govern the intermediate layers directly, they are called hidden layers. The outputs of hidden layers are also sometimes referred to as features \parencite{Goodfellow-et-al-2016}. To ensure that the deep learning model capacity is not just restricted to modelling linear functions, some non-linearity has to be introduced in the model. It is accomplished by applying linear transformations not directly to $x$, but to a transformed input $g(x)$, where $g$ is a non-linear transformation function, also called the activation function. Activation function decides whether a neuron should be activated or not by calculating weighted sum and further adding bias with it. The purpose of the activation function is to introduce non-linearity into the output of a neuron. Some examples of activation functions are
\begin{align} \label{eq:activation}
\begin{split}
    \text{Sigmoid} &: g(z) = \frac{1}{1+e^{-z}} \\
    \text{Tanh} &: g(z) = \frac{e^z-e^{-z}}{e^z+e^{-z}}\\
    \text{Softmax} &: g(z_i)= \frac{e^{z_i}}{\sum_{i=1}^{n}e^{z_i}}\\
    \text{ReLU}  &:  g(z)=max(0,z)\\
    \text{ELU} &: g(z)=\left\{
	\begin{array}{ll}
		z  & z > 0\\
		\alpha(e^{(z)}-1) & z \leq 0 	
	\end{array}
\right\}
\end{split}
\end{align}
In the activation function examples in Eq. \ref{eq:activation}, the sigmoid activation function takes a real value as an input and then outputs a value between 0 and 1. It is non-linear, continuously differentiable, monotonic, has a fixed output range and is a good classifier. Tanh squashes a real-valued number to the range [-1, 1]. Unlike Sigmoid, its output is zero-centred. Therefore, in practice, the tanh non-linearity is always preferred over the nonlinearity of sigmoid. Softmax is typically used as an activation function for the output layer of a classification network. The result from the softmax could be interpreted as a probability distribution. Softmax function calculates the probability distribution over ‘n’ different events. In general, this function will calculate the probabilities of each target class over all the possible target classes. Later, the calculated probabilities will help determine the target class for the given inputs. Rectified Linear Units (ReLU) activation function just compares the input value with zero \parencite{activation}. The formula is deceptively simple: max(0,z). Despite its name and appearance, it’s not linear and provides the same benefits as sigmoid but with better performance. Finally, Exponential Linear Unit, also known as ELU activation function tend to converge the cost to zero faster and produces accurate results. Unlike the other activation functions, ELU has an extra constant 'alpha' $\alpha$ which should always be a positive number. ELU is similar to ReLU except for negative inputs.  They are both an identity function form for non-negative inputs. On the other hand, ELU becomes smooth slowly until its output equals to $-\alpha$ whereas ReLU sharply smoothens. ELU produces activations instead of letting them be zero when calculating the gradient. Figure \ref{fig:activation} shows the plots of different activation functions. ELU activation gives promising results for regression problems. As our task is to primarily build a neural network to predict the coronavirus cases and deaths, ELU activation function is used throughout the research.

\begin{figure}[ht]
    \centering
    \includegraphics[width=9cm, height=6cm]{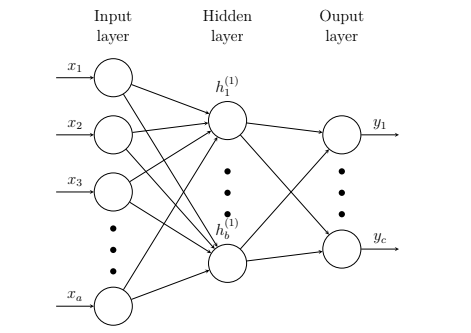}
    \caption{An Example of a feedforward network. The network input is $x=[x_1,x_2,...,x_a]$ and the output is $y=[y_1,y_2,...,y_c]$ \parencite{Goodfellow-et-al-2016}}
    \label{fig:network}
\end{figure}

Due to the chained structure of feedforward network, the output of a layer is the input for the next layer. Each layer could be considered as a non-linear mapping of vectors \parencite{Goodfellow-et-al-2016}. If a layer contains $c$ neurons and the previous layer comprises $b$ neurons, then the mapping by the layer with $c$ neurons is $\mathbb{R}^b \rightarrow \mathbb{R}^c$. For the first layer, the input would directly be the network input i.e. $x$ and the output from first layer is $$h^{(1)}=f^{(1)}(x)=g(W^{(1)}(x)+b^{(1)})$$, where $W^{(1)}$ is the 2D matrix having the weights used for the linear transformation, while $b^{(1)}$ is the vector containing the bias for each neuron, and $g$ is the activation function \parencite{activation}. Weights and biases are learnable parameters. We continue feeding this network into the forward direction i.e., the output from this layer is given as an input for the next layer. Hence, for any layer $l$, forward pass can be expressed as $$h^{(l)}=f^{(l)}(h^{(l-1)})=g(W^{(l)}(h^{(l-1)})+b^{(l)})$$, where $l \in \{0,1,.....,L \}$ \parencite{Goodfellow-et-al-2016}. $L$ here being the total number of layers in the network. Output from the last layer would be the output of network, i.e. $$y=f^{(L)}(h^{(L-1)})=g(W^{(L)}(h^{(L-1)})+b^{(L)})$$ If $n^{(l)}$ is the number of neurons in the layer $l$, then dimensions of matrix $W^{(l)}$ would be $n^{(l)} \times n^{(l-1)}$ and the size of vector $b^{(l)}$ is $n^{(l)}$. An example of a feedforward network can be seen in Figure \ref{fig:network}. In this example, the input vector has $a$ elements, hence there is $a$ number of neurons in the input layer. There is a hidden layer with $b$ neurons and finally the output layer with $c$ neurons. Fully connected networks are a type of feedforward networks. As the name suggests, fully connected networks are the feedforward networks, in which the layers of the network are fully connected.

\subsection{Training Feedforward Networks}
The goal of training the network is to find the parameters for our network that best approximates the function $f^*$. For this, we first need a loss function that can compare how far the network output $f(x;\theta)$ is from the target output $f^*(x)$. Then we minimize the loss via training. For the classification problem, where the network predicts the probabilities of input belonging to each class, loss based on the cross-entropy is mostly used. For using cross-entropy loss, the network should predict the probabilities in the classification problem. The cross-entropy is just negative log-likelihood of the probabilities predicted by the network. If $y$ is the network output for an example input $x$ to the network $f$ parameterized by $\theta$, and the label for the example is $t$, the cross-entropy loss is described by
$$J(\theta)=\sum_{c=1}^{C} t_c log(f(c;\theta)_c)$$.

In the above equation $c \in \{1,2,...,C\}$, where $C$ is the total number of classes. The subscript $c$ for the label and network output denote the $c^{th}$ component of the vector. With training, we minimize $J(\theta)$, i.e. training a neural network is an optimization problem. For this optimization, the algorithm used is called \emph{backpropagation} \parencite{LeCunBoserDenkerEtAl89}.
\newline
\par
Backpropagation \parencite{LeCunBoserDenkerEtAl89} is a gradient-based algorithm. It uses the gradient of the loss function $J(\theta)$ with respect to the parameters $\theta$ or with respect to the activation of each layer. The backpropagation is briefly explained in the following steps: 
\begin{enumerate}
    \item Forward pass: We first pass the input sample $x$ through the network to get network output, and then we compute the loss function $J(\theta)$ \parencite{NN}
    \item Computing intermediate term: We now aim to compute the gradient of the loss function with respect to the activation of each layer. We start by computing the gradient with respect to the activation of the last layer of the network. Then we use the chain rule to compute gradients with respect to activation of previous layers. To calculate gradients, we first compute a term $\delta^{(L)}$ for the last layer \parencite{NN}.
    $$\delta^{(L)} = \frac{\partial J(\theta)}{\partial h^{(L)}} \odot g'(W^{(L)}h^{(L-1)}+b^{(L)}) $$
    The operator $\odot$ is the Hadamard product, which represents element-wise multiplication. Now, we propagate backwards and calculate intermediate terms with respect to activations of preceding layers. For any layer $l \in \{1,2,...,L\}$, the intermediate term $\delta^{(l)}$ is calculated using \parencite{NN}
    $$\delta^{(l)} = W^{(l+1)^T}\delta^{(l+1)} \odot g'(W^{(l)}h^{(l-1)}+b^{(l)}) $$.
    \item Computing the gradients: Now we compute gradients with respect to the parameters i.e. weights and bias, as \parencite{NN}
    \begin{align*}
       \frac{\partial J(\theta)}{\partial W^{(l)}} =& \delta^{(l)} h^{(l-1)^T} \\
       \frac{\partial J(\theta)}{\partial b^{(l)}} =&\delta^{(l)}
    \end{align*}
    \item Optimize: Once we have the gradients, we can apply any gradient-based optimization algorithm like gradient descent to update the weight and bias. We repeat the steps until a local minimum of loss function $J(\theta)$ is reached. We also have to provide the learning rate for the optimization algorithm. Very large learning rates result in the optimizer taking large steps towards the local minima, but it could be highly unstable. On the other hand, if the learning rate is too low, the model might take longer to reach minima. Hence, to find adequate learning rate, the model is trained using various different learning rates. In batch gradient descent, all the training data is taken into consideration to take a single step. We take the average of the gradients of all the training examples and then use that mean gradient to update our parameters. So that’s just one step of gradient descent in one epoch. Another commonly used optimization algorithm is the mini-batch gradient descent. In this, instead of using all the input data together, only a small subset of data called a mini-batch is used at a time to update the weights \parencite{Goodfellow-et-al-2016}. We repeat the steps and update the parameters using the mini-batches until all the data has been used once. This is called one epoch. Iterating over the whole dataset once more would be two epochs and so on. In Stochastic Gradient Descent (SGD), we consider just one example at a time to take a single step. Depending on the available data and the model, different algorithms can be used for optimization. Modified variants of SGD are also used for better performance \parencite{journals/nn/Qian99}. SGD with momentum is one such modification that helps accelerate the optimization process in the right direction, which results in faster convergence. Applying momentum makes updates not just based on the current gradient, but it also takes into account, the previous gradients \parencite{journals/nn/Qian99}. 
\end{enumerate}

In machine learning, the learning of the target function from training data is referred to as inductive learning. Generalization refers to how well the concepts learned by a machine learning model apply to specific examples not seen by the model when it was learning. The goal of a good learning model is to generalize well from the training data to any data from the problem domain. This allows us to make predictions in the future on data the model has never seen. Overfitting and underfitting are the two biggest causes for the poor performance of machine learning algorithms. Overfitting occurs when the model learns in detail the training set including the noise to an extent that it negatively impacts the performance. The network, therefore, fails to generalize well as it learns noise and fluctuations and therefore gives bad results. Overfitting results in a good performance on the training data but poor generalization to other data. Underfitting refers to a model that can neither model the training data nor generalize to new data. It results in poor performance on the training data and poor generalization to other data. Underfitting is easy to detect and can easily be rectified by either including more data when available or by using a different algorithm. Overfitting is a major problem in neural networks. This especially is true in modern networks, which often have very large numbers of weights and biases. Increasing the training data or reducing the network size are basic ways to reduce overfitting. To limit overfitting k fold cross-validation can be used. Fortunately, there are other techniques which can reduce overfitting, even when we have a fixed network and fixed training data. These are known as regularization techniques. L1 and L2 regularizations are the most commonly used regularization techniques. The idea of L1 and L2 regularization is to add an extra term to the cost function, a term called the regularization term \parencite{NN}. Early stopping and dropouts can also be used to regulate our network. Since we deal with a large amount of data, overfitting becomes an increasing threat and must be avoided for good predictions.

\section{Loss functions}
One key feature of our suggested method is the choice of the loss function. In any deep learning algorithm, the loss function configuration is one of the most important steps to ensure the model will work in an intended manner. The loss function can give a lot of practical flexibility to the neural networks and will define how exactly the output of the network is connected with the rest of the network. All the algorithms in machine learning try to minimize or maximize the objective function. The group of functions that are minimized are called the loss functions. A loss function is a measure of how good the model functions in terms of predicting the expected target. There is not a single loss function that works for all kind of data. It depends on a number of factors including the presence of outliers, choice of the machine learning algorithm, time efficiency of gradient descent, ease of finding the derivatives and confidence of predictions. Loss functions can be classified into 2 types: classification and regression loss. \parencite{loss} investigated some representative loss functions and analysed the latent properties of them. The main goal of the investigation was to ﬁnd the reason why bilateral loss functions are more suitable for regression task, while unilateral loss functions are more suitable for classiﬁcation task. This section covers in detail the different loss functions which can be used for our regression problem as discussed by \parencite{lossfuncations}.

\subsection{Mean Square Error (MSE), Quadratic loss, L2 Loss}
This function originates from the theory of regression, least-squares method. Mean Square Error (MSE) is the most commonly used regression loss function. MSE is the sum of squared distances between our target variable $y$ and predicted values $y_{p}$.
\begin{equation}
\mathbf{M.S.E.} = \frac{\sum_{i=1}^{n} (y^{i}-y_{p}^{i})^{2}}{n}
\end{equation}
It is well known that, here, few distant points outweighs the closer points. MSE is sensitive towards outliers and given several examples with the same input feature values, the optimal prediction will be their mean target value.  The MSE is great for ensuring that our trained model has no outlier predictions with huge errors since the MSE puts larger weight on these errors due to the squaring part of the function. MSE is thus good to use if the target data, conditioned on the input, is normally distributed around a mean value, and when it’s important to penalize outliers extra much. It has a continuous derivative and therefore the minimisation with gradient methods works well. The main disadvantage of MSE is when the model makes a single very bad prediction which would magnify the error due to squaring. 

\subsection{Mean Absolute Error (MAE), L1 Loss}
Mean Absolute Error (MAE) is just the mean of absolute errors between the actual value $y$ and the value predicted $y_{p}$.  So it measures the average magnitude of errors in a set of predictions, without considering their directions.
\begin{equation}
\mathbf{M.A.E.} = \frac{\sum_{i=1}^{n}|(y^{i}-y_{p}^{i})|}{n}
\end{equation}

\begin{figure}[ht]
  \centering
  \begin{tabular}[b]{c}
    \includegraphics[width=6cm]{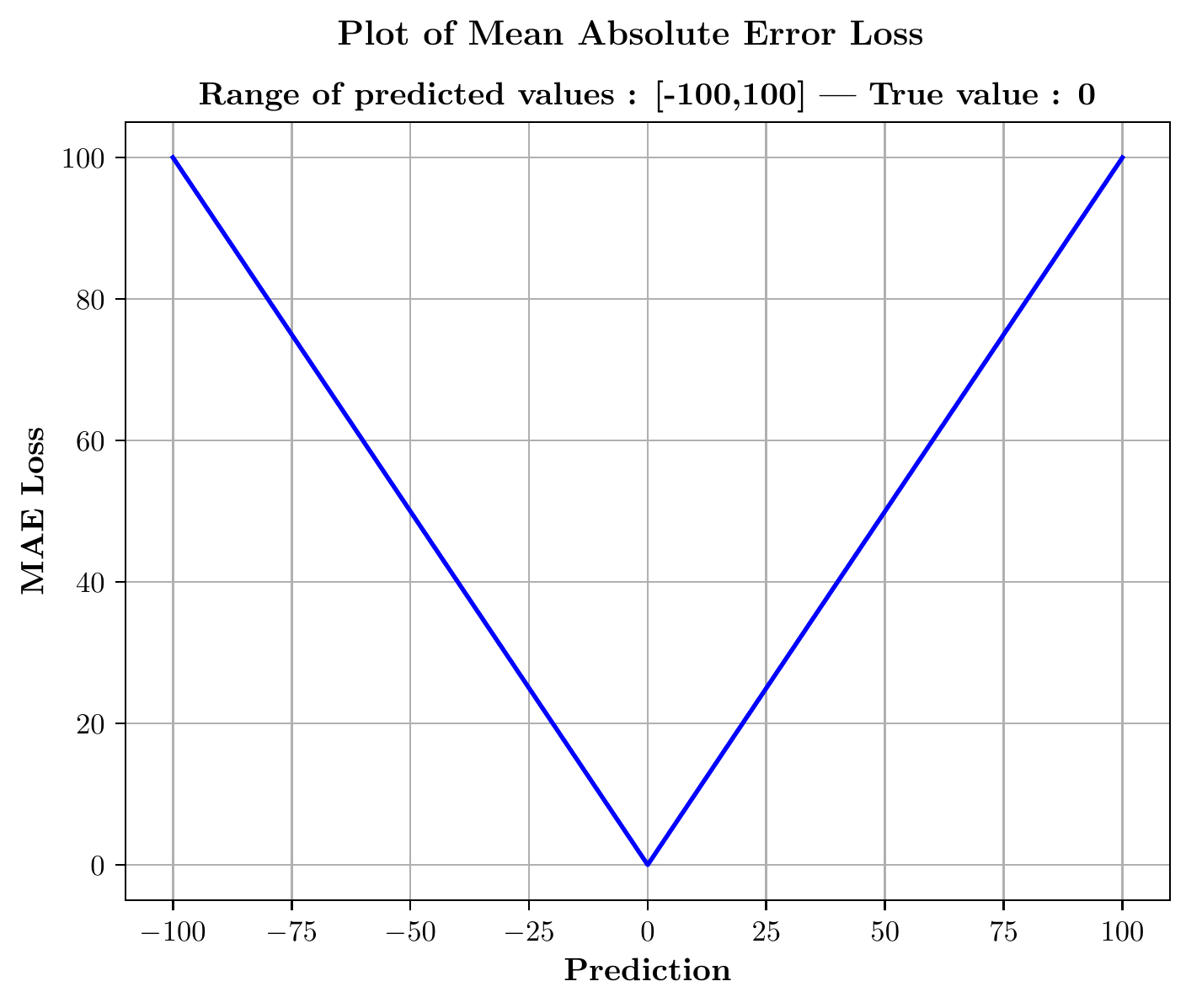} 
  \end{tabular} \qquad
  \begin{tabular}[b]{c}
    \includegraphics[width=6cm]{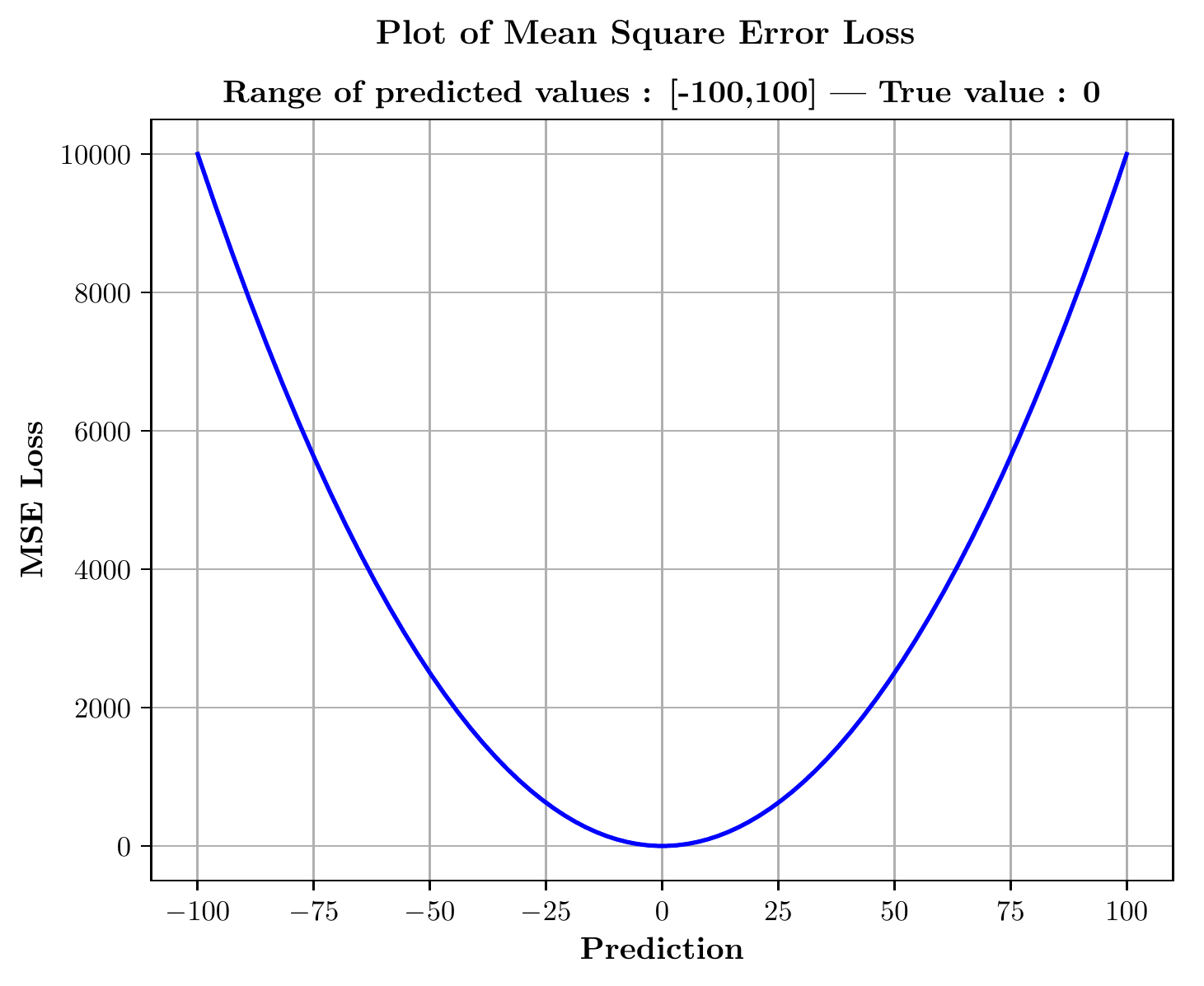} 
  \end{tabular}
  \caption{Plot of Mean Absolute Error (left) vs Mean Square Error (right)}
  \label{fig:MAE_MSE}
\end{figure}

As one can see, for this loss function, both the big and small distances contribute the same. The advantage of MAE covers the disadvantage of MSE. As we consider the absolute value, the errors will be weighted on the same linear scale. Therefore, unlike the previous case, MAE doesn't put too much weight on the outliers and the loss function provides a generic and even measure of how well our model is performing. However, it does not have a continuous derivative and thus does not always provide a stable solution. Figure \ref{fig:MAE_MSE} compares the plots of mean absolute error and the mean square error loss functions where the true target value is 0, and the predicted values range between -100 to 100. The loss (Y-axis) reaches its minimum value at prediction (X-axis) = 0. The range is 0 to $\infty$. During gradient descent, the MSE does a better job in finding the minima as it has a continuous derivative and provides a stable solution when occasional outliers don't exist.\\
\par
Now, since our data might have some outliers, we would not want our predictions to be biased towards these outliers (by using $\mathbf{M.S.E.}$), nor do we want to ignore the outliers (by using $\mathbf{M.A.E.}$). Hence, we need to use some other loss function for our problem.

\subsection{Huber Loss, Smooth Mean Absolute Error}
Huber loss is just the absolute error but transforms to squared error for small values of error. Huber loss is less sensitive to outliers in data than the squared error loss. It’s also differentiable at 0. It’s basically absolute error, which becomes quadratic when the error is small. How small that error has to be to make it quadratic depends on a hyperparameter $\delta$, which can be tuned. Huber loss approaches MSE when $\delta \rightarrow 0 $ and MAE when $\delta\rightarrow \infty $ (large numbers). It is defined as 
\begin{equation}
L_{\delta}(y,y_{p}) =
\left\{
	\begin{array}{ll}
		\frac{1}{2}(y-y_{p})^{2}  & \mbox{if } |y-y_{p}| \leq \delta\\
		\delta|y-y_{p}|-\frac{1}{2}\delta^{2} & \mbox{otherwise } 	
	\end{array}
\right\}
\end{equation}
The choice of $\delta$ becomes increasingly important depending on what one considers as an outlier. Residuals larger than delta are minimized with L1 while residuals smaller than delta are minimized with L2. Hubber loss combines the advantages of both the loss functions. It can be really helpful in some cases, as it curves around the minima which decreases the gradient. However, the problem with Huber loss is that we might need to train hyperparameter delta which is an iterative process.

\subsection{Log-Cosh Loss} \label{logcoshloss}
Log-cosh is another loss function used in regression tasks which is smoother than L2. Log-cosh is the logarithm of the hyperbolic cosine of the prediction error. Given the actual value $y$ and the predicted value $y_{p}$, the log-cosh is defined as 
\begin{equation}
L(y,y_{p}) = \sum_{i=1}^{n}|\operatorname{log}(cosh((y^{i}-y_{p}^{i})))|
\end{equation}

\begin{figure}[ht]
    \centering
    \includegraphics[width=7cm]{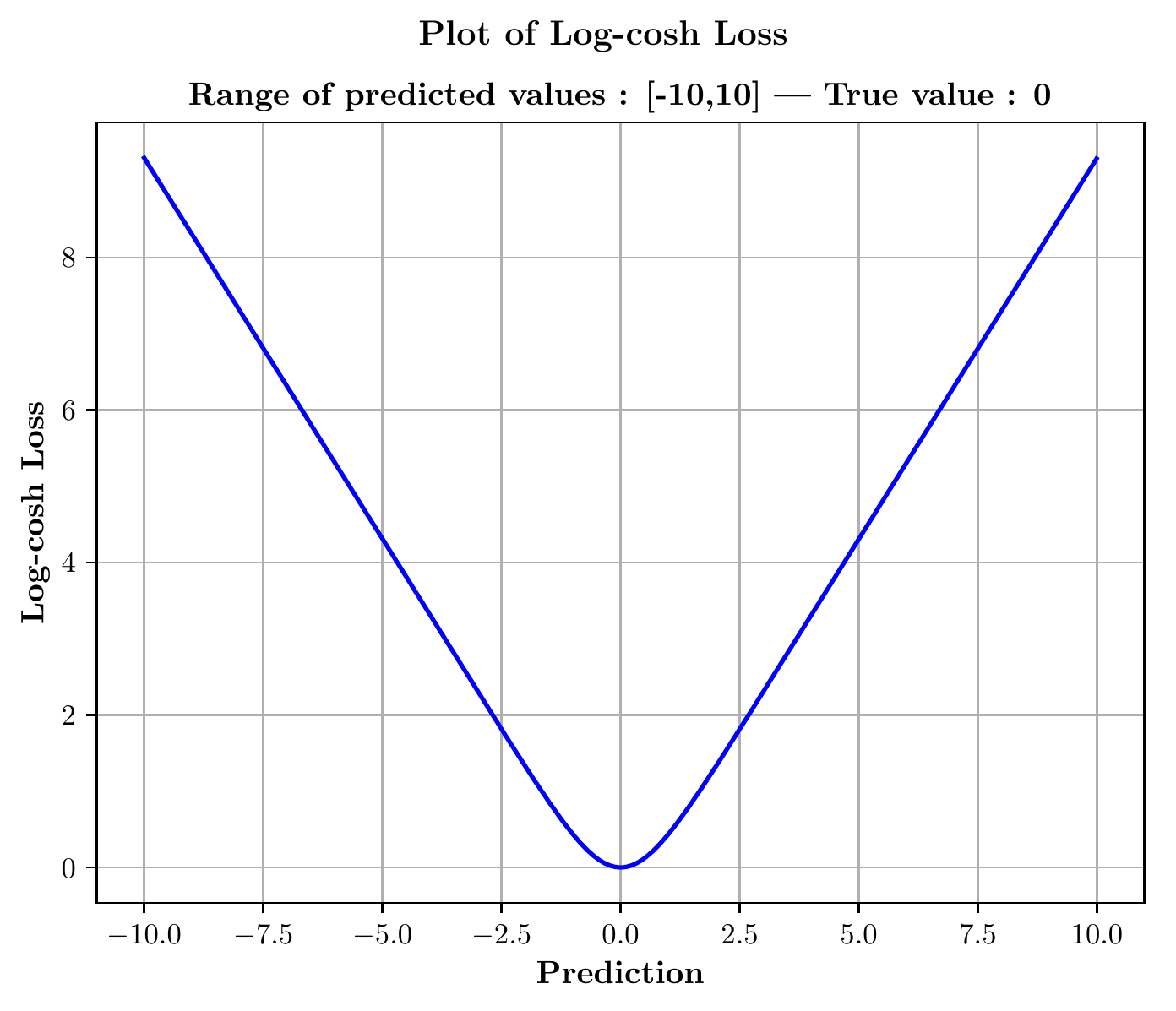}
    \caption{Plot of log-cosh loss function}
    \label{fig:logcosh}
\end{figure}

$\operatorname{log}(cosh(x))$ is approximately equal to $\frac{x^2}{2}$ for small values of x and to $|x|-\operatorname{log}(2)$ for larger values. Therefore, the log-cosh loss function is similar to mean square error but will not be largely affected by occasional wrong predictions. It is twice differentiable everywhere unlike Huber loss. Figure \ref{fig:logcosh} shows the plot of logcosh loss vs the prediction. In our research, log-cosh loss function was used as it showed a positive result in classifying the data based on the hidden features. The logcosh loss function for neural networks has been developed to combine the advantage of the absolute error loss function of not overweighting outliers with the advantage of the mean square error of continuous derivative near the mean, which makes the last phase of learning easier. Also, when clustered data is present, an artificial neural network with logcosh loss function learns the bigger cluster rather than the mean of the two and hence can be used to classify the clustered data. In the case of MSE, due to the squaring of the error function, few faraway points are weighted more than the nearby points. When learning clustered data, the network with MSE loss function gets affected by these outlying clusters and tries to find the minima between them and thereby fails to learn the bigger cluster.  However, for linearly growing loss functions like logcosh and MAE, just the sum of distances count and few far-away points do not count more than several nearby points and therefore, a regression value near or through the heavier cluster is learnt.  Though the MAE loss function tries to learn the bigger cluster, it is non-smooth and has a non-continuous derivative resulting in oscillating behaviour.  As MAE is unstable during the last part of minimization, it oscillates between the clusters. As mentioned above, since the logcosh loss function is a combination of MAE for larger values and MSE for the smaller values, it successfully learns the bigger cluster and gives a stable solution. This feature of logcosh loss function is exploited in our research.

\section{Feature Selection and Correlation}

Machine learning models are highly depended on the selected data. That is the reason data scientists spend hours pre-processing and cleansing the data. Only the features that would best represent the selected model is selected. This process is called feature selection. Feature Selection is the process of selecting the attributes that can make the predicted variable more accurate or eliminating those attributes that are irrelevant and can decrease the model accuracy and quality. Data and feature correlation is considered as an important step in feature selection. In statistics, correlation or dependence is any statistical relationship, whether causal or not, between two random variables or bivariate data. In the broadest sense, correlation is any statistical association, though it commonly refers to the degree to which a pair of variables are linearly related. Using correlation, one can get insights on how different attributes depend on each other or forms a cause for another attribute. Features with high correlation are more linearly dependent and hence have almost the same effect on the dependent variable. So, when two features have a high correlation, we can drop one of the two features.
\newline
\par
Feature selection methods are intended to reduce the total number of input variables to a minimum number of input variables which are believed to be the most useful to represent the model. Unsupervised feature selection does not use the target variable and is mainly used in methods to remove redundant variables whereas, in the supervised feature selection, the target variables are utilised to remove irrelevant variables.   Filter-based feature selection methods use statistical measures to score the correlation or dependence between input variables that can be filtered to choose the most relevant features. Statistical measures for feature selection must be carefully chosen based on the data type of the input variable and the output or response variable \parencite{FeatureSelection}.
\newline
\par
Correlation among the input features and among the input features and the target variables becomes an important measure in feature selection. Correlation can help in predicting one attribute from others, which is a great way to input the missing values. It is also used to reduce the redundancy in the input variables. Positive Correlation means if a feature A increases, then the feature B also increases i.e. both features move in tandem, and they have a linear relationship. In the other way round, a negative correlation means that if a feature A increases then the feature B decreases and vice versa. If the chosen data set has a perfectly positive or negative correlation, there is a high probability that the performance of the model is impacted by \emph{Multicollinearity}. Multicollinearity happens when one predictor variable in a multiple regression model can be linearly predicted from the others with a high degree of accuracy. This can lead to skewed or misleading results \parencite{FeatureCorrelation}. Therefore, the data should be analysed for correlation and only the required input feature must be selected. For high dimensional data sets, heat maps are usually drawn to analyse the data before training. A perfect correlation of 1 rarely exists and hence any correlation above 0.7 is considered to be highly correlated. It is always advised to select data with mixed correlation (both positive and negative correlation) when selecting the input features.
\newline
\par
However, correlation is often considered as causation which is a big misconception. Any highly correlated variables must be examined carefully. Correlations are very useful in many applications, especially when conducting regression analysis. However, it should not be mixed with causality and misinterpreted in any way. One should also always check the correlation between different variables in the data set and gather some insights as a part of the exploration and analysis.

\chapter{Classification of multi-valued functions} 
\label{Chapter3} 

In mathematics, a multi-valued function, also called multi-function, many-valued function, set-valued function, is similar to a function but may associate several values to each input. Since these functions have several outputs $y$ for a given input $x$, it is difficult to classify them based on multivaluedness. This chapter discusses in detail the work of \parencite{classification} to find hidden-feature depending laws inside a data set and classifying the data based on the hidden features using Neural Network. Section \ref{section:1} discusses the basic problem setting. In this chapter, we create a multi-valued data set and examine the behaviour of the neural network trained using this data set. Our focus in this section is to not only build an efficient neural network for our chosen problem but also to examine the behaviour of the network on the multi-valued data set. To simplify the hypothesis, a simple 1 dimensional and a 2-dimensional problem was selected. The constructed data set was split into training and testing data, mostly in the ratio of 80:20. The network was trained with the 80 percent of data which was then tested on the remaining 20 percent of the data to check the behaviour of the network. In this chapter, we aim to classify the data by using a network with logcosh loss function. This forms the base of the experiment that we followed to check if the tuberculosis vaccine indeed provided immunity for the COVID-19. 

\section{Problem Setting}
\label{section:1}
Let 
\begin{equation}
\left\{ (\vek x, \vek y) \in\realnr^d \times \realnr^{d} \right\}
\end{equation}
be a set of data points. It is assumed that $\vek y$  depends on $\vek x$. 
To simplify the setting and also due to the fact that artificial neural networks do not encourage vector valued output, 
we restrict ourselves to 
\begin{equation}
X = \left\{ (\vek x, \vek y) \in \realnr^d \times \realnr \right\}
\end{equation}
Let 
\begin{equation}
X\subset \Omega \subset \realnr^d \times \realnr 
\end{equation}
be a subset that contains the data points where $\Omega$ is the domain where the function 
\begin{align}
\label{1}
\Phi_i: X & \longrightarrow  \realnr \\ \label{2}
\vek x &\mapsto  \Phi_i(\vek x) =  Y(\vek x)
\end{align}
is defined and $\Phi$ is the function that maps $X$ and $Y$ is such that one parameter $x$ might have multiple outputs $y$. Now we consider  
\begin{align}
X_i\subset& X, \quad i = 1,...,N<\infty\\
\phi_i: X_i & \longrightarrow  \realnr \\
\vek x &\mapsto  \phi_i(\vek x) =  y(\vek x)
\end{align}
where $\phi_i$ is a single valued function, for which each point in the domain, has a unique value in the range.
\newline
\par
Now, the data is such that, one parameter $\vek x$ may have more than one image under $\Hat{\phi}$ -- 
 some $\vek x$ may have distant $\vek y$ in spite of being very close to each other. If looking for continuous functions $\Phi$, this means that one looks for a possibly  \emph{set-valued} function: 
\begin{align}
\Hat X_i\subset& X, \quad i = 1,...,M<\infty\\
\Hat{\phi}_i:
\Hat X_i & \longrightarrow  \realnr \\
\vek x &\mapsto \Hat{\phi}_1(\vek x) = y_1(\vek x)\\
\vek x &\mapsto \Hat{\phi}_2(\vek x) = y_2(\vek x)
\end{align}
where each component of $\Hat{\phi}$  represents one of the possible outcomes which are not distinguishable by $\vek x$ a priori, for  which, a rule is valid depending on some hidden property: 
The set-valued function captures the property that the data input-output pairs indeed belong to different situations or populations. It is known in the beginning which data belongs to which situation, or the different rules for the different outcomes of $\hat{\phi}(\vek x)$. We assume that the entries of $\hat{\phi}$ exist and possess some smoothness. They may coincide in parts of $\Omega$, i.e. there exists a certain set of $\{X\}$ which gives the same output $\{Y\}$ and the remaining set which gives a multi-valued output. The set-valuedness in this nomenclature is expressed by this vector-valuedness. $\Phi$ can be re-written as a combination of $\phi$ and $\Hat \phi$ as 
\begin{align}
\Phi (\vek x) = \left\{ \begin{array}{cc} 
                \phi \\
                \Hat{\phi}\\
                \end{array} \right\}.
\end{align}
\newline
\par
The task to solve in this nomenclature is: Given the set $X$, find the rules $ \Phi_i$ and the subsets $X_i$ where they are valid.
\newline
\par
The discussed problem setting creates a multi-valued data set which can be otherwise described as clustered data. As for some $x$, there are 2 possible $y$, the artificial neural network trained with this as an input considers it as clustered data. As discussed in section \ref{logcoshloss}, the logcosh loss function can be used to train such a setting to classify the bigger cluster. The logcosh loss function combines the advantages of both MAE and MSE and therefore manages the outliers much better and exhibits good performance during gradient descent. The network trained with logcosh loss tries to learn the bigger cluster efficiently, without being affected by the smaller clusters and thus classifies the clustered data. In our research, we give different weights to the clusters and train the network with logcosh loss function in an aim to classify the clustered data \parencite{classification}.  This approach is demonstrated using a simple 1-dimensional and 2-dimensional problem.

\section{1 dimensional case}
We now consider a simple 1D example based on the concept discussed in the section~\ref{section:1}. Two simple single-valued polynomial functions were selected and combined in different fractions to achieve a multi-valued data set. This section discusses the problem setting of the 1-dimensional case and thereafter the network behaviour based on the chosen data set.
\subsection{Network Architecture}
\label{subsection:Network1D}
A simple fully connected neural network was used for the problem. 2000 data points were considered to form the multi-valued data set out of which 80\% i.e. 1600 data pairs were used for training and the remaining 400 data points were used for testing. The data set was split randomly. The fully connected network contained 1 neuron in the input and 1 neuron in the output as the input and output dimension was 1. It contained 4 hidden layers with 50 neurons each with the ELU activation function. Logcosh loss function and Adam optimiser with an initial learn rate of $10^{-3}$ was used for optimization. A batch size of 32 was used and trained for 100 epochs. Finally, the concept of early stopping was used to avoid overfitting.

\subsection{Training Strategy}
To create a multi valued Data set, 2 simple functions were selected as below.
\begin{equation}
f_{1}(\vek x) = {((x-4)(x+4))}^2,\hspace{5mm} x \in [-6,6]
\end{equation}
\begin{align}
f_{2}(\vek x) = \left\{ \begin{array}{cc} 
                {((x-4)(x+4))}^2 & \hspace{5mm} x \in [-6,-4) \\
                0 & \hspace{5mm} x \in [-4,4] \\
                {((x-4)(x+4))}^2 & \hspace{5mm} x \in (4,6] \\
                \end{array} \right\}.
\end{align}

where $f_1$ and  $f_2$ are two single-valued functions which are defined within the interval $[-6,6]$. The data set was split such that $80\%$ of the data were used for training and the rest $20\%$ were used as test data. Initially, both the functions were trained individually with a basic regression neural network as discussed in  section~\ref{subsection:Network1D} and then tested on the test data to validate the network.
\newline
\par
\begin{figure}[ht]
  \centering
  \begin{tabular}[b]{c}
    \includegraphics[width=6cm]{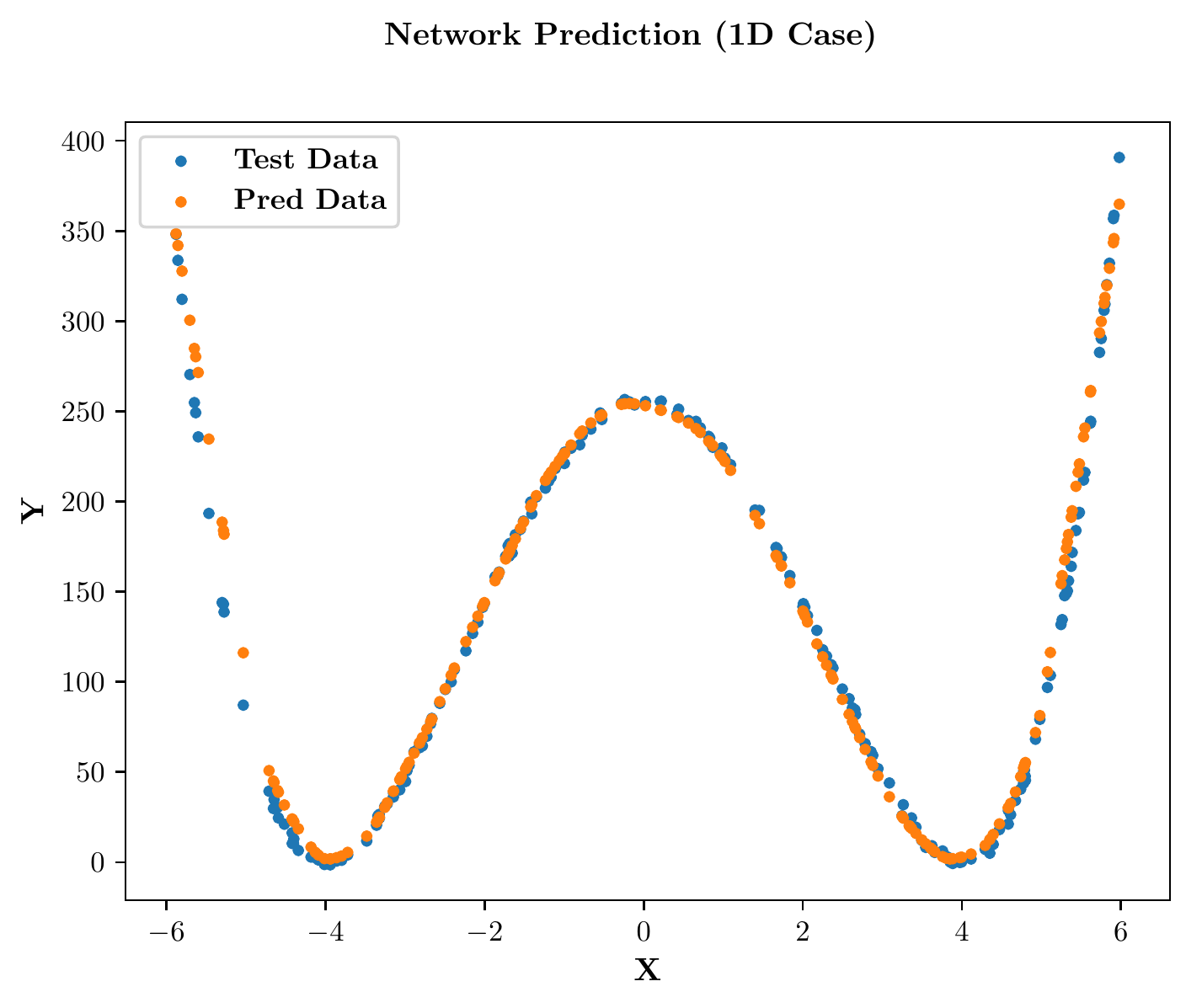} \\
    \small (a) $f_1(\vek x)$
  \end{tabular} \qquad
  \begin{tabular}[b]{c}
    \includegraphics[width=6cm]{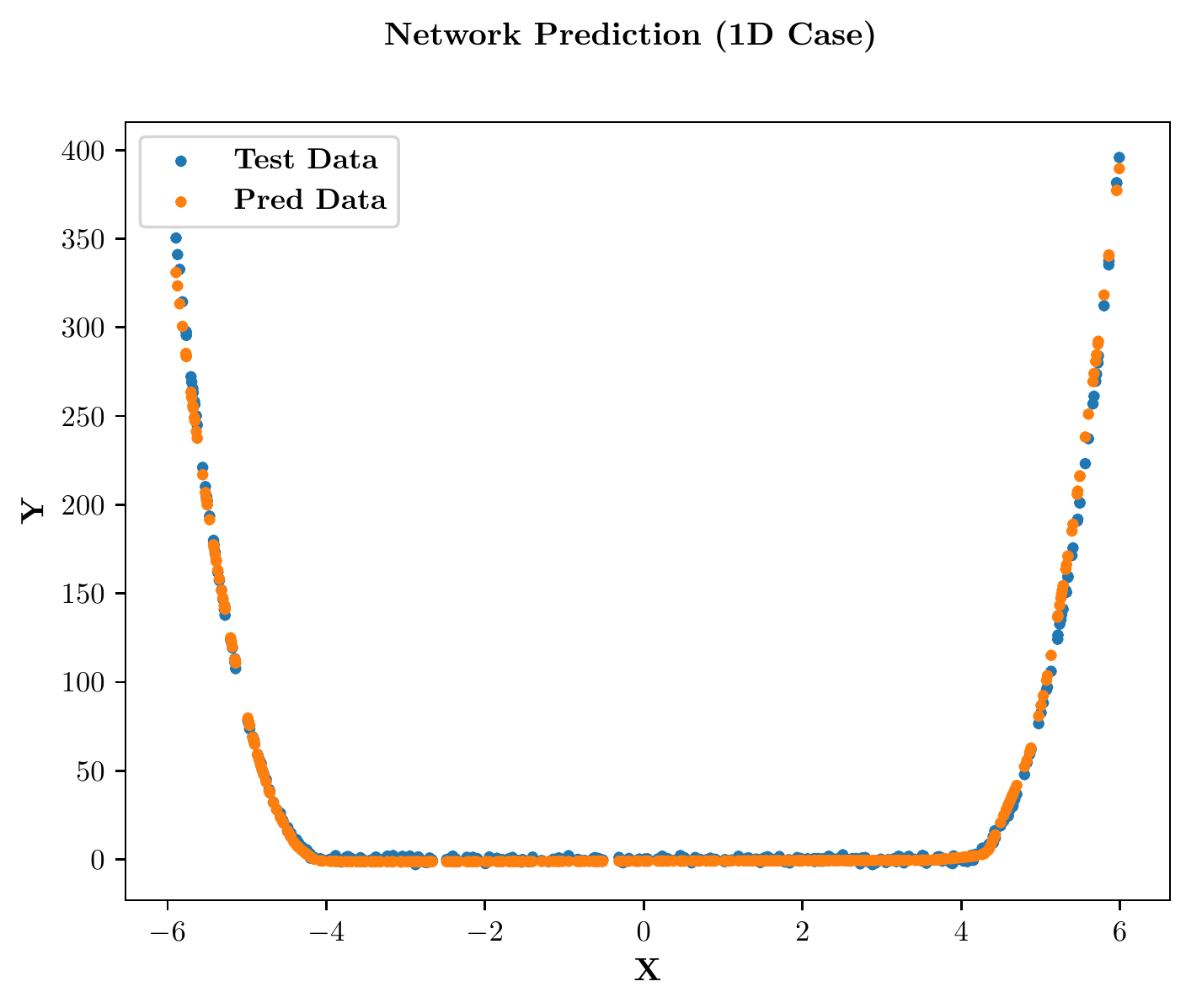} \\
    \small (b) $f_2(\vek x)$
  \end{tabular}
  \caption{Plot of Test and Predicted Data for functions $f_1$ and $f_2$ for 1 dimensional case.}
  \label{fig:3.1}
\end{figure}

As seen in figure \ref{fig:3.1}, it is clear that the neural network was able to approximate the given functions by reducing the loss function to the minimum. As discussed in section~\ref{section:1} we need to set up a multi-valued data set i.e. we combine a fraction of both the functions $f_1$ and $f_2$ to form a new data set as per our requirement. The two data sets were combined in different fractions, trained using our neural network and then tested on the test data which is $20\%$ of the new combined data. The noise was added to the data set to replicate the real-world data. The network was trained using logcosh loss function to examine how the network behaves when clusters of data exists. By using logcosh loss function, we aim to classify the clusters of data.
\newline
\par
\begin{figure}[ht]
    \centering
    \includegraphics[width=7cm]{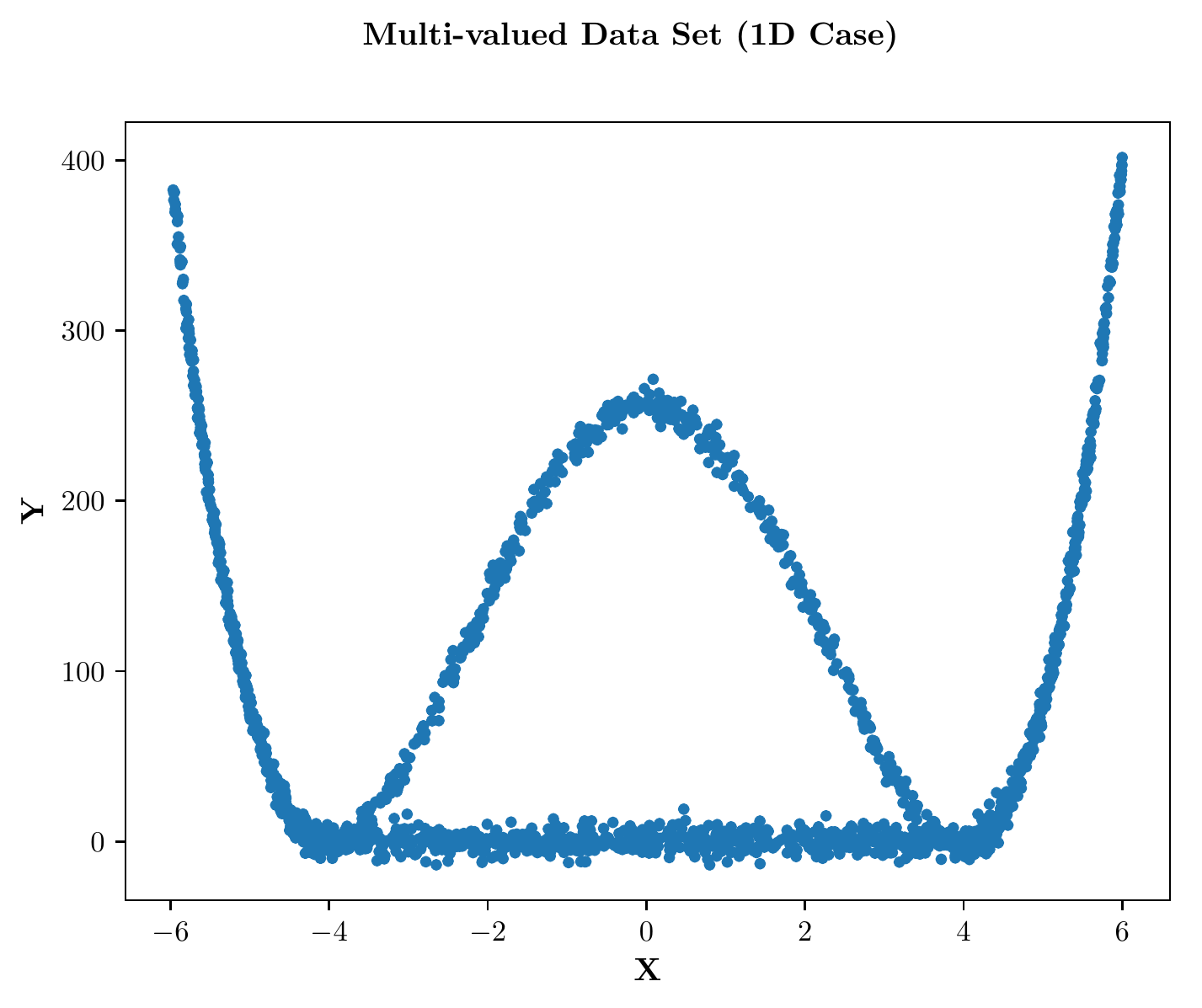}
    \caption{Plot of data set with noise for 1 dimensional case}
    \label{fig:3.2}
\end{figure}

The combined data set can be written as follows :
\begin{align}
\Phi (\vek x) = \left\{ \begin{array}{cc} 
                \phi & \hspace{5mm} x \in [-6,-4) \\
                \Hat{\phi} & \hspace{5mm} x \in [-4,4] \\
                \phi & \hspace{5mm} x \in (4,6] \\
                \end{array} \right\}.
\end{align}
where $\Phi(\vek x)$ is a combination of both single and multi-valued function. $\phi$ represents the common region in the interval $[-6,4)$ and $(4,6]$ and $\Hat{\phi}$ represents the multi-valued region where each $\vek x$ has two possible outputs $\vek y$ as shown in the figure~\ref{fig:3.2}. 

\subsection{Network behaviour}
\begin{figure}[ht]
  \centering
  \begin{tabular}[b]{c}
    \includegraphics[width=6cm]{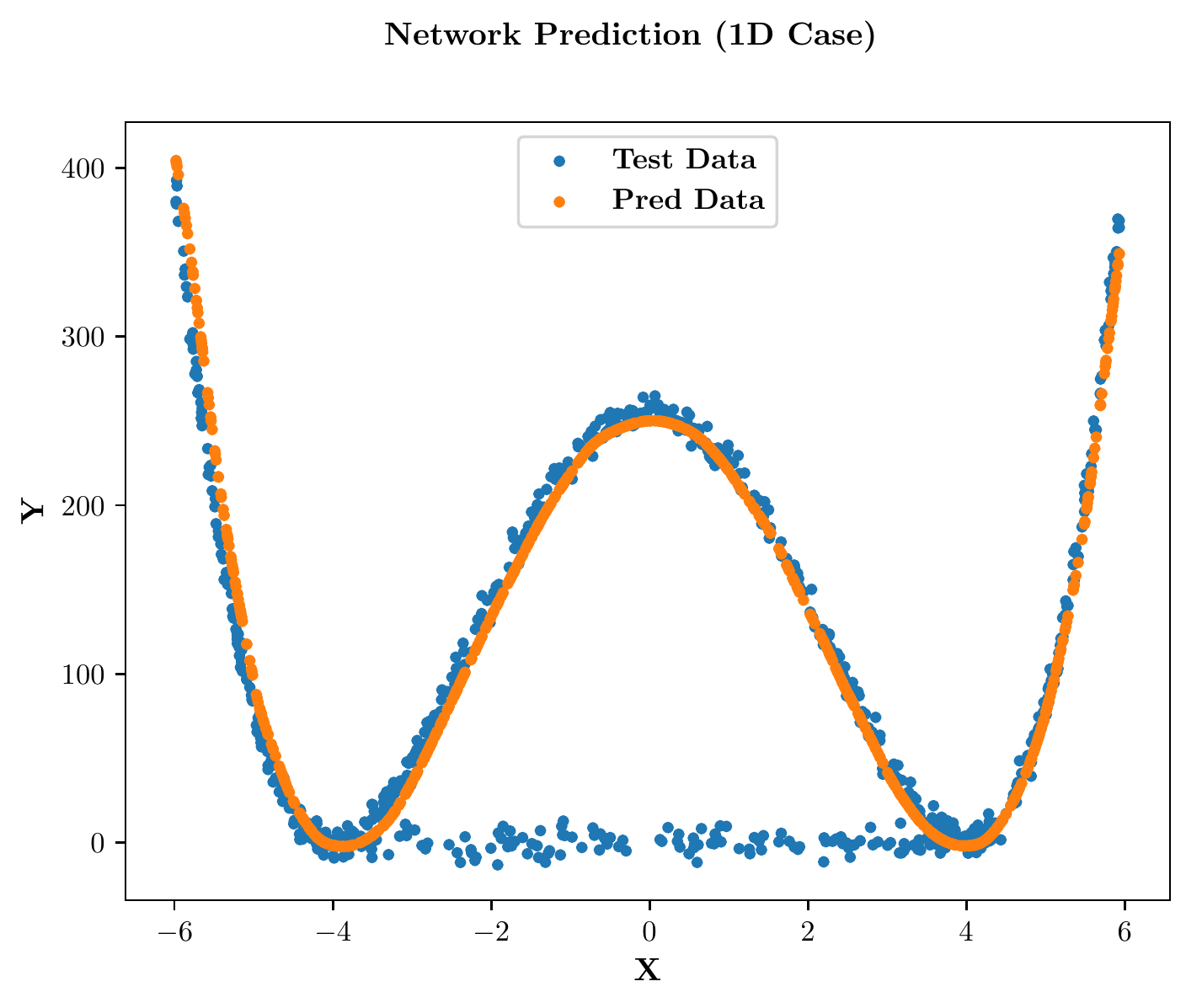} \\
    \small (a) $60\%$ or more of function $f_1$
  \end{tabular} \qquad
  \begin{tabular}[b]{c}
    \includegraphics[width=6cm]{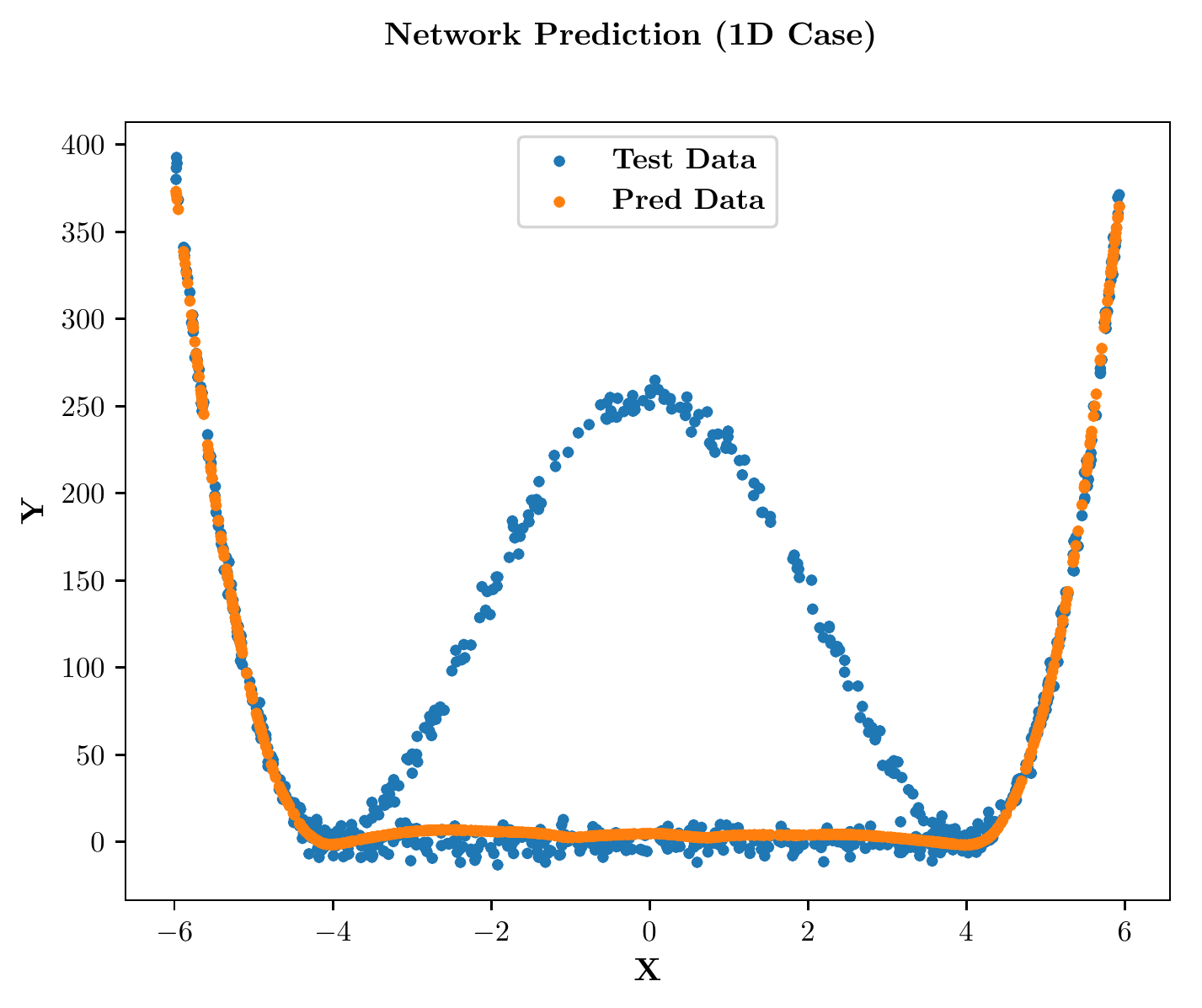} \\
    \small (b) $60\%$ or more of function $f_2$
  \end{tabular}
  \caption{Plot of Test and Predicted Data for the mixed data set for 1 dimensional case. The plots illustrate the behaviour of the network when tested upon 20\% of data set. The network accurately predicts one of the 2 functions accurately based on the fraction of data considered.}
  \label{fig:3.3}
\end{figure}

In this section, the behaviour of our network based on the chosen network architecture is discussed. As discussed earlier the network was trained with a different fraction of the two chosen functions and then tested on the test data. It can be seen that the network predicted one of the two chosen function with high accuracy and not the mean of the two functions. The network predicted the function $f_1$, when $60\%$ or more of the function $f_1$ was chosen in the combined data set and it predicted the function $f_2$ otherwise, as shown in the figure~\ref{fig:3.3}.
It is clear from the figure ~\ref{fig:3.3} that the logcosh loss function learns the bigger cluster of data, unlike the mean square error which would learn the mean of the two functions or the absolute error which would oscillate between the two chosen functions.

\section{2 dimensional case}
We now choose a 2-dimensional case based on the concept discussed in section~\ref{section:1}. Similar to the 1D case, two 2 dimensional single-valued functions were combined in different fractions to form the multi-valued data set, trained by the neural network and finally, the behaviour of our network based on these data sets were analysed and concluded.

\subsection{Network Architecture}
\label{subsection:Network2D}
The network architecture for the 2-dimensional case was very similar to that of the 1-dimensional case as discussed in section \ref{subsection:Network1D}. 160000 data points were chosen. Therefore, the network was trained with [128000 x 2] training examples and was evaluated on [32000 x 2] test data. Similar to the previous case, a fully connected network with 2 neurons at the input and 50 neurons each for the 4 hidden layers were used. ELU activation function was used for the hidden layers. Logcosh loss function and Adam optimizer were used in the network. As the input size was high, a batch size of 200 was used to train the network.

\subsection{Training Strategy}
Two simple 2D functions were chosen to validate our claim.
\begin{equation}
f_{1}(\vek x, \vek y) = xy(2x+2y)
\end{equation}

\begin{equation}
f_{1}(\vek x, \vek y) = xy(x^2+y^2)
\end{equation}

The two functions were then used on the sigmoid function. The main reason to use the sigmoid function was to keep its range between (0,1).
\begin{equation}
\Hat f_{1}(\vek x, \vek y) = \text{sigmoid}(f_{1}(\vek x, \vek y)) = \frac{1}{1+e^{-f_1(\vek x, \vek y)}}
\end{equation}
\begin{equation}
\Hat f_{2}(\vek x, \vek y) = \text{sigmoid}(f_{2}(\vek x, \vek y)) = \frac{1}{1+e^{-f_2(\vek x, \vek y)}}
\end{equation}
To set up a multi-valued data set we combined both the data sets of the functions $ \Hat f_1$ and  $ \Hat f_2$ in different fractions to form a combined data set as per our requirement. The noise was added to the data set to replicate the real-world scenario. The neural network was trained with this data set and then predicted on the test data which is $20\%$ of the total combined data.
\begin{figure}[ht]
    \centering
    \includegraphics[width=7cm]{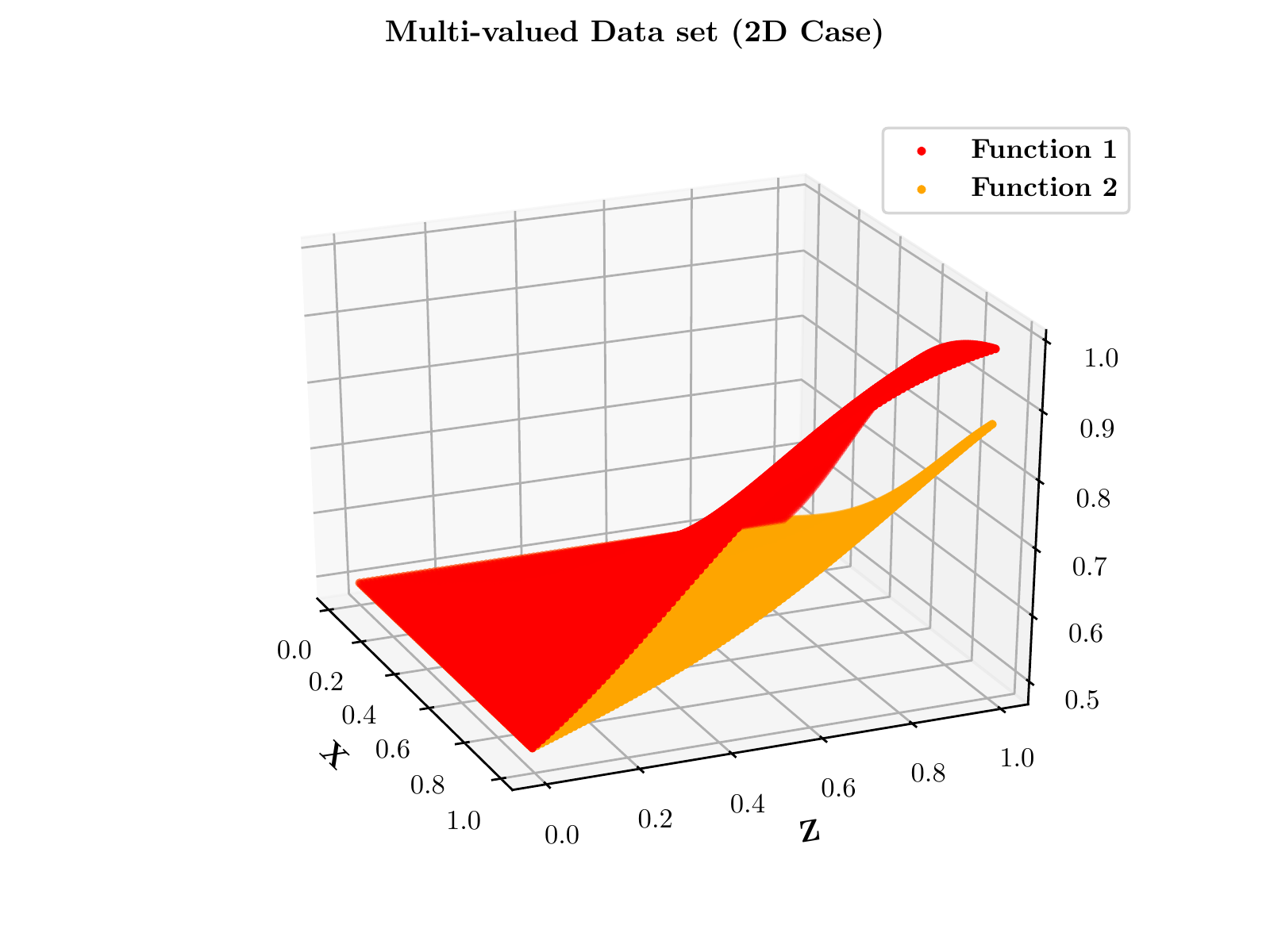}
    \caption{Plot of data set without noise for a 2 dimensional data set}
    \label{fig:3.4}
\end{figure}
Figure~\ref{fig:3.4} shows the plot of the combined data set without noise, where red and orange represent the function $\hat f_1$ and function $\hat f_2$ respectively. As discussed earlier, in this case, for a given $\vek x,\vek y$, we have two distant values $\vek z_1$ and $\vek z_2$ despite being very close to each other. Therefore, the network with logcosh loss function is trained with different fractions of the functions $\hat f_1$ and $\hat f_2$ in an aim to classify the two based on the weight given to the functions.

\subsection{ Network behaviour}
A very noisy data set was used to train the network in which the 2 populations cannot be easily distinguished by visualization. After training the network with a combination of different fractions of the two chosen functions $\hat f_1$ and $\hat f_2$, similar to the 1D case, a clear rule was visible when the logcosh loss function was used. 
\begin{figure}[ht]
  \centering
  \begin{tabular}[b]{c}
    \includegraphics[width=6cm]{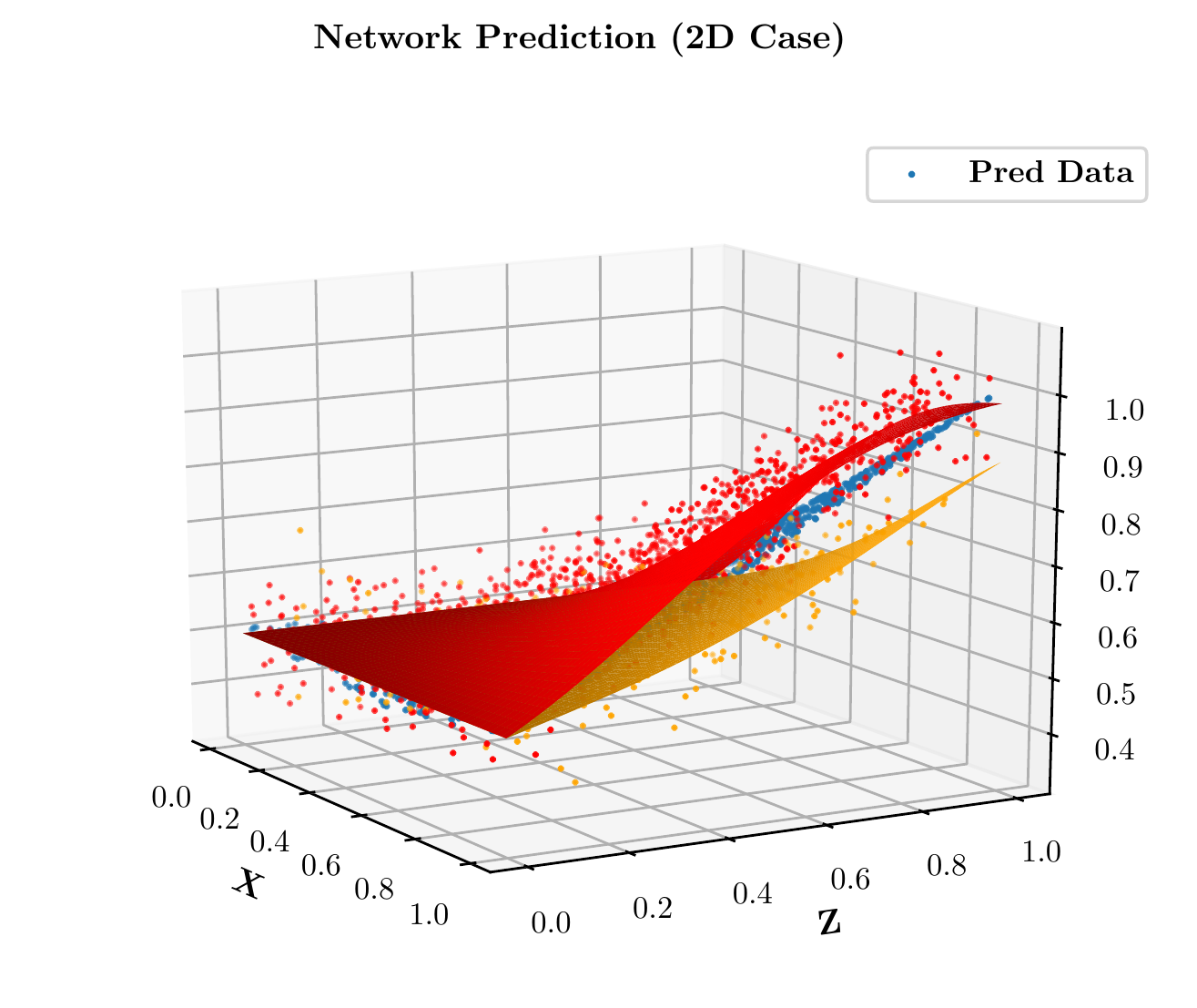} \\
    \small (a) $60\%$ or more of function $\hat f_1$
  \end{tabular} \qquad
  \begin{tabular}[b]{c}
    \includegraphics[width=6cm]{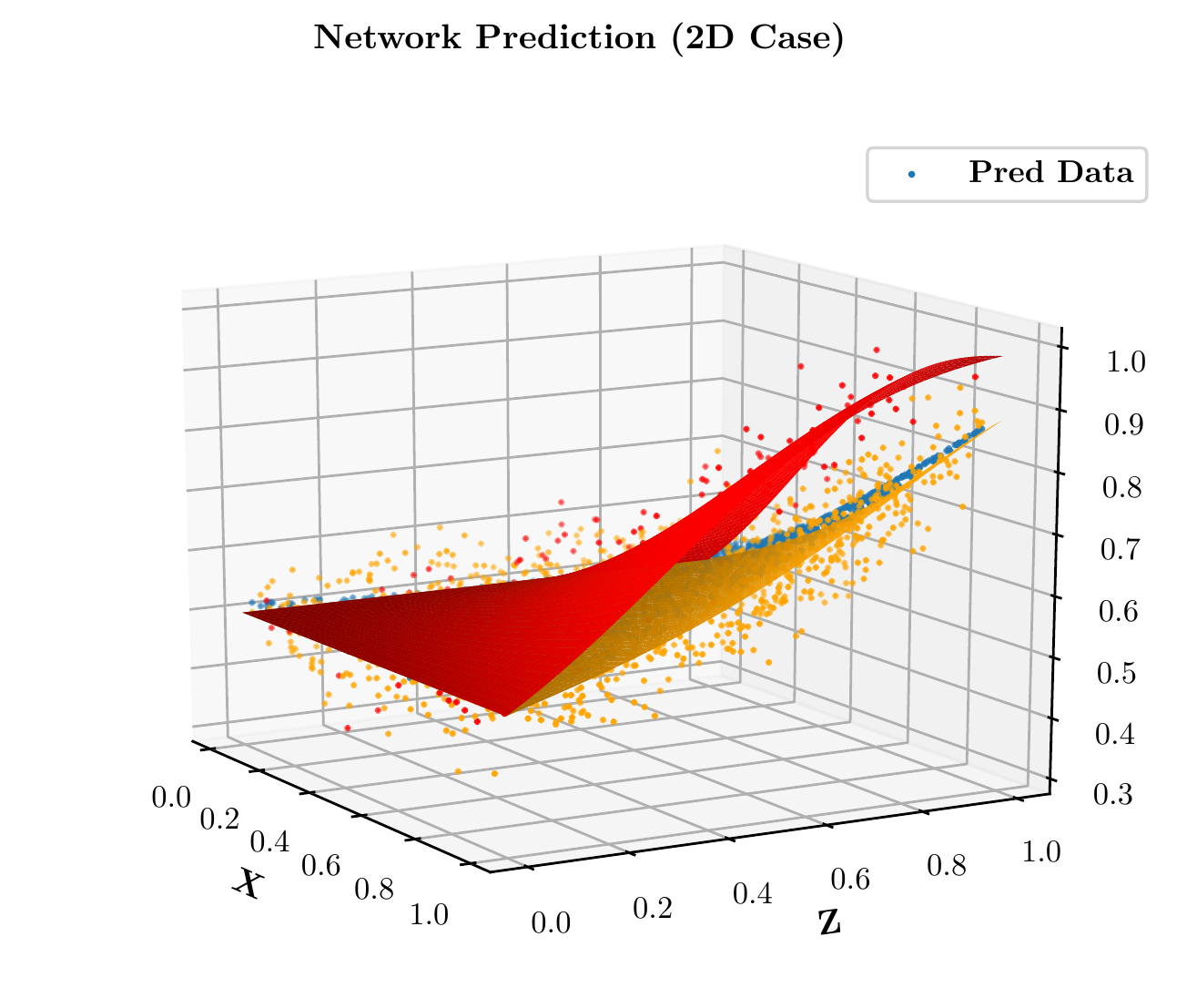} \\
    \small (b) $60\%$ or more of function $\hat f_2$
  \end{tabular}
  \caption{Plot of Test and Predicted Data for the mixed data set for a 2 dimensional data set. The network predicts one of the two functions accurately depending upon the fractional composition of the data set. }
  \label{fig:3.5}
\end{figure}
The network predicted the function $\hat f_1$ when $60\%$ or more of the function $\hat f_1$ was chosen in the fraction of the combined data set and vice versa as shown in the figure~\ref{fig:3.5}. In the plots, the red scatter points represent the function $\Hat{f_1}$ with noise and the red surface plot represents the function $\Hat{f_1}$ without noise. Similarly, for the function $\Hat{f_2}$, orange scatter points and orange surface plot represents the function with and without noise respectively. Finally, the blue scatter points represent the predicted value. The functions were plotted without noise for better visualisation. From figure \ref{fig:3.5}, it is clear that the network learnt one of the 2 functions accurately without being influenced by noise. It can be therefore confirmed that the neural network predicts the bigger of the two clusters when logcosh loss function is used.

\section{ANN Regression quality as a classification criterion}
\label{secion:regclass}
Based on the network behaviour, we claim that a network with logcosh loss function can be used to classify the data when clusters of data exist. It can be concluded that in case of clustered data, an artificial neural network with logcosh learns the bigger cluster rather than the mean of the two. Even more so, the ANN when used for regression of a set-valued function, will learn a value close to one of the choices, in other words, one branch of the set-valued function, while a mean-square-error NN will learn the value in between \parencite{classification}. Based on the above result we have a neural network that not only helps in classifying the data based on the invisible features but also predicts the majority cluster with high accuracy. In the real world scenario, the unavailability of enough parameters to build the regression model is always a major problem and therefore it becomes increasingly difficult to represent the model based on the available limited data. Using this theory, we can classify the clusters of data based on an invisible feature which is not available. It can be also used to check if there are enough features to represent the model. In other words, we can confirm if a feature is required to represent the model based on our theory. A simple example is the tuberculosis vaccine which was compulsory for the citizens of East Germany but not of West Germany before unification. A network can be built without using this information i.e. as an invisible feature, to check the effect of tuberculosis vaccine on the COVID-19. If this parameter is indeed an invisible feature, the network then classifies the data set into two classes: Eastern and Western German districts which are discussed in detail in the upcoming chapters.

\chapter{Experimental Setup / Methodology} 
\label{Chapter4} 

In this chapter, we set up a model based on certain relevant parameters to predict the total number of coronavirus cases, the logarithm of the total cases, total Deaths, active cases and few other parameters and then attempt to classify the data based on an invisible feature. We use the artificial neural network regression quantity as a classification criterion as discussed in the section~\ref{secion:regclass}. \emph{Invisible feature} here represents a feature which is not exclusively provided to the model during training. This feature must be a parameter that classifies the data into clusters. When this feature is included in the model, the input feature set will be able to completely represent a model and give accurate predictions. The model is trained without including this important parameter which classifies the data. This parameter becomes invisible/hidden to the model and hence we call it an invisible feature. When clusters of data exist, the model tries to find rules to classify the data based on an invisible feature as discussed in the previous chapter. In our research, we try to classify the data by using such an invisible feature.
\newline
\par
It was speculated that the tuberculosis vaccine plays a vital role by reducing the spread of the coronavirus. We validate our theory by considering the tuberculosis BCG vaccine as an invisible feature. Since we know as a fact that the Eastern German districts were vaccinated before unification, the network must be able to find two clusters of data, namely — East and West German districts, based on this invisible feature, if the speculation of the vaccine is true. Based on the network behaviour, we then can decide if the tuberculosis vaccine plays a role in reducing the spread of the novel coronavirus. Though a large amount of researchers tries to link between the COVID-19 and the tuberculosis BCG vaccine, no work has been done to find the link using the vaccine as an invisible feature for the neural network. Moreover, the research works on the district-based data of Germany from the beginning of the pandemic and successfully provides a network to predict the number of cases, deaths and other useful data based on certain highly correlated input parameters which are discussed later. 

\section{Why BCG Vaccine}
Bacillus Calmette–Guérin (BCG) vaccine is a vaccine which is used against the Tuberculosis (TB) disease. Tuberculosis is an infectious disease usually caused by the bacteria Mycobacterium tuberculosis (MTB) which generally affects the lungs and also other parts of the body \parencite{WHOTB}. In places where TB is common, one dose of the vaccine is recommended to be given to the babies soon after their birth. In areas where tuberculosis is not common, only children at high risk are typically immunized, where suspected cases of tuberculosis are individually tested for and treated. The protection rate of the vaccine varies and protects for up to 20 years \parencite{WHOBCG}. In recent years, a new concept of trained immunity has emerged, which has helped improve the understanding of the role of BCG vaccination in shaping the innate immune memory response. Innate immune cells, such as macrophages, monocytes, or NK cells, can change their epigenome after exposure to infection, vaccination, or other stressors, which modifies their expression profile and cell physiology \parencite{Neteaaaf1098}. This suggested that the vaccine provides a long term or sometimes lifetime immunity. It has also been studied that the vaccine has shown to provide protection to a wide range of viral infections, mainly respiratory diseases \parencite[page 335]{covidBCG}.
\newline
\par
\parencite{Escobar202008410} suggested that a strong correlation exists between the BCG index, an estimation of the degree of universal BCG vaccination deployment in a country and COVID-19 mortality rate in different socially similar European countries. The coarse study indicated a negative correlation between the BCG index and COVID-19 mortality rate, i.e. every 10\% increase in the BCG index was associated with a 10.4\% reduction in COVID-19 mortality. However, the research was made on coarse data and was not able to provide solid proof for the research. The article suggested that the consistent association between reduced severity of COVID-19 and BCG vaccination observed in the epidemiological explorations is remarkable, but not sufficient to establish causality between BCG vaccination and protection from severe COVID-19. \parencite{covidBCG} also tried to associate between the BCG induced trained immunity and its effect on the coronavirus and concluded that though the data suggests a correlation, the study doesn't provide proof for the same.

\subsection{BCG Vaccination in East and West Germany}
This section familiarizes the BCG vaccination policy that existed in the politically divided Germany (1949-1989) before the reunification. East Germany ruled by the communist government had a strong BCG vaccination policy which made it mandatory for its citizens to take one dose of the vaccination and therefore 99.8\% of the newborns were vaccinated by day 3. However, the voluntary BCG vaccination in the west counterpart was far less common due to low evidence of the disease after the world war. In the early years, only 7–20\% of all newborns were vaccinated in Western Germany, with almost complete cessation of vaccination between 1975 and 1977. Therefore, the comparison of mortality of COVID-19 among the two parts of Germany would be strongly informative as suggested by \parencite{covidEASTWEST}. The article discussed the work of \parencite{Miller2020} and tried to compare it with the two parts of Germany.
\\

In the upcoming sections, we discuss the formulation of the data sets and the network to predict data related to COVID-19 for each district. Based on our theory, we try to see if the network classifies West and East German districts i.e. finds if there are two clusters of data present based on the invisible feature.

\section{Data Set}
In this section, we discuss the detailed formulation of the data set which was used to train our neural network. Most of the scientific research is based on gathering and analysis of relevant data. Feature selection is the process of identifying and selecting a subset of input variables that are most relevant to the target variable. The simplest case of feature selection is numerical input and target for regression predictive modelling because the strength of the relationship between the two can be easily calculated by using correlation.

\subsection{Factors influencing COVID-19}
Though numerous factors affect coronavirus cases or mortality rates of the virus, we use the available demographic data to form our data set. One of the most important factors that affect the COVID-19 transmission and fatality rate is the \emph{age structure} of the country. \parencite{Dowd9696} examined the role of age structure in the spread and deaths caused by the virus. It illustrated how the pandemic effects cities/countries with similar population sizes but different age structures. It was noted that the case fatality rate (CFR) was much lower in the lower age group than the older group. \parencite{AgegroupGermany}'s work analysed the policies implemented by the German government and conducted an empirical analysis to access the factor which plays a major role in countries' fight to defeat the virus. The results suggested that \emph{population density} and \emph{disposable income} played an important role in determining the number of cases and death rates. The study showed a positive correlation between the number of cases and the population density, whereas there existed a negative correlation between the disposable income and the total cases, i.e. higher income resulted in lower cases/death rates.
\\

Apart from these, there are many others factors like Healthcare Expenditure, availability of beds, immunity, Diabetes, social distancing norms, lockdown strategy and most importantly human behaviour towards the virus which affects the number of cases in a given region. Since our study is based on different German districts of a single country, most of the other factors like healthcare expenditure, immunity, lockdown strategy, governance etc can be considered constant throughout the country and some other minor factors can be ignored for simplicity.

\subsection{Formulation}
Many limitations exist while forming the data set. Therefore, only a few of the most important parameters were selected. As we selected districts of one single country for our study, most of the features remain constant and are hence ignored. The data set was formed based on the following assumptions: 
\begin{itemize}
\item Many features related to heath-care, governance, lockdown strategy etc were considered constant thought the country.
\item Human behaviour towards the virus varies and cannot be parameterized and therefore ignored.
\item Climate or other environmental factors were considered constant within the country.
\item Travel history is sometimes an important factor but it cannot be easily tracked and therefore ignored.
\end{itemize}

The information on the COVID-19 for each district in Germany was received from \parencite{coviddata}. The data contained district-based information of population, the total number of cases, cases per million, deaths, active cases and other important data.  Most of the other data were taken from the Federal Statistical Office which is a German federal authority in the division of the Federal Ministry of the Interior, which collects and analyzes statistical information on the economy, society and the environment. \parencite{area} provided the area in square kilometer of each district in Germany. \parencite{agestructure} provided the age structure of different German districts. The data set tabulated the total population and the population based on different age groups for different districts. \parencite{income} sorted the districts of the Federal Republic of Germany according to their disposable household income per inhabitant in euros, based on the data from the Federal Statistical Office. The income per capita is less distorted than the gross domestic product per capita and is therefore a better benchmark for wealth. However, the differences in the cost of living between individual regions are not taken into account.
\\
\par
Finally, in the later course of the research, a time series data were used where the total number of cases, deaths and other data was predicted based on the age of the pandemic. The time-series data by \parencite{timeseries} provided detail information on the number of new cases, new deaths, new recovery and also the age group of the infected person on a certain day for each of the German districts.
\\
\par
The raw data mentioned above had a lot of missing values and it required processing. The data was processed and analysed using the data analysis library in python called pandas. After processing the data, a total of 365 data entries were available, i.e. 365 German districts were tabulated based on different parameters. Different features have different relevance in a particular problem. Some are highly related while some are not. Though we have considered the features based on the literature research, it is important to find the correlation of the variables to validate the data set.
\newline
\begin{figure}[ht]
    \centering
    \includegraphics[width=14cm, height=12cm]{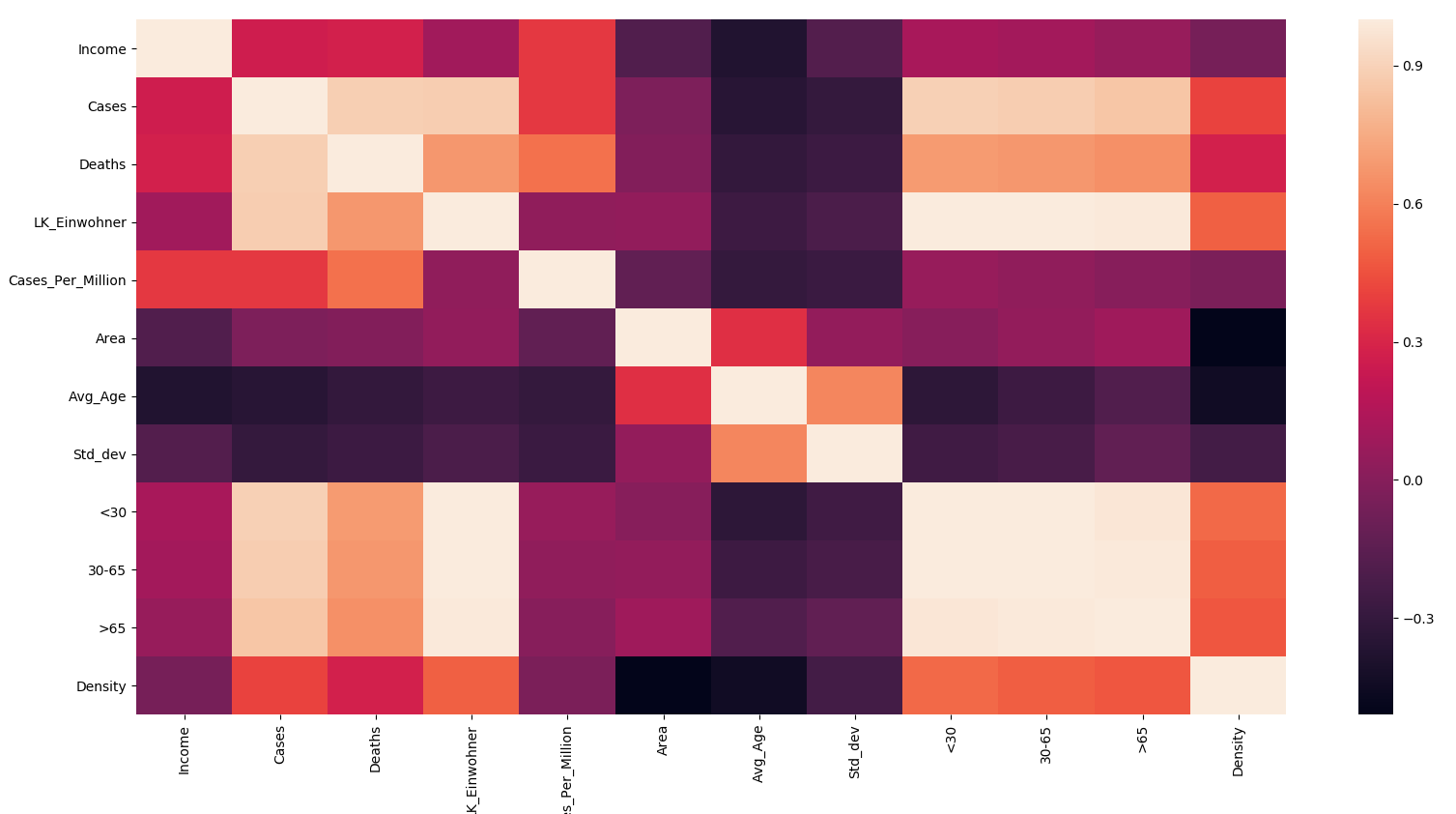}
    \caption{Heat/Correlation Map of the COVID-19 Data Set. The heat map shows correlation of different possible input features among themselves and the correlation of the input features with the target.}
    \label{fig:4.1}
\end{figure}
\par
Figure~\ref{fig:4.1} shows the correlation between different parameters of the data set with each other. The selection of the age group is based on the age group of people who were vaccinated. The group (0-30) includes the population after German unification which was not vaccinated and the age group (30-65) includes population which was possibly vaccinated. This second group plays a major role because it includes population which was vaccinated in Eastern Germany and which was possibly not vaccinated in Western Germany. Therefore, it is an important feature of our classification. From the figure~\ref{fig:4.1} it can be clearly seen that there exists a positive correlation (0.4) between the disposable income and the number of cases/deaths. There also exists a high correlation (0.8-0.9) between the population and the number of cases. A similar correlation exists between the different age groups selected. There is an obvious negative correlation that exists between the area and cases i.e. larger the area, lesser the number of cases. However, in our study, we have chosen density as one of our input feature which is a combination of both population and area (Density = Population / Area) because it is not only a better statistical data but also reduces the number of input features selected. Average age and the standard deviation were initially used as an input feature but later omitted because it was mostly similar for most of the German cities and also made the learning slower.
\\
\par
\begin{figure}[ht]
    \centering
    \includegraphics[width=14cm, height=12cm]{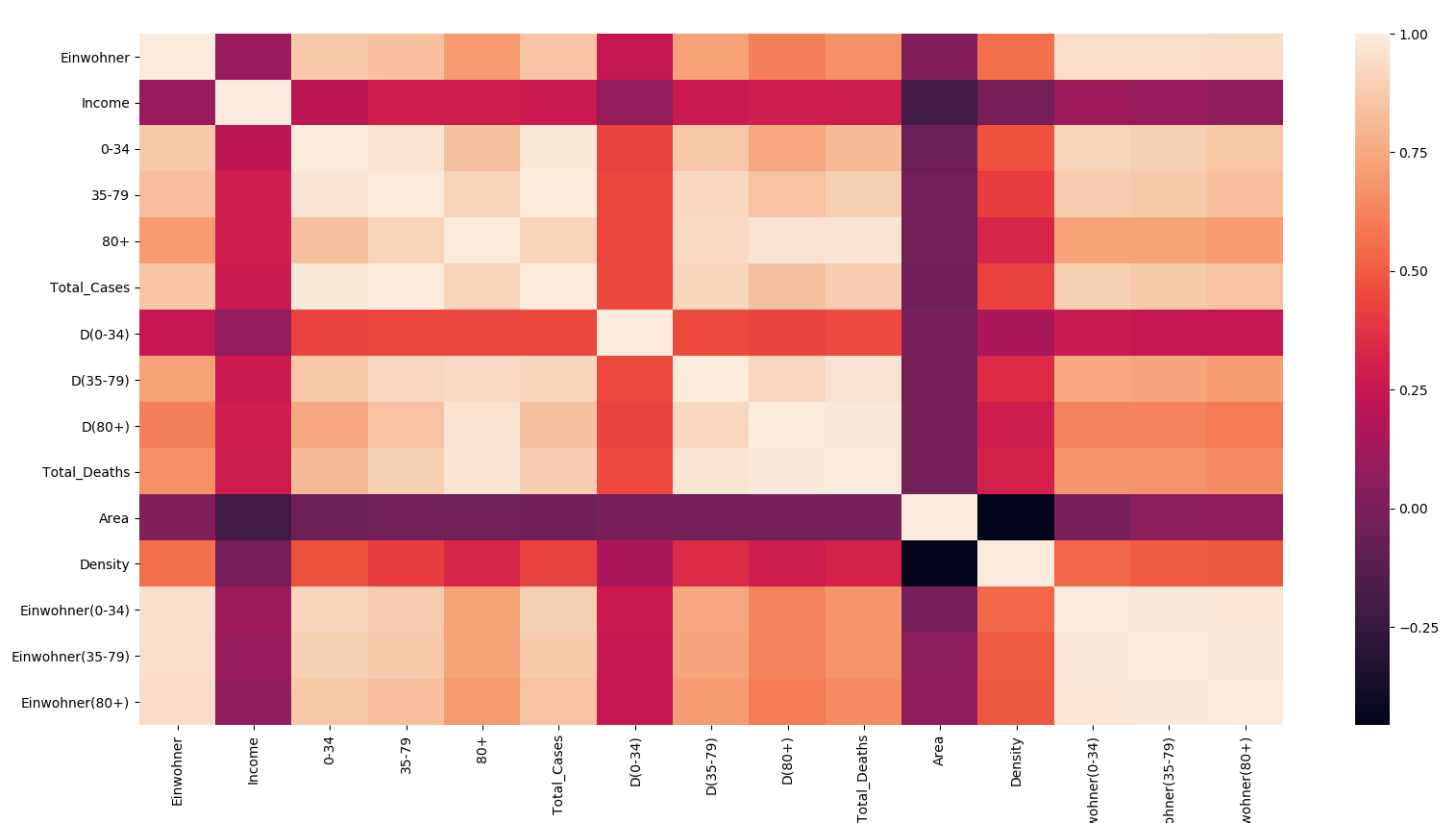}
    \caption{Heat/Correlation Map of the Data Set with age group. The heat map considers the total population, infected population and total deaths in the age groups 0-34, 35-79 and 80+}
    \label{fig:4.2}
\end{figure}

\parencite{timeseries} gave a better insight into our problem by giving information on the age group of the infected and the deceased people. Figure~\ref{fig:4.2} shows the correlation between the parameters after the consideration of the age group information. (0-34), (35-79) and (80+) represents the age group of the infected people, D(0-34), D(35-79) and D(80+) represents the age group of the deceased people and finally the last 3 age groups represents the population in the respective age groups. 
It can be seen that a similar correlation exists in the figure~\ref{fig:4.2} as discussed before. The age group range was changed because of the limitation in the information provided based on the age group of the infected patients by \parencite{timeseries}. 
\\
\par
Also, it must be noted that correlation may not mean causation. Because two variables are related, does not mean that one directly caused the other. Based on our literature survey, correlation map and the preliminary network behaviour, we decided to move forward with the chosen parameters for our network. 
\\
\par
Therefore, in the accumulated data set, the income, density and the age group information were used as the input feature to predict the number of cases or deaths for each district. In the later part of the research, a time series data was used as an input feature which takes into account the age of the pandemic from the day 0, being the first day, to the last day considered for the research. Also, the first-day information was provided to the network in the time series data for better learning. Lastly, the relative population of the age groups were used to predict the relative cases per day. Section~\ref{secion:stratergy} explains in detail the input features, target, training strategy and other important details of the experiment.

\section{Network Architecture}
\label{section:Archi}
In this section, the network architecture of both the accumulated and time series data set is discussed in detail. For the accumulated data, a fully connected neural network with 5 hidden layers was selected. For the output layer, 1 or 3 neurons were used based on the output dimensions. Each of the hidden layers had 100 neurons each and was activated using the ELU activation function. Normal distribution was used to initialize the weights. As one of the fey features of our method is the selection of loss function, logcosh loss function was used with Adam optimizer for minimization. Since our training examples are relatively small, a batch size of 8 was used to train the network with 25 epochs. 
\newline
\par
For the time-series data, a network with more hidden layers was selected due to the large size of the training data. The network consisted of 15 hidden layers with 50 neurons each, which were activated using the ELU activation function. logcosh loss function and Adam optimiser with reducing learn rates of $10^{-3}$, $10^{-4}$ and $10^{-5}$ were used for the optimization. A batch size of 100 was used to fit the model. 15 epochs were used to train the data. As 3 reducing learn rates were used, the model iterates 3 times with 15 cycles through the full training set per iteration.

\section{Learning Strategy}
\label{secion:stratergy}
In this section, we discuss in detail the network selected, input features, target and the training strategy. In the beginning, an accumulated data set was used which included the data of different districts and the total number of cases and deaths till a particular date. For a more focused study, in the later stages, a time series data set was used which included the age of the pandemic.

\subsection{Accumulated Data}
In the initial stage of our research, an accumulated data set was used which provided the COVID-19 details up to a particular day. The data set only provided information about the total number of cases or deaths but did not provide any details on how the pandemic affected each district over time. Our research aimed to observe if there was a clear rule that separates the two clusters if they exist. For doing so, it was important to build a simple network that can be used to predict the required parameters. It is important to note that if the network is able to classify the two former parts of Germany with the limited information provided, it is a candidate to be the hidden feature and thus the criterion that divides the data into two populations. It would suggest that the vaccine information is one of the most important features to represent the data. However, it was unlikely that the network would predict the existence of 2 classes with the information provided by the accumulated data because of the small size of the training set and the unavailability of information about the effect the age of the pandemic towards the cases.

\subsubsection{Data with age group}
The primary network was built with income, population, area, average age and standard deviation of the age as input features and was predicted on the total number of cases and deaths. A simple fully connected neural network of 5 layers with 100 neurons each and batch size of 8 was used with logcosh loss function and Adam optimiser as discussed in section~\ref{section:Archi}. Because of the inability to predict the target accurately, average age and standard deviation were omitted and the different classes of age were considered as described in section \ref{section : avgage}. Instead of considering the population and area separately, the population density was considered, as it not only reduces the number of input feature but also results in better prediction. Moreover, the population information was already provided in the age groups which we have selected. The 3 age groups selected were 0-30, 30-65 and 65+ based on the vaccination information. Since the vaccine was given a few days after the birth, the age group 30-65 and 65+ included the population which was possibly vaccinated; definitely vaccinated in Eastern Germany and mostly not vaccinated in the western counterpart. 
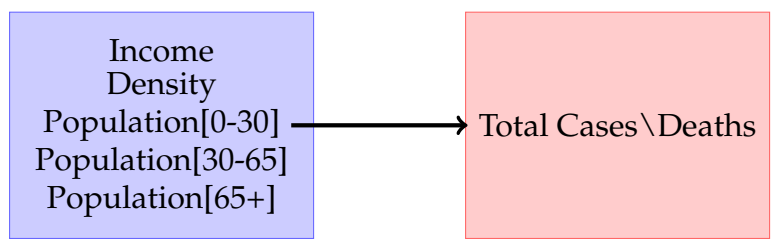
\begin{figure}[ht]
\centering
\begin{tikzpicture}
    \filldraw[fill=blue!20, draw=blue!60] (-5,1.5) rectangle (-1,-1.5);
    \filldraw[fill=red!20, draw=red!60] (1,1.5) rectangle (5,-1.5);
    \node (x1) at (-3,1.0) {Income};
    \node (x2) at (-3,0.5) {Density};
    \node (x3) at (-3,0) {Population[0-30]};
    \node (x4) at (-3,-0.5) {Population[30-65]};
    \node (x5) at (-3,-1.0) {Population[65+]};
    \node (y1) at (3,0) {Total Cases\textbackslash Deaths};

    \draw[->,ultra thick] (x3) -- (y1);
\end{tikzpicture}
    \caption{Mapping diagram of input features and target for the network considering age groups.}
    \label{fig:map_agegroup}
\end{figure}
Therefore, income, density and population in the age groups of 365 districts were taken as the input feature to train the network to predict the total number of cases, total deaths and cases per million of the respective districts as shown in the mapping diagram (Figure \ref{fig:map_agegroup}). 

\subsubsection{Data with age group information of infected patients}
\begin{figure}[ht]
\centering
\begin{tikzpicture}
    \filldraw[fill=blue!20, draw=blue!60] (-5,1.5) rectangle (-1,-1.5);
    \filldraw[fill=red!20, draw=red!60] (1,1.5) rectangle (5,-1.5);
    \node (x1) at (-3,1.0) {Income};
    \node (x2) at (-3,0.5) {Density};
    \node (x3) at (-3,0) {Population[0-34]};
    \node (x4) at (-3,-0.5) {Population[35-79]};
    \node (x5) at (-3,-1.0) {Population[80+]};
    \node (y1) at (3,0.5) {Total Cases[0-34]};
    \node (y2) at (3,0) { Total Cases[35-79] };
    \node (y3) at (3,-0.5) {Total Cases[80+]};

    \draw[->,ultra thick] (x3) -- (y2);
\end{tikzpicture}
    \caption{Mapping diagram of input features and target for the network considering age groups of infected patients. Here population information is used to predict the total cases in the different age groups. }
    \label{fig:map_agegroup_infected1}
\end{figure}
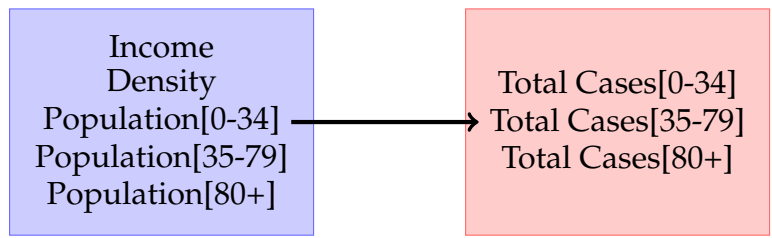

Due to some success in the prediction of the target, a similar feature set and target were used for the neural network with the main difference being the age group information of the infected and deceased patients that were considered for the experiment. A similar network with 5 layers and 100 neurons each in the hidden layer was used with ELU activation function, batch size 8, Adam optimiser and logcosh loss function as explained in the section \ref{section:Archi}. In this strategy, 2 different sets of input data and target were used. Firstly, the income, density and the population in the different age groups were used to predict the target. The network here was build to predict the number of infected patients and deaths in the age group class. Therefore, the network had an input dimension of 5 and output dimensions of 3 i.e. 5 input features were used to predict 3 parameters. The age group class used for the work was 0-34, 35-79 and 80+. The age group classes were changed because of the limitation of information available about the age of the infected patients. The use of the new information provided by \parencite{timeseries} also reduced the total number of entries to 349 because some information of the districts were either not available or considered differently. Therefore, initially, the income, density, population in the group 0-34, 35-79 and 80+ were used to predict the total infected patients in the same age group as shown in the mapping diagram (Figure \ref{fig:map_agegroup_infected1}). 
\\
\par
\begin{figure}[ht]
\centering
\begin{tikzpicture}
    \filldraw[fill=blue!20, draw=blue!60] (-5,1.5) rectangle (-1,-1.5);
    \filldraw[fill=red!20, draw=red!60] (1,1.5) rectangle (5,-1.5);
    \node (x1) at (-3,1.0) {Income};
    \node (x2) at (-3,0.5) {Density};
    \node (x3) at (-3,0) {Total Cases[0-34]};
    \node (x4) at (-3,-0.5) {Total Cases[35-79]};
    \node (x5) at (-3,-1.0) {Total Cases[80+]};
    \node (y1) at (3,0.5) {Total Deaths[0-34]};
    \node (y2) at (3,0) { Total Deaths[35-79] };
    \node (y3) at (3,-0.5) {Total Deaths[80+]};

    \draw[->,ultra thick] (x3) -- (y2);
\end{tikzpicture}
    \caption{Mapping diagram of input features and target for the network considering age groups of infected patients. Here information regarding the total cases is used to predict the total deaths in the different age groups. }
    \label{fig:map_agegroup_infected2}
\end{figure}
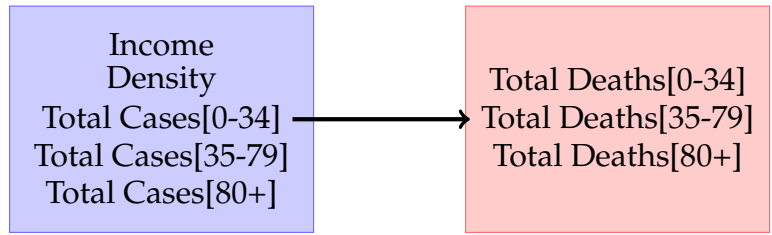

In the second part of the study the income, density and the information regarding the total infected patients were used to predict the total deaths in the same age groups as shown in the mapping diagram (Figure \ref{fig:map_agegroup_infected2}). It must be noted that we aim to not only build an efficient neural network that can predict our target but also to check if two different clusters exist based on our speculation. Therefore, the plots are saved and carefully analyzed.
\subsection{Time Series Data}
After the use of accumulated data set, the information about the age of the pandemic were included. Apart from the already given input features, the age of the pandemic was also used as an input feature. The first day of the pandemic in Germany i.e. 28.01.2020 was considered as day 0 and every consecutive day was considered till a particular day. For the given input features along with day information, the number of cases, active cases and deaths for the corresponding days were predicted. The last day considered for the research was day 129, therefore the size of the data set was 349 x 129. The data set available lacked the exact information of when the patient was tested negative for coronavirus but only had information if a patient was cured or not. Hence, it was important to use a criterion to calculate the active cases which were also used as our target. Since we know that the infected person usually gets cured in 14 days, the same 14 days window was used to form the active cases' data set. The total number of recoveries on a particular day was pushed forward by 14 days and then subtracted from the total number of cases per day, if the patient had not deceased. The predicted and the target data were plotted against the age of pandemic for each district and therefore 349 plots were generated after each run for comparison. The main aim was to build an efficient network to predict the target with high accuracy and check if 2 classes of data exists in our data set.

\subsubsection{Time series data with cumulative cases}
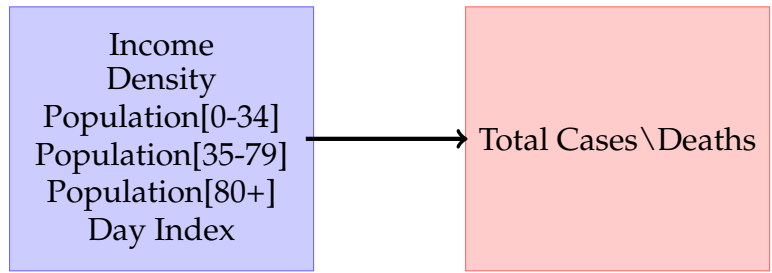
\begin{figure}[ht]
\centering
\begin{tikzpicture}
    \filldraw[fill=blue!20, draw=blue!60] (-5,1.75) rectangle (-1,-1.75);
    \filldraw[fill=red!20, draw=red!60] (1,1.75) rectangle (5,-1.75);
    \node (x1) at (-3,1.25) {Income};
    \node (x2) at (-3,0.75) {Density};
    \node (x3) at (-3,0.25) {Population[0-34]};
    \node (x4) at (-3,0) {\hspace{3.5cm} };
    \node (x5) at (-3,-0.25) {Population[35-79]};
    \node (x6) at (-3,-0.75) {Population[80+]};
    \node (x7) at (-3,-1.25) {Day Index};
    \node (y1) at (3,0) {Total Cases\textbackslash Deaths};

    \draw[->,ultra thick] (x4) -- (y1);
\end{tikzpicture}
    \caption{Mapping diagram of input features and target for the network considering time-series data.}
    \label{fig:map_cumulative}
\end{figure}

The initial network of the time series data was built similar to the accumulated data including the age of the pandemic. In the beginning, different network architectures were tried until the most ideal network was achieved. Finally, a network with 15 hidden layers with 50 neurons each and ELU activation function was built. The network was trained with logcosh loss function and Adam optimiser. Initially, only a single learn rate of $1 \times 10^{-3}$ was used to train the network but later on 3 different reducing learn rates:  $1 \times 10^{-3}$, $1 \times10^{-4}$ and $1 \times 10^{-5}$ were used for the Adam optimiser as it resulted in a considerable reduction of the losses. Here, the density, income, population in the 3 different age groups and the age of the pandemic were used as the input feature to predict mainly the total number of cases / cumulative cases and deaths on the respective days as shown in the mapping diagram (Figure \ref{fig:map_cumulative}). The number of cases was a more noteworthy parameter because some district had very low death rates. The input dimensions were [349 x 129] x 6 to predict [349 x 129] x 1 number of cases and deaths separately. The resultant plots were saved and then examined if the Western and Eastern districts showed different trends. 
\subsubsection{Time series data considering first day information}
Due to the shortcomings of the initial experiment, the first-day information had to be considered as one of the input features. Since we know by a fact that the epidemic in Germany started on different days in different districts, we needed to feed this data to the network as an input, so that the network gets a better idea of the data provided and hence give better generalization. The first day data provided the day on which the pandemic stared based on a simple criterion which helped the network to learn the curve better. After several trials, the following criterion was used: 
\begin{equation}\label{eq:crit}
    \frac{\text{Total Daily Cases}}{\text{Population}} \geq \frac{1}{100000}
\end{equation}
where Total Daily Cases are the total number of cases up to a certain day and Population is the total population of the district. Therefore, the first day of the pandemic is the day when the criterion \ref{eq:crit} is met. This method is used to calculate the first day of all the 349 districts. It can be argued about why the actual start of the pandemic in the respective districts was not used as the first day. It is mainly because the criterion used considers the relative quantity which takes into account the population. The ratio of total daily cases to the population gives a different fraction for different districts and these form a better criterion to select the first day. This criterion becomes mainly important in the districts where the number of cases doesn't rise after the first case was reported. For example, in some districts, the cases remain constant for a certain period after the beginning of the pandemic until the actual behaviour is exhibited. This extra parameter acts as a correction for such cases and helps the network to generalize the behaviour efficiently. Hence, it helps the network identify the curve of the total number of cases on each day precisely.
\\
\par
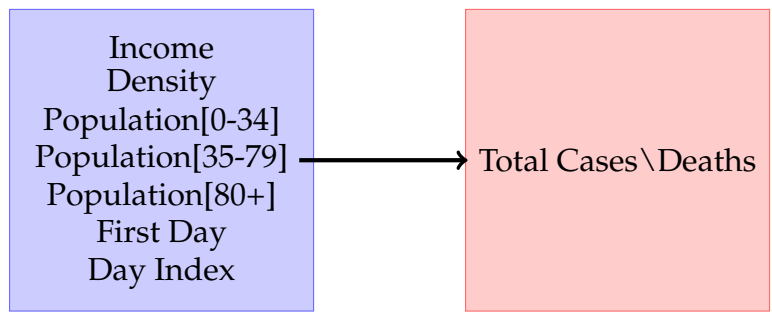
\begin{figure}[ht]
\centering
\begin{tikzpicture}
    \filldraw[fill=blue!20, draw=blue!60] (-5,2) rectangle (-1,-2);
    \filldraw[fill=red!20, draw=red!60] (1,2) rectangle (5,-2);
    \node (x1) at (-3,1.5) {Income};
    \node (x2) at (-3,1) {Density};
    \node (x3) at (-3,0.5) {Population[0-34]};
    \node (x4) at (-3,0) {Population[35-79] };
    \node (x5) at (-3,-0.5) {Population[80+]};
    \node (x6) at (-3,-1) {First Day};
    \node (x7) at (-3,-1.5) {Day Index};
    \node (y1) at (3,0) {Total Cases\textbackslash Deaths};

    \draw[->,ultra thick] (x4) -- (y1);
\end{tikzpicture}
    \caption{Mapping diagram of input features and target for the network considering time-series data with first day information.}
    \label{fig:map_firstday}
\end{figure}

The network used here is very similar to the one discussed in the previous section. A very similar network with 15 hidden layers with 50 neurons each was used with 3 sets of decreasing learn rates as it had shown a further reduction of the loss function. The input features used to train the network were the density, disposable income, the population in the 3 different age groups, the first-day information and finally the day index. The network was used to predict the total number of cases, deaths and also the active cases for each day. Figure \ref{fig:map_firstday} shows the mapping between the input features and target for the network considering the first-day information. The plots of each district with respect to the target and the age of the pandemic were then analysed.
\\
\par
The network was finally used for validation of our theory. Three identical networks were used with different input data. The joint network learns the entire data set including West and East German districts, the west network only learns the western districts and the east networks learns the eastern districts. The west network which was trained with data set of western districts were evaluated on the eastern districts and vice versa. The networks were also evaluated on its own train data to check how well the network had generalized. The joint network was used to evaluate on the entire data set. A total of 277 western and 72 eastern districts exist. If all the networks give similar predictions for all the districts, it justifies that the vaccine doesn't play a major role in preventing the coronavirus. If the vaccine indeed plays a role, then the western network over predicts the eastern districts and the eastern network under predicts the western districts, clearly suggesting the existence of two clusters of data.

\subsubsection{Time series data considering past data}
\begin{figure}[ht]
\centering
\begin{tikzpicture}
    \filldraw[fill=blue!20, draw=blue!60] (-5,2.25) rectangle (-1,-2.25);
    \filldraw[fill=red!20, draw=red!60] (1,2.25) rectangle (5,-2.25);
    \node (x1) at (-3,1.75) {Income};
    \node (x2) at (-3,1.25) {Density};
    \node (x3) at (-3,0.75) {Population[0-34]};
    \node (x4) at (-3,0.25) {Population[35-79] };
    \node (x5) at (-3,0) {\hspace{3.5cm} };
    \node (x6) at (-3,-0.25) {Population[80+]};
    \node (x7) at (-3,-0.75) {First Day};
    \node (x8) at (-3,-1.25) {Average 7-day cases};
    \node (x9) at (-3,-1.75) {Day Index};
    \node (y1) at (3,0) {Active Cases};

    \draw[->,ultra thick] (x5) -- (y1);
\end{tikzpicture}
    \caption{Mapping diagram of input features and target for the network considering time-series data along with past data.}
    \label{fig:map_pastdata}
\end{figure}
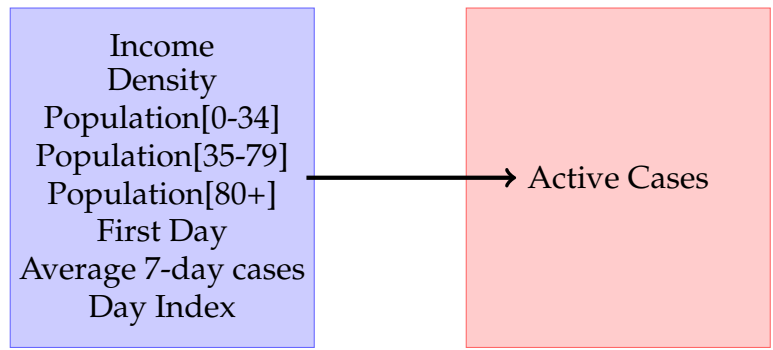

To increase the accuracy of the prediction, past 7-day information of the total number of cases were also given as an input. Initially, the network was put into a setting where past 7 days of the active cases were used to predict the new case i.e. $(Case_{n-7}+Case_{n-6}+...+Case_{n-1})$ were inputted along with other input data to predict the new case $Case_{n}$. Finally, in a slight modification to the above, an average of the past cases was used to predict the target. To predict the $n^{th}$ day case, the average of the number of cases in the last 7 days i.e $(Case_{n-7}+Case_{n-6}+...+Case_{n-1})/7$ was used as one of the input features among the others. Hence, the density, income, population in the three different age groups, first-day information, average of the cases in the last 7 days and the day index were used to train the network to predict the active cases for each district as shown in the mapping diagram (Figure \ref{fig:map_pastdata}). Apart from the above change, a very similar network was used to predict the target. The plots were finally compared to check for clusters.

\subsubsection{Time series data considering log of cases}
Due to the disadvantage of the previous approach as discussed in section \ref{result : pastdata}, the consideration of past information was omitted in further study. The logarithm of cumulative cases per day was considered instead of absolute values. In the log graph, the vertical scale or the y-axis are graduated by the order of magnitude 10, 100 and so on and not by equal intervals. This basically “squashes” the y-axis, so large numbers do not skew the whole graph i.e. it pulls the extreme values to the middle of the distribution. If an epidemic is growing exponentially, it arguably makes more sense to plot it this way because the trend line can keep up with the numbers instead of shooting off into the stratosphere. On the log scale, these exponential increase appears as a straight line and only bends when the growth rate changes.  The slope of a log-scaled graph measures the relative change in the variable of interest. This makes it a powerful tool to assess growth rates, which are particularly meaningful in the context of a global health crisis. Apart from this, it is also contemplated that the network learns the log of data better than exponential data as the log linearizes the exponential curve.
\\
\par
\begin{figure}[ht]
\centering
\begin{tikzpicture}
    \filldraw[fill=blue!20, draw=blue!60] (-5,2) rectangle (-1,-2);
    \filldraw[fill=red!20, draw=red!60] (1,2) rectangle (5,-2);
    \node (x1) at (-3,1.5) {Income};
    \node (x2) at (-3,1) {Density};
    \node (x3) at (-3,0.5) {Population[0-34]};
    \node (x4) at (-3,0) {Population[35-79] };
    \node (x5) at (-3,-0.5) {Population[80+]};
    \node (x6) at (-3,-1) {First Day};
    \node (x7) at (-3,-1.5) {Day Index};
    \node (y1) at (3,0) {Log of cases};

    \draw[->,ultra thick] (x4) -- (y1);
\end{tikzpicture}
    \caption{Mapping diagram of input features and target for the network considering time-series data to predict the log of cases.}
    \label{fig:map_log}
\end{figure}
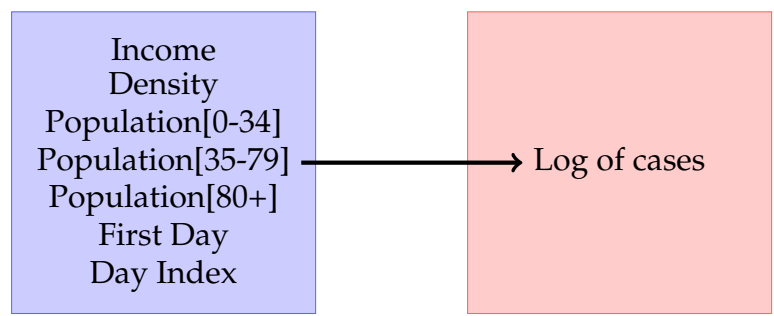

A similar neural network with input features: density, income, the population in the age groups, first day and the day index were used to predict the log of total cumulative cases per day. The natural log of the cumulative cases was taken as the target. Figure \ref{fig:map_log} shows the mapping between the input features and the target for the above case. As we know that the cases were initially 0 up to a certain day for most of the districts and since the log of 0 doesn't exist, they were replaced by 1 and then the natural log operator was used. In this study, the 3 major districts Berlin, Hamburg and Munich were omitted due to the high number of cases. It had been observed that the network generalizes better when these districts were not considered. It is important to note that a lot more parameters affect these big cities and cannot be easily generalized when included in the input feature. In other words, it adds unwanted noise to the network. A network with 15 hidden layers, 50 neurons each with ELU activation function was used with logcosh loss function and Adam optimiser with decreasing learn rates as mentioned before. The plots of the log of cumulative cases vs the age of the pandemic were saved and analysed.
\\
\par
Similar to the strategy discussed in the previous section, 3 identical networks with different input data were used to validate our research. A west network which learns from western districts were used to predict the eastern districts and vice versa. Finally, the combined network was used to predict all the districts. To confirm that 2 sets of clusters exist, a district from the east was chosen and evaluated on the eastern network, the western network and finally on the combined network. A similar strategy was used for the western district. If the classes exist, then the chosen district must over or under predict in the other network (west network over predicts eastern districts and vice versa), predict accurately in its own network and again over, underpredict or predict accurately in the combined network i.e. the districts from dominating class must be predicted accurately in the joint network.
\\
\par
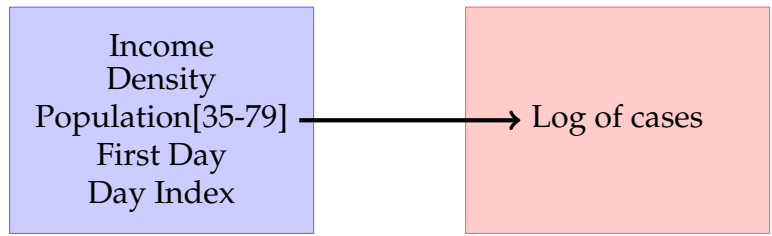
\begin{figure}[ht]
\centering
\begin{tikzpicture}
    \filldraw[fill=blue!20, draw=blue!60] (-5,1.5) rectangle (-1,-1.5);
    \filldraw[fill=red!20, draw=red!60] (1,1.5) rectangle (5,-1.5);
    \node (x1) at (-3,1.0) {Income};
    \node (x2) at (-3,0.5) {Density};
    \node (x3) at (-3,0) {Population[35-79] };
    \node (x4) at (-3,-0.5) {First Day};
    \node (x5) at (-3,-1.0) {Day Index};
    \node (y1) at (3,0) {Log of cases};

    \draw[->,ultra thick] (x3) -- (y1);
\end{tikzpicture}
    \caption{Mapping diagram of input features and target for the network considering only the mid-age group for the time-series data to predict the log of cases.}
    \label{fig:map_log2}
\end{figure}

Finally, for the last variant, the same strategy was followed without considering all the population groups but only the mid-age group of interest: 30-79. The target data, which is the log of the total cumulative cases were only considered for the chosen age group. The start of the pandemic for the chosen age group was 02.02.2020 and therefore the last day considered was day 125. The input dimension of the network reduced by 2 as the other two age groups were not considered for our study. Figure \ref{fig:map_log2} shows the mapping between input features and target for the case considering the mid-age group. Apart from the above difference, a very similar approach was followed and the results were examined.

\subsubsection{Time series data considering relative proportions}
In the final strategy, a more generalized approach was used to represent the data. Till this point, we considered the absolute population in the different age group but in the final network, we considered the relative proportion of the population in each of the age group which was achieved by dividing the population into each of the age groups by the total population of the district. In the previous strategies since we considered the absolute population in each age group, the network learnt the absolute number in each of the age groups to predict the target. Since the absolute number does not give the distribution of people in the age group, the network finds it difficult to find a rule with the absolute population. Once we consider the relative population or the population proportion in each age group, the distribution of population can be explained better. Also, since the sum of the 3 proportions of the age group is 1, one of the age group was omitted from the feature space. Since we considered the relative proportion of people in the age groups, we also needed to select the network target as a relative proportion. Therefore, the new target of the network was the relative number of cases. In this context, the relative number of cases refers to the cumulative cases divided by the total population of the district. The relative number of cases goes maximum up to $0.1-0.2\%$.
\newline
\par
\begin{figure}[ht]
\centering
\begin{tikzpicture}
    \filldraw[fill=blue!20, draw=blue!60] (-7,1.75) rectangle (-1,-1.75);
    \filldraw[fill=red!20, draw=red!60] (1,1.75) rectangle (7,-1.75);
    \node (x1) at (-4,1.25) {Income};
    \node (x2) at (-4,0.75) {Density};
    \node (x3) at (-4,0.25) {Relative Population[0-34]};
    \node (x4) at (-4,0) {\hspace{5cm} };
    \node (x5) at (-4,-0.25) {Relative Population[35-79]};
    \node (x6) at (-4,-0.75) {First Day};
    \node (x7) at (-4,-1.25) {Day Index};
    \node (y1) at (4,0) {Log of relative cases};

    \draw[->,ultra thick] (x4) -- (y1);
\end{tikzpicture}
    \caption{Mapping diagram of input features and target for the network considering relative proportions for time-series data.}
    \label{fig:map_relative1}
\end{figure}
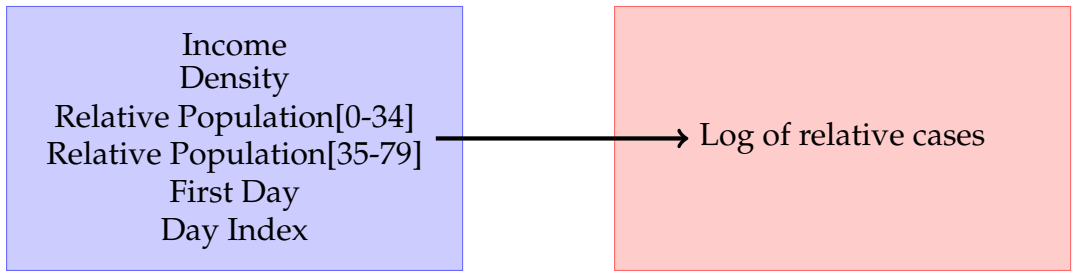

The density, disposable income, relative proportion of people in the age groups 0-34 and 35-79, first-day information and the day index were used to predict the log of the relative number of cases as shown in the mapping diagram (Figure \ref{fig:map_relative1}). Due to the advantages of using the log as the target as seen in section \ref{section:resutlog}, the network was used to predict the log of the relative cases. Similar to the previous section, as the relative number of cases remain 0 till the beginning of the pandemic for each district, the relative number of cases were approximated to the relative cases on the first day, i.e. all the 0 entries were replaced by the relative cases of the first day. The plots for all the districts were finally saved and then analysed. Again similar to the previous case, 3 networks namely the joint network, west network and east network were used to validate our study.
\\
\par
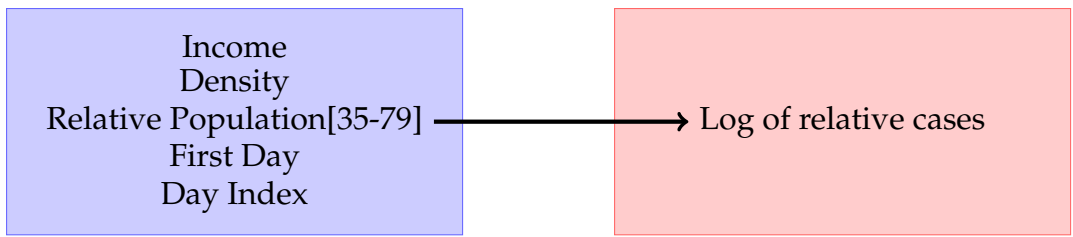
\begin{figure}[ht]
\centering
\begin{tikzpicture}
    \filldraw[fill=blue!20, draw=blue!60] (-7,1.5) rectangle (-1,-1.5);
    \filldraw[fill=red!20, draw=red!60] (1,1.5) rectangle (7,-1.5);
    \node (x1) at (-4,1) {Income};
    \node (x2) at (-4,0.5) {Density};
    \node (x3) at (-4,-0) {Relative Population[35-79]};
    \node (x4) at (-4,-0.5) {First Day};
    \node (x5) at (-4,-1) {Day Index};
    \node (y1) at (4,0) {Log of relative cases};

    \draw[->,ultra thick] (x3) -- (y1);
\end{tikzpicture}
    \caption{Mapping diagram of input features and target for the network considering the relative proportion for the mid-age group for time-series data.}
    \label{fig:map_relative2}
\end{figure}

In the final experiment, only the relative population in the required age group i.e. age group of 35-79 was considered along with the other data as the input and the log of the relative number of cases in the selected age group was used as the target as shown in the mapping diagram (Figure \ref{fig:map_relative2}). A similar strategy of using 3 networks was used to validate our study. Few districts from West and East Germany were selected and evaluated with all the 3 networks. Based on the detailed study of the plots, the final decision was made.

\chapter{Results and Discussion} 
\label{section : Result}

In the previous chapter, we discussed how we set up our model to predict the parameters related to COVID-19 to check if two classes of data exist. For doing so, we built a fully connected neural network and followed different strategies to predict the target with high and thereby analysed the results for different cases. In this chapter, we discuss the results of the different strategies used in detail for both accumulated (section \ref{accumulated}) and time-series data (section \ref{timeseries}) and give possible explanations for the same. Only a few important plots are presented here. For more plots refer to Appendix.

\section{Accumulated Data} \label{accumulated}
In this section, all the results of our model using the accumulated data set are explained in detail. The section walks through the different attempts made to predict the required data and hence analyses the results from the plots. The strategy using the accumulated data set gave all the basic information regarding the behaviour of the network based on our constructed data set. Though the strategy didn't give the most promising results, this strategy formed the base of our research and vaguely presented the behaviour of the network. 

\subsection{Data with age group}\label{section : avgage}
\begin{figure}[ht]
  \centering
  \begin{tabular}[b]{c}
    \includegraphics[width=6cm, height=5.5cm]{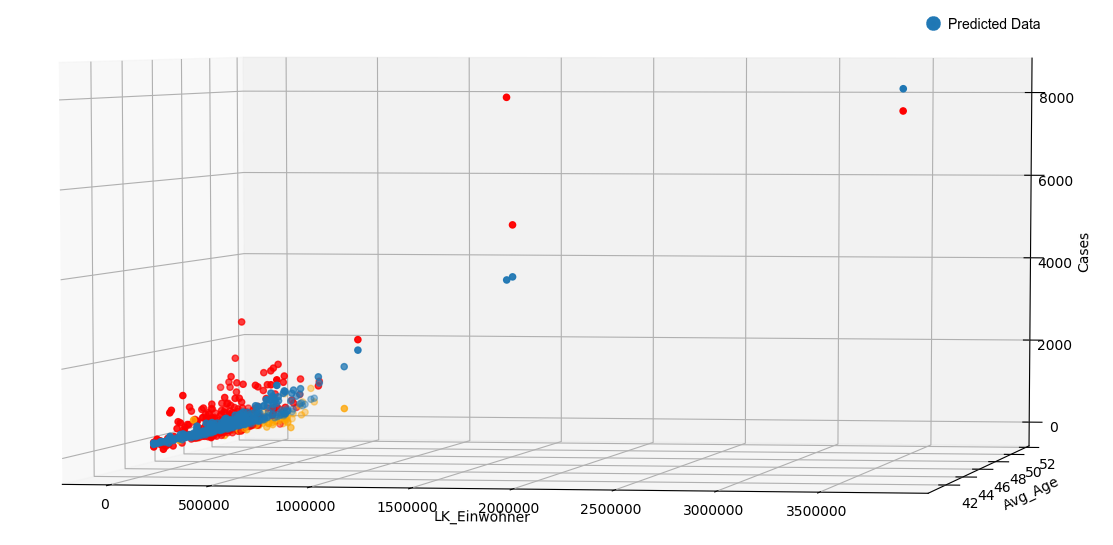} \\
    \small (a) Population vs Average Age \\ vs Cases
  \end{tabular} \qquad
  \begin{tabular}[b]{c}
    \includegraphics[width=6cm, height=5.5cm]{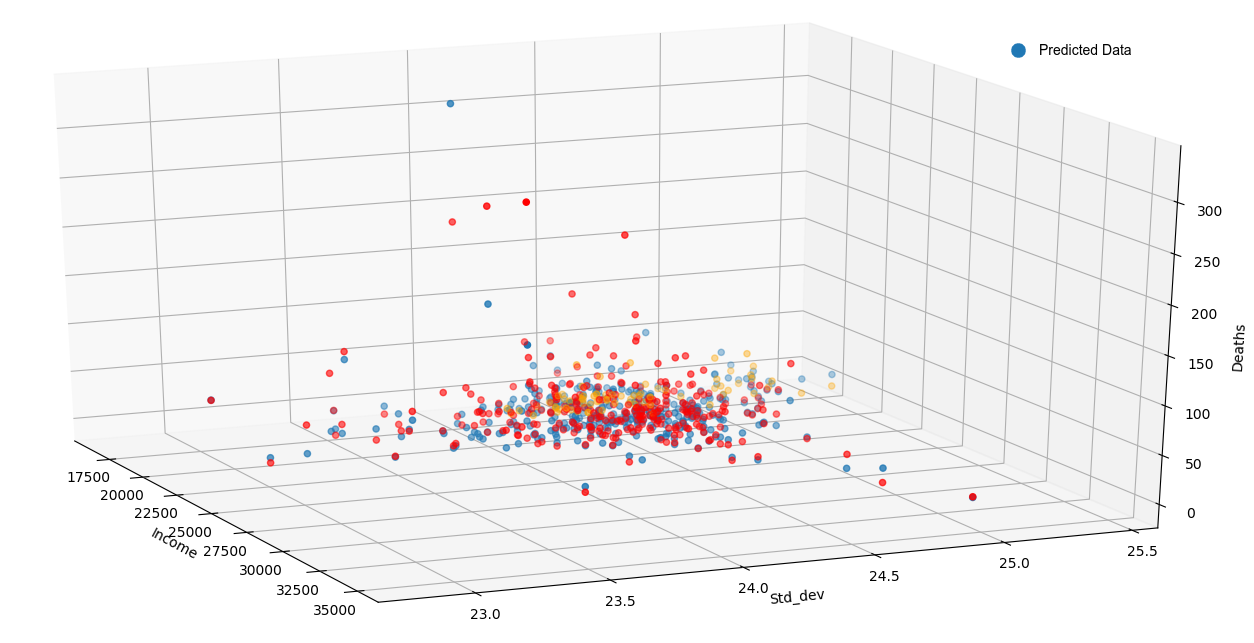} \\
    \small (b) Income vs Standard Deviation \\  vs Death
  \end{tabular}
  \caption{Plot considering average age and standard deviation}
  \label{fig:5.1}
\end{figure}

In the initial network, the average population and standard deviation of the age groups were used along with other input features like Income, population and area to train the network. Figure~\ref{fig:5.1} shows the plot of total cases and total deaths with respect to other variables.
\newline

To easily distinguish the eastern and western districts, the eastern districts were represented in orange and the western in red. The blue points in the scatter plots are the predicted data and rest are the test data. This analogy is followed to represent all the further plots. Though the data can be plotted based on different axes, only the major 2 plots are shown in figure \ref{fig:5.1} as these are sufficient to analyse the details of the plot. The network shows very similar behaviour for prediction of the total cases and total deaths. The plot of total population vs average age vs total cases shows that the network is not completely successful in predicting the total cases based on the input parameters. It can be realized that the network predicted the rise of cases with population linearly while the actual data is much more scattered as seen from the population axis. Similarly, in the second plot between income, standard deviation and total deaths, it is evident that the network is not able to correctly represent the data. The scatter plot shows that the predicted data are distorted with respect to the test data and are unable to predict the cases in the higher range. Hence, it is clear that the selected input data doesn't provide a good prediction. One of the major reasons for this behaviour could be similarities in the average age and standard deviation for the different districts. It can be concluded that the average age and the standard deviation doesn't represent the data accurately and hence can be omitted from the input feature set.
\\
\par
\begin{figure}[ht]
  \centering
  \begin{tabular}[b]{c}
    \includegraphics[width=6cm, height=5.5cm]{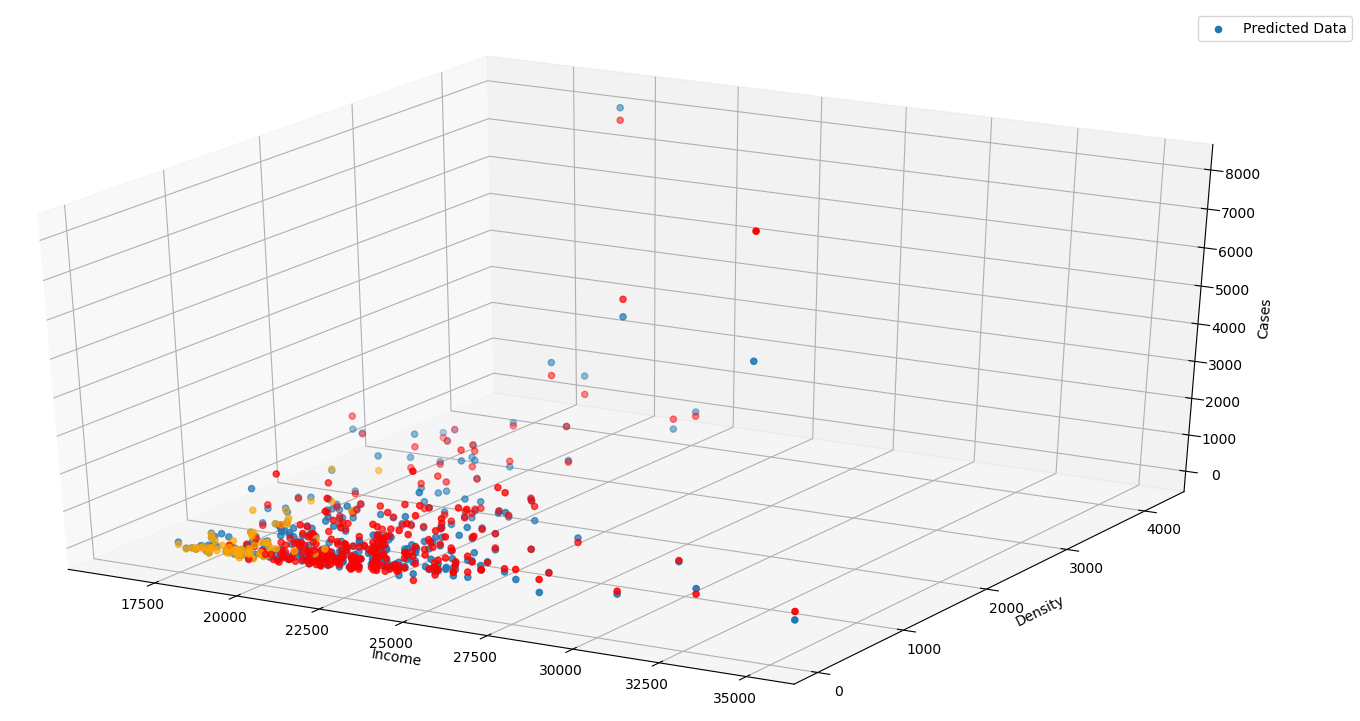} \\
    \small (a) Income vs Density vs Cases
  \end{tabular} \qquad
  \begin{tabular}[b]{c}
    \includegraphics[width=6cm, height=5.5cm]{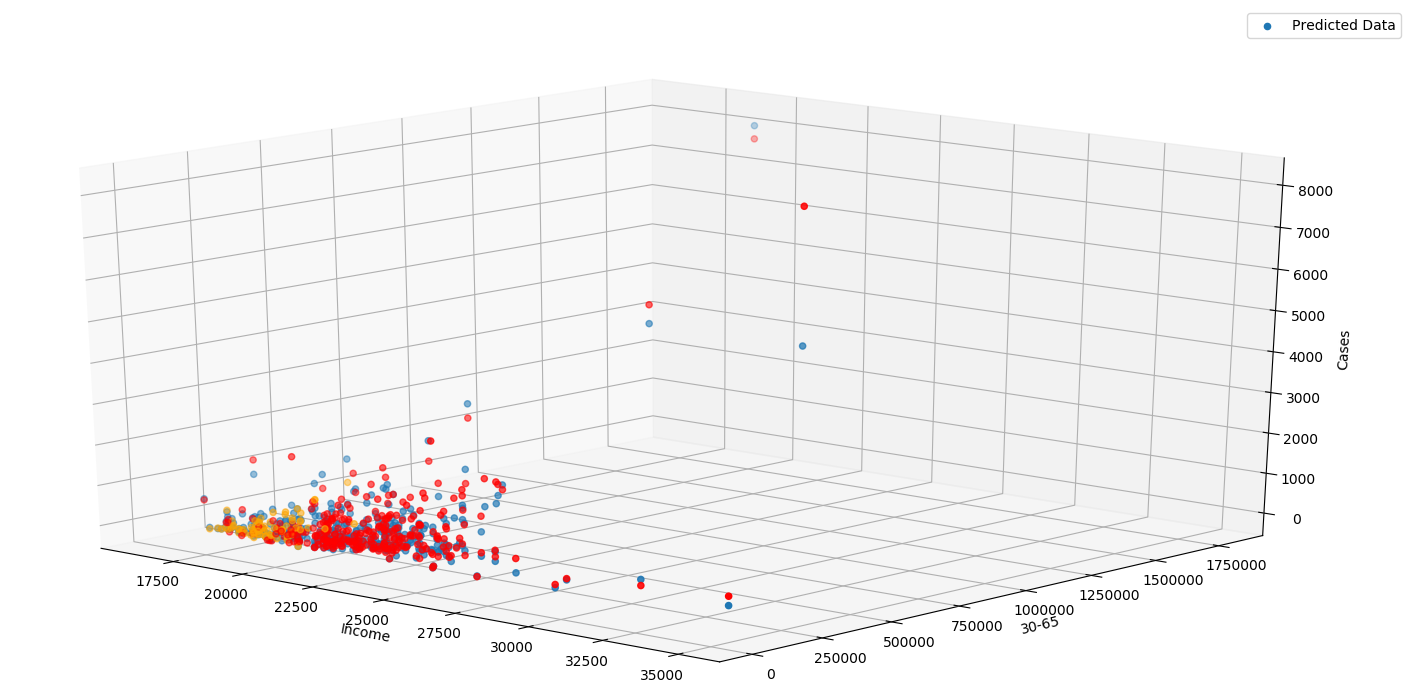} \\
    \small (b) Income vs 30-65 vs Cases
  \end{tabular}
  \begin{tabular}[b]{c}
    \includegraphics[width=6cm, height=5.5cm]{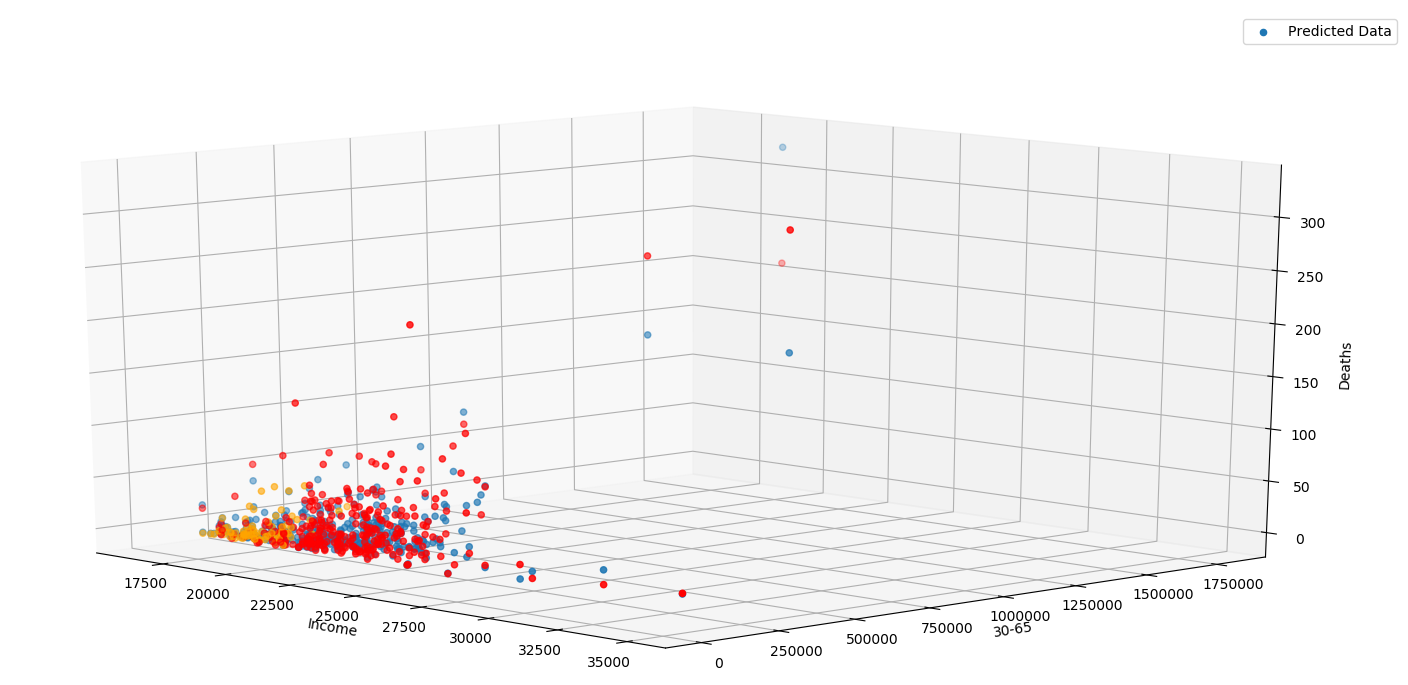} \\
    \small (b) Income vs 30-65 vs Death
  \end{tabular}
  \begin{tabular}[b]{c}
    \includegraphics[width=6cm, height=5.5cm]{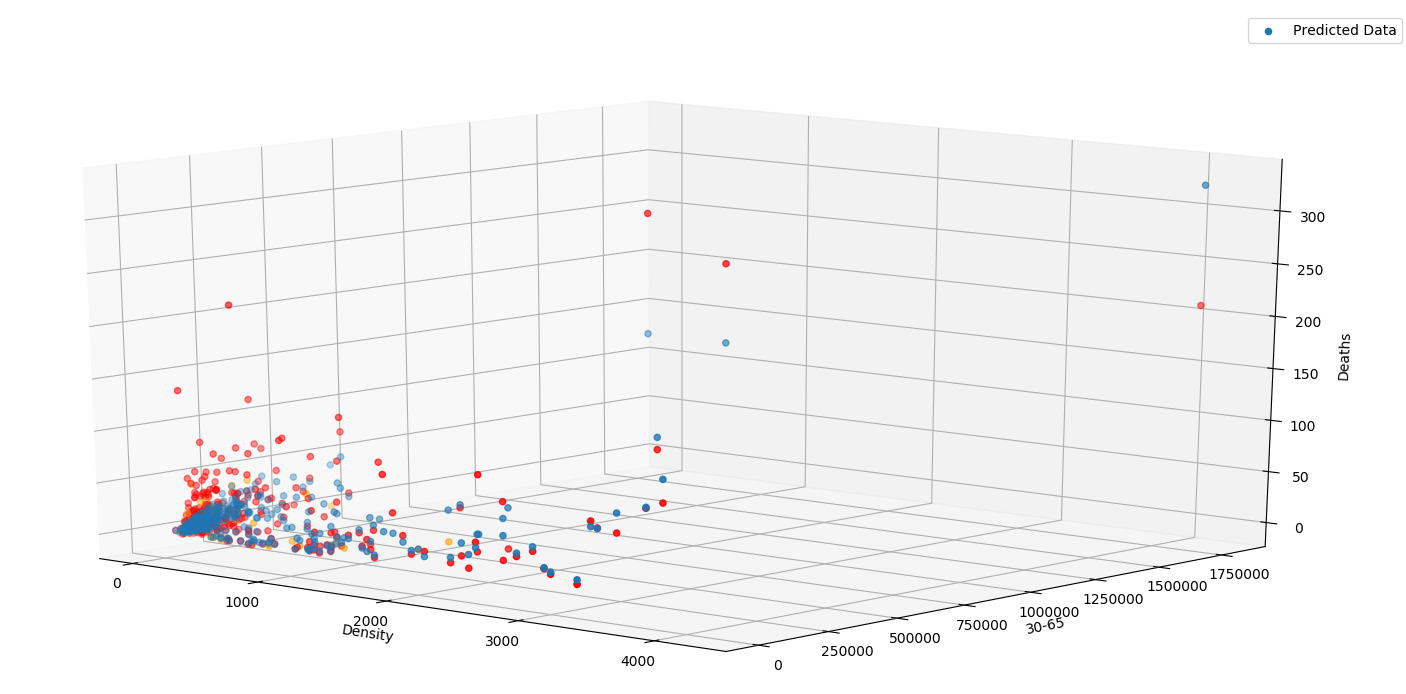} \\
    \small (b) Density vs 30-65 vs Deaths
  \end{tabular}  
  \caption{Plot considering absolute population in the age groups}
  \label{fig:5.2}
\end{figure}

Due to the above disadvantage, the absolute population in the different age groups were selected instead of the omitted parameters. The other major change was that the population and area information was combined and then represented as density which helped in reducing the size of the data set. As we see in figure \ref{fig:5.2}, the predictions had improved from the previous case. The network was able to predict higher points in the plot with some degree of accuracy. However, the network prediction to predict the total number of deaths were not accurate as compared to that of the total cases. In the network to predict deaths, the network was not successful in capturing some extreme points.  As we see in the plots, mainly the mid-age group 30-65 is selected for visualisation due to its relation to the tuberculosis vaccine. Though the prediction has improved, we do not see any clear existence of classes or clusters i.e. the network was unable to distinguish between the western and the eastern districts.

\subsection{Data with age group information of infected patients}\label{section : avgageinfected}
\begin{figure}[ht]
    \centering
    \includegraphics[width=8cm, height=7cm]{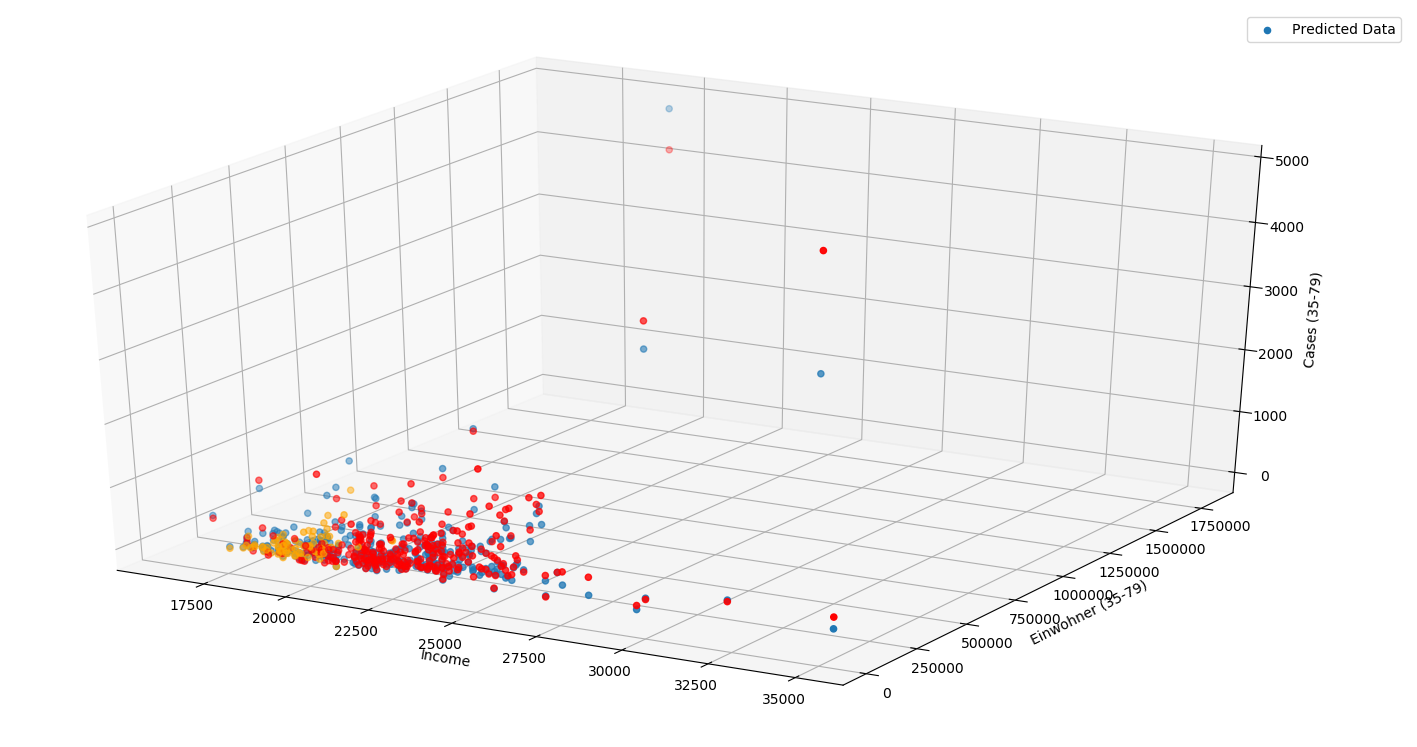}
    \caption{Plot of actual and predicted cumulative cases in age group 35-79 with respect to income and population in the same age group.}
    \label{fig:5.3}
\end{figure}

\begin{figure}[ht]
    \centering
    \includegraphics[width=8cm, height=7cm]{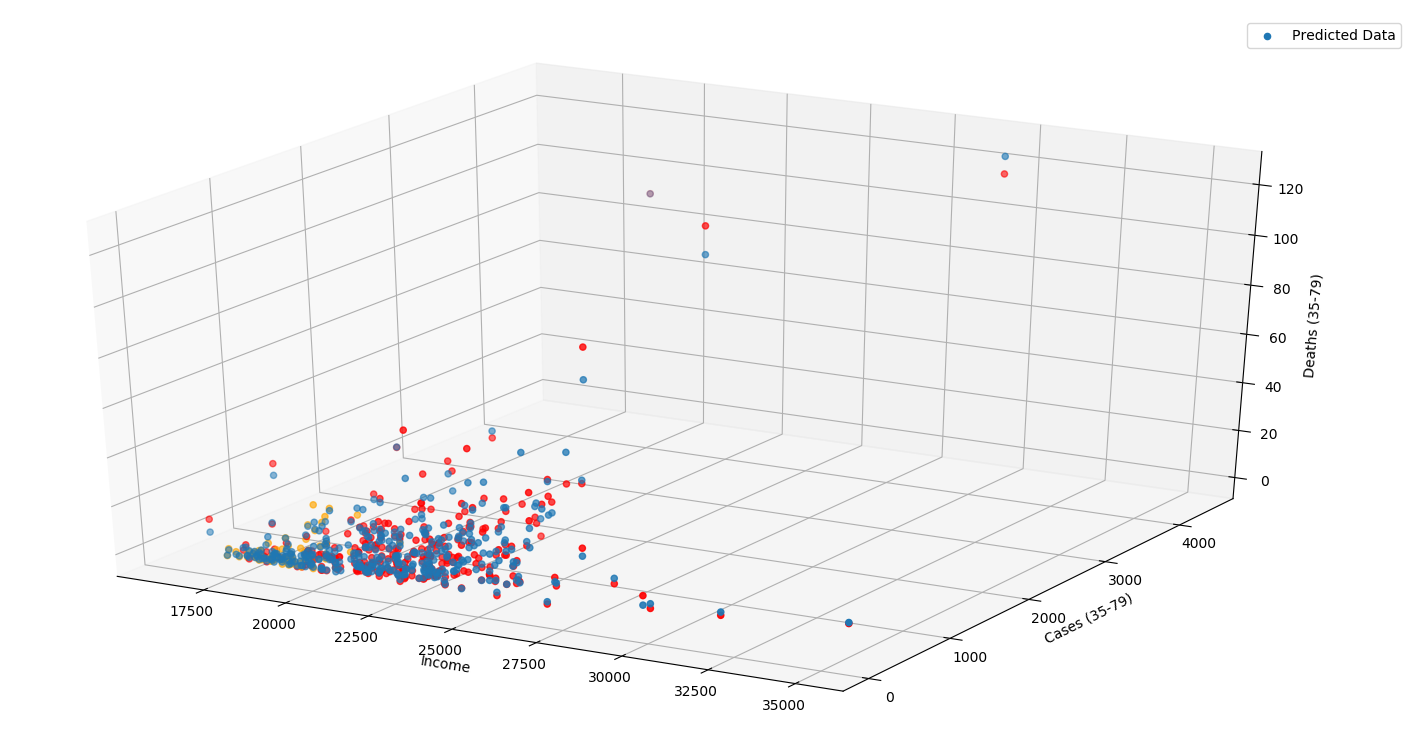}
    \caption{Plot of actual and predicted deaths in age group 35-79 with respect to income and infected patients in the same age group.}
    \label{fig:5.4}
\end{figure}

As discussed earlier, a slightly different approach was used to predict the total cases and total deaths. The total income, density, population distribution in the age groups 0-34, 35-79, 80+ were used to predict the total number of cases in the same age group. Figure \ref{fig:5.3} shows the plot of income vs the population in the age group 35-79 vs the total cases in the age group 35-79. Similar to the previous case, orange represents the Eastern German districts whereas the red represents the western districts and finally the blue points depict the predicted data. Only the mid-age group 35-79 was used for visualisation due to its significance to our study. It can be seen from the graph that the network was very well able to approximate the total number of cases in a particular age group when the total population of the same age group was provided. The network was able to capture most of the extreme points with high accuracy.
In a slight modification to the previous case, the network was trained with the information regarding the total cases in the chosen age group to predict the deaths in the same age group. As seen in figure \ref{fig:5.4}, it is evident that the network was successful in predicting the total deaths when provided with the total number of cases. The above two networks can be successfully used to predict the total cases or the total deaths of a certain region.
\\
\par
Despite the good performance of the network, we do not see the existence of classes as both the western and eastern districts were predicted with almost the same degree of accuracy. It would be premature to conclude at this stage about the non-existence of clusters in the data because it should also be noted that our input feature dimension is particularly small. As we have only considered around 350 districts in our study, it might be plausible that the network was not able to learn a rule to differentiate the two classes based on the hidden feature. To get a more distinct insight, we need to utilize the day series information for each of the districts. The next section discusses in detail the results when a day series data is used to predict the target.

\section{Time Series Data} \label{timeseries}
Due to the disadvantage of using the accumulated data, a time series data set which included the age of the pandemic was used for the further part of our study as explained before. This section deals with the results of different strategies used to set up and train the network. The section walks through all the different results for the strategies discussed in the previous chapter and gives the reasons for the same. After different trials with different network architectures, the network consisting of 15 layers with 50 neurons each with ELU activation function gave the most satisfying results. Logcosh loss function and Adam optimiser with 3 different reducing learn rates were used as it showed a better reduction in the loss function and hence resulted in better prediction.

\subsection{Time series data with cumulative cases}
This section motivates the consideration of the first day of appearance of the virus in the districts as one of our input features. In the initial trials, the density, income, absolute population in the age group and the age of the pandemic were used to predict the total cases and the deaths. As the total cumulative cases form a better parameter for visualisation, only the plots of the total cases are shown below. 
\begin{figure}[ht]
  \centering
  \begin{tabular}[b]{c}
    \includegraphics[width=6cm, height=5.5cm]{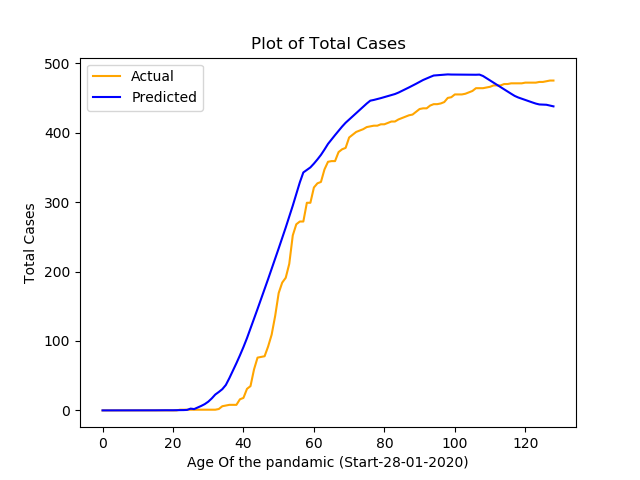} \\
    \small (a) Ingolstadt (west)
  \end{tabular} \qquad
  \begin{tabular}[b]{c}
    \includegraphics[width=6cm, height=5.5cm]{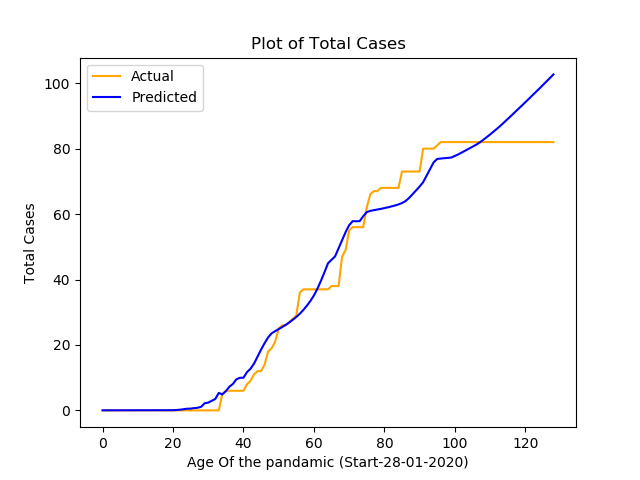} \\
    \small (b) Ostprignitz-Ruppin (East)
  \end{tabular}
  \begin{tabular}[b]{c}
    \includegraphics[width=6cm, height=5.5cm]{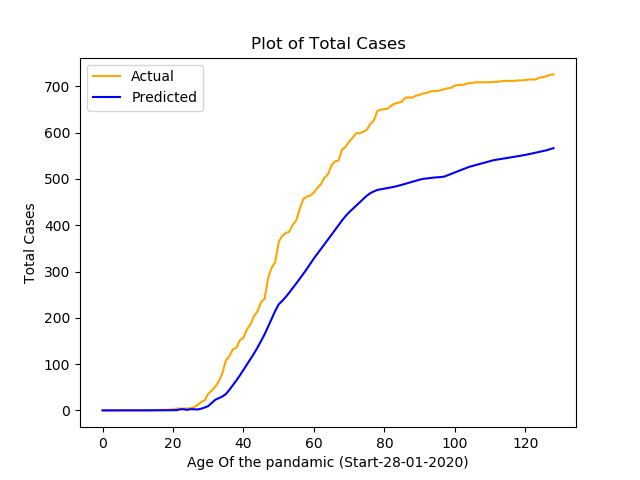} \\
    \small (c) Viersen (West)
  \end{tabular}
  \begin{tabular}[b]{c}
    \includegraphics[width=6cm, height=5.5cm]{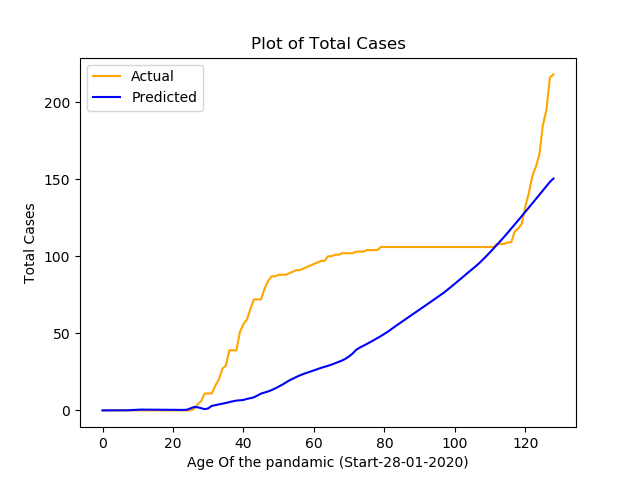} \\
    \small (d) Magdeburg (East)
  \end{tabular}  
  \caption{Plots of actual and predicted cumulative cases vs the age of the pandemic for different German districts}
  \label{fig:5.5}
\end{figure}
The network successfully learnt the data and predicted total cumulative cases per day with a certain accuracy. The network was trained with 3 different learn rates for Adam optimiser : $1\times10^{-3}$, $1\times10^{-4}$, $1\times0^{-5}$ with a batch size of 100 and 15 epochs each. During training, the logcosh loss function reduced the loss from 388 to 83 in the 3 runs with reducing learn rates. The actual and the predicted cumulative cases were plotted against the age of the pandemic for all the districts. Due to the space constraints, only selected plots are exhibited in the report. Figure \ref{fig:5.5} shows the plots of the total cumulative cases vs the age of the pandemic for 4 cities, 2 of each west and east. The orange represents the actual number of cases and the blue represents the predicted data. The x-axis represents the age of pandemic with 0 being the first day — 28.01.2018 and 129 being the last day considered for the study. As seen in the plot, the network had certain difficulties in finding the beginning of the curve i.e. the network in certain cases didn't predict the beginning of the pandemic accurately for some districts. The network predicted many of the targets quite accurately whereas some were over or underpredicted.
\newline
\par
Our aim of the study is to not only build a network to predict the cases but also to find the existence of classes if any. There are mainly two possible behaviour that the network might exhibit if BCG vaccine is indeed a hidden feature. The first possibility is that the network predicts one of the complete set i.e. west or east set accurately and fails to predict the other set. The second possibility is that the network mainly under predicts one set of data and over predicts the other. In this case, the network must accurately predict most of the western districts and over predict the eastern districts, if the classes exist. But as we see from the figure \ref{fig:5.5}, no clear rule exists which distinguish the western or eastern districts and therefore no conclusions can be made.  

\subsection{Time series data considering first day information}
\label{result : firstday}
\begin{figure}[ht]
  \centering
  \begin{tabular}[b]{c}
    \includegraphics[width=6cm, height=5.5cm]{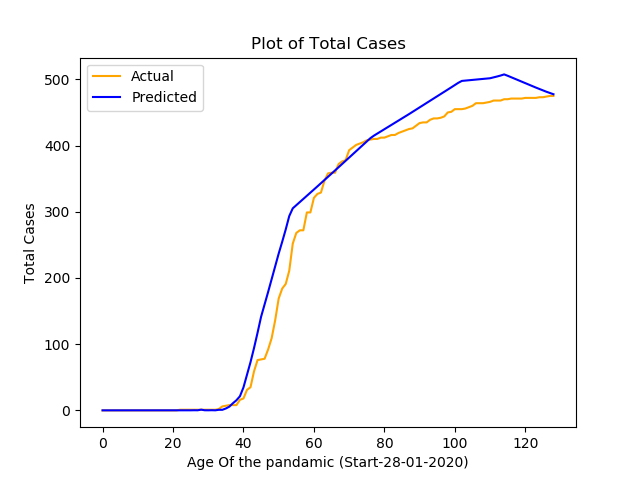} \\
    \small (a) Ingolstadt (west)
  \end{tabular} \qquad
  \begin{tabular}[b]{c}
    \includegraphics[width=6cm, height=5.5cm]{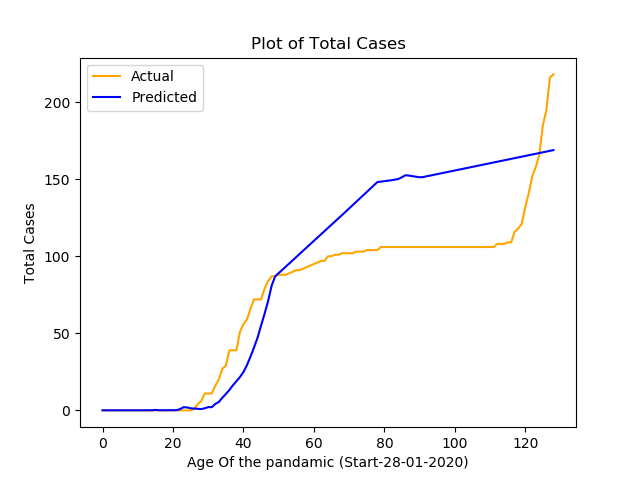} \\
    \small (b) Magdeburg(East)
    \end{tabular} \qquad
  \caption{Plots of actual and predicted cumulative cases including the first day information}
  \label{fig:5.6}
\end{figure}

In this strategy, as discussed in the previous chapter, the first-day information was used to train the network along with the other input features. The network reduced the loss function from 348 to 84. Though there wasn't a further reduction in the loss function compared to the previous case, the inclusion of the first-day information showed significant improvement in the prediction. No changes were made in the network other than the inclusion of a new parameter. For better understandability and visualisation, the plots of the same two districts: Ingolstadt and Magdeburg are shown in figure \ref{fig:5.6}. It is evident from the figure that due to this extra information provided, the network was able to generalize the curve much better and therefore was able to predict the start of the curve i.e. the beginning of the pandemic for all the chosen districts. Apart from the previous advantage, the added information also helped the network learn the reproduction rate of the virus much more accurately and therefore resulted in better prediction of the total cumulative cases.
\\
\par
Over $60\%$ of the districts were correctly predicted while others showed deviation from the actual cumulative cases. Just like the previous case, some districts in both east and west were over or under predicted but there was no clear rule that differentiated the western cities from the eastern cities. Since around $40\%$ of the districts were not accurately predicted, it is too premature to conclude the non-existence of classes. Therefore, a validation scheme was used to validate our speculation. Here one district from each West and East Germany was selected and tested on all the 3 separate networks as discussed before: the west network, the east network and the combined network. The west network is a network which is trained only by the western districts, the east network is trained only by the eastern districts and finally, the joint network is trained by all the districts selected. This approach was used for all the eastern and western districts but only the plots of few districts are presented here. More plots are available in Appendix \ref{Code and Dataset}.
\\
\par
\begin{figure}[ht]
  \centering
  \begin{tabular}[b]{c}
    \includegraphics[width=6cm, height=5.5cm]{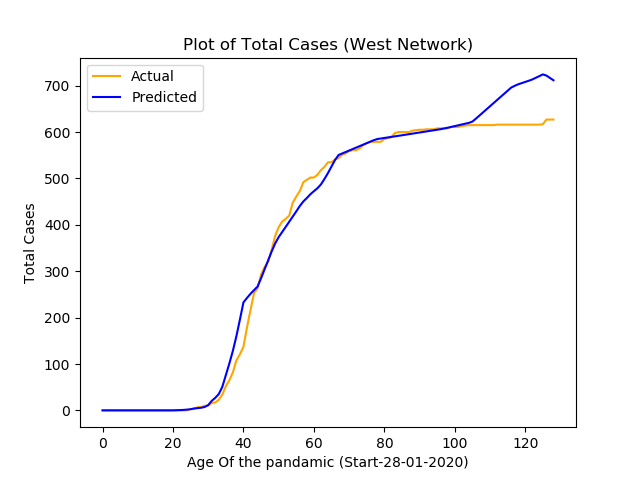} \\
    \small (a) Erding (West Network)
  \end{tabular} \qquad
  \begin{tabular}[b]{c}
    \includegraphics[width=6cm, height=5.5cm]{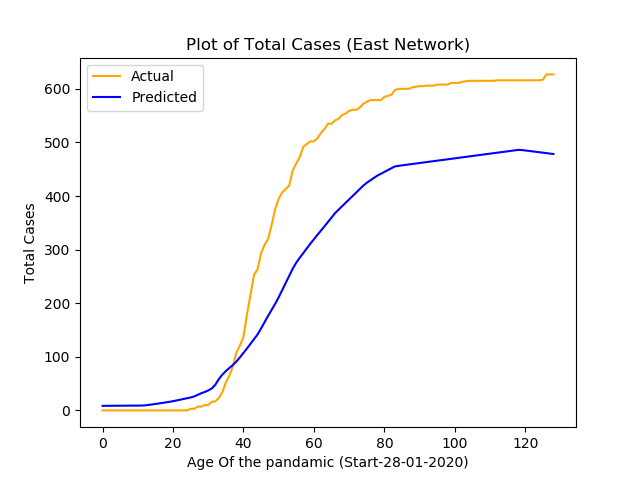} \\
    \small (b) Erding (East Network)
    \end{tabular} \qquad
      \begin{tabular}[b]{c}
    \includegraphics[width=6cm, height=5.5cm]{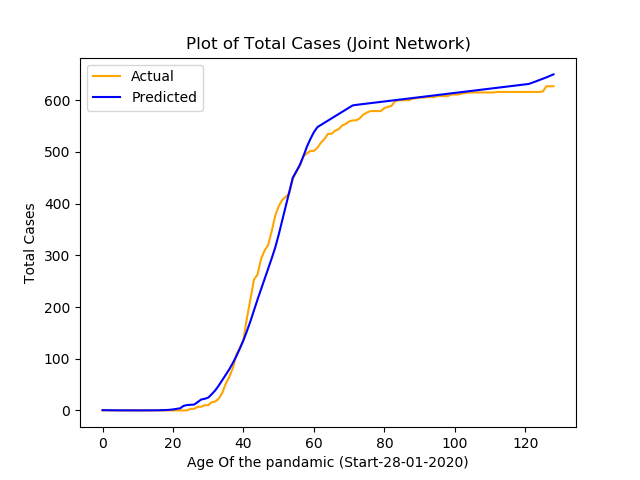} \\
    \small (c) Erding (Joint Network)
    \end{tabular} \qquad
  \caption{Plots of actual and predicted cumulative cases of Western district "Erding" using (a) west, (b) east and (c) joint network}
  \label{fig:5.7}
\end{figure}

A Western district Erding was selected and tested on the 3 different networks. Ideally, the western district must be exactly predicted by its western network, almost precisely predicted by the joint network and underpredicted by the eastern network. Figure \ref{fig:5.7} shows the plot of the western district Erding in all the 3 different networks. As we see from the figure, the plots show a similar trend as discussed. Most of the western districts were accurately predicted by the western network and many western districts were underpredicted by the eastern network. But in the joint network, only around 60\% - 70\% of the western districts were precisely predicted while others were both over or under predicted. Though it hinted towards the existence of clusters, it cannot be concluded unless a clear rule is visible.  
\\
\par
As per our method, the eastern network must also show a similar trend. An eastern district Saale-Holzland-Kreis was selected to be tested on 3 different networks. The eastern district should ideally show the same trend to point towards the existence of clusters. Nevertheless, many of the eastern districts didn't follow the trend. As the number of the eastern districts are relatively lower compared to the western districts, it is highly unlikely that the joint network gives precise prediction for the eastern districts. Therefore, the eastern district must ideally be predicted precisely by its own network, overpredicted by the joint network and also overpredicted by the western network.
\newline

\begin{figure}[ht]
  \centering
  \begin{tabular}[b]{c}
    \includegraphics[width=6cm, height=5.5cm]{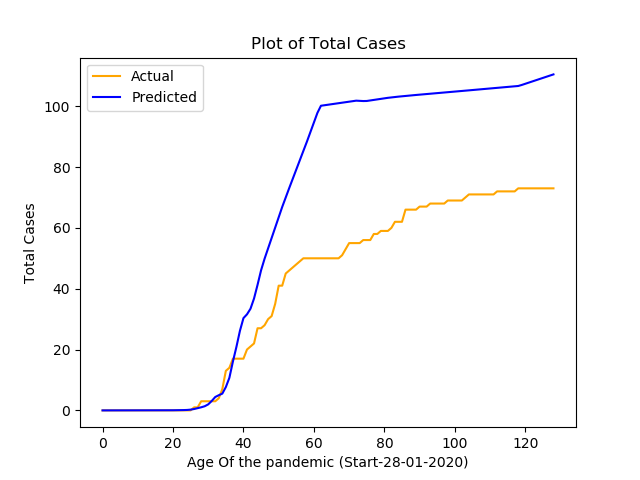} \\
    \small (a) Saale-Holzland-Kreis \\ (West Network)
  \end{tabular} \qquad
  \begin{tabular}[b]{c}
    \includegraphics[width=6cm, height=5.5cm]{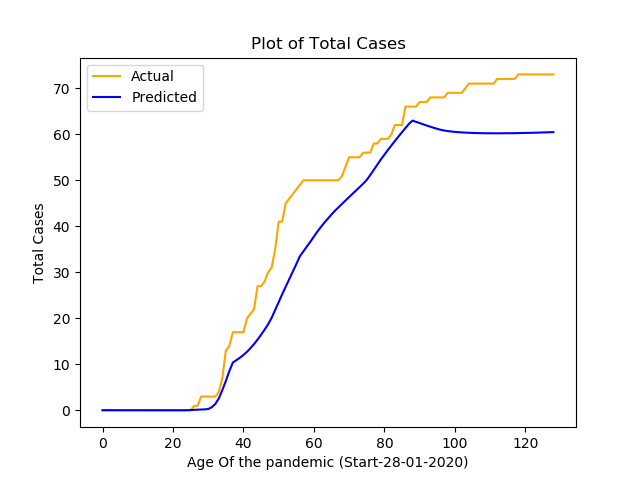} \\
    \small (b) Saale-Holzland-Kreis \\ (East Network)
    \end{tabular} \qquad
      \begin{tabular}[b]{c}
    \includegraphics[width=6cm, height=5.5cm]{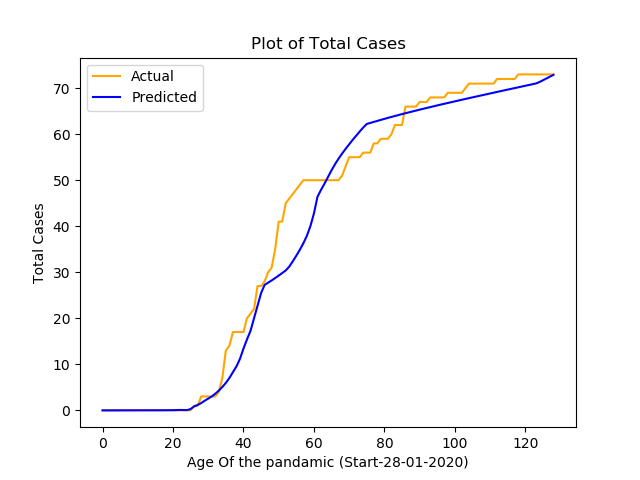} \\
    \small (c) Saale-Holzland-Kreis \\ (Joint Network)
    \end{tabular} \qquad
  \caption{ Plots of actual and predicted cumulative cases of Eastern district "Saale-Holzland-Kreis" using (a) west, (b) east and (c) joint network}
  \label{fig:5.8}
\end{figure}

Figure \ref{fig:5.8} shows the plot of the eastern district Saale-Holzland-Kreis using the 3 different networks. As we see in the figure, the district was overpredicted by the western network. Around 80\% of the eastern districts were overpredicted by the western network. Unlike the previous case, the eastern network was not completely successful in generalizing the data when evacuated on its own data. The joint network gave mixed results. Around 40\% of the eastern districts were accurately predicted by the joint network whereas 30\% were over predicted and the remaining 30\% were under predicted. The eastern districts didn't follow a unique trend and therefore, at this stage, it is not possible to conclude that the eastern and western districts form independent clusters. No clear rule was seen that would suggest that the eastern districts form a cluster. Since the eastern network was trained using eastern districts alone, it is unnatural that the eastern districts were not well predicted by its own network.
\newline
\par
To prove our claim, it is important that the eastern and the western network generalizes the data well and gives accurate predictions when evaluated by its own network. However, within the eastern and western districts, it could also be possible that they do not show a unique behaviour in the rise of cases and hence not possible for the network to generalize the curve for its network. Nevertheless, different approaches were tried in an aim to build a network that can generalize the data better and possibly classify the data as speculated.  Therefore, in the upcoming strategies, certain changes were made in the input parameters to improve the learning and thereby give better predictions.

\subsection{Time series data considering past data}
\label{result : pastdata}
To improve the learning, the active cases in the past 7 days were added to the input as discussed in the previous chapter. This strategy aimed to improve learning for a better prediction of the target. Unlike the previous case, the active cases were used as the target. The network that was trained using the past 7 days' information gave extremely precise prediction with the upper limit of error between the actual and predicted data being $\pm 5$ cases. 
\begin{figure}[ht]
  \centering
  \begin{tabular}[b]{c}
    \includegraphics[width=6cm, height=5.5cm]{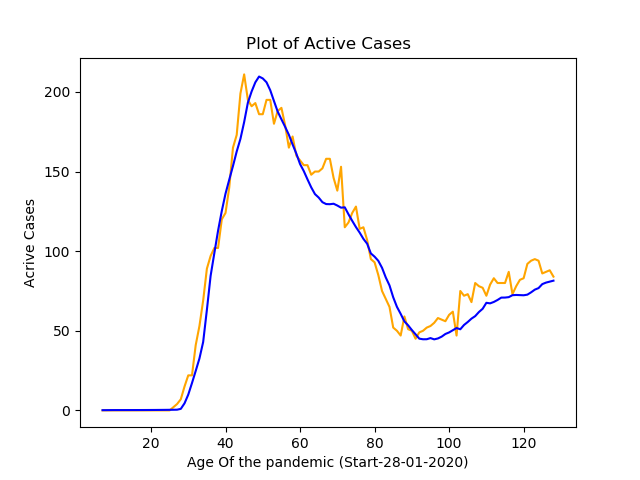} \\
    \small (a) Kleve (West)
  \end{tabular} \qquad
  \begin{tabular}[b]{c}
    \includegraphics[width=6cm, height=5.5cm]{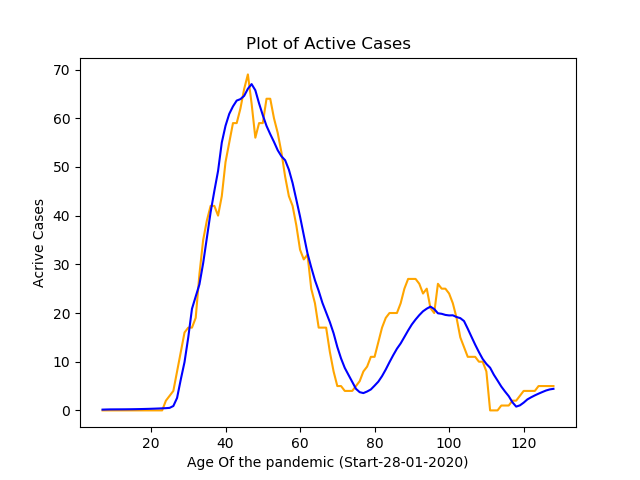} \\
    \small (b) Rostock (East)
    \end{tabular} \qquad
  \caption{Plots of actual and predicted active cases for (a) Kleve and (b) Rostock using past data}
  \label{fig:5.9}
\end{figure}
As too much information was provided to the network, in a slight modification to the previous case, the average of the cases in the last 7 days was considered. The network again gave efficient results and was able to predict the active cases of all the selected districts. The network successfully learnt the bell curve and predicted both the eastern and the western districts accurately. Plots of active cases of Kleve and Rostock, west and east districts respectively, are shown in the figure \ref{fig:5.9}. Orange represents the actual active cases and blue represents the predicted cases. Since the average past information is provided along with the day information, the network was able to learn the reproduction rate precisely and was able to correct the direction of the curve. The network ignored the minor variation in the active cases and found a smooth fit through the curve of active cases. As the past information is provided, the network easily predicted the new active case for all the districts.  No clusters of districts were seen which were over/under predicted. Since the data is able to completely generalize the curve, this strategy cannot be used to classify the data based on our speculation. However, the network can be used to predict future active cases for a certain number of days. The network will evidently not change the direction of the curve but can be used for minor extrapolation. As the above scheme cannot be used to find the clusters, this method is not continued in our future strategies. 

\subsection{Time series data considering log of cases}\label{section:resutlog}
The use of logarithm of total cases resulted in good learning and in some cases, better results were achieved. It is evident that the network learns linear data much better than the exponential data. As the cumulative cases are mostly exponential curves, the logarithm of the cumulative cases resulted in a linear curve which was well captured by the network, improving the overall performance of the network. The plots of all the districts with respect to the log of cases and the age of the pandemic were plotted and analysed. Many districts were precisely predicted. It is important to consider that since the log of total cases is used, the upper part of the y-axis is squashed and therefore a small difference in the actual and predicted cases in the log plot would result in a much larger difference in the plot without considering the log. Overall, the network gave accurate results for many districts while some districts from both east and west were over or under predicted. The east and the west districts did not show a tendency to form clusters. Similar to section \ref{result : firstday}, the validation strategy was used to check the predictions in different networks. In this section, one district from each West and East Germany is selected and is evaluated on the three separate networks: west network, east network and the joint network.
\\
\begin{figure}[!htb]
  \centering
  \begin{tabular}[b]{c}
    \includegraphics[width=6cm, height=5cm]{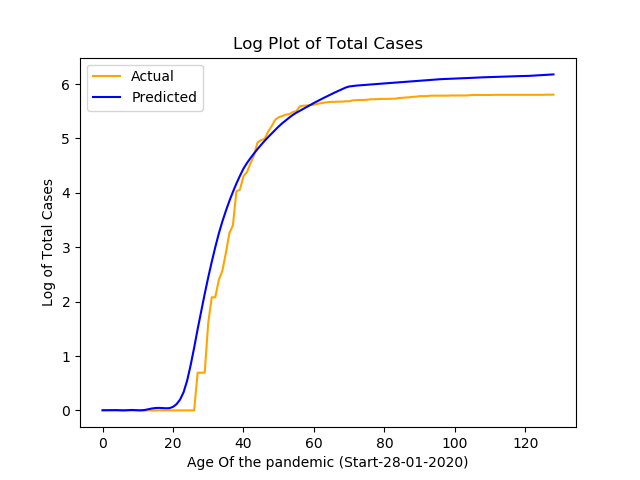} \\
    \small (a) Braunschweig (W) \\ West Network
  \end{tabular} \qquad
  \begin{tabular}[b]{c}
    \includegraphics[width=6cm, height=5cm]{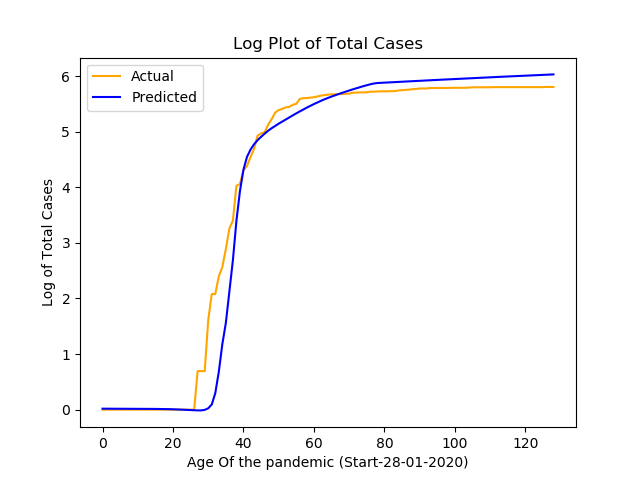} \\
    \small (b) Braunschweig (W) \\ East Network
    \end{tabular} \qquad
  \begin{tabular}[b]{c}
    \includegraphics[width=6cm, height=5cm]{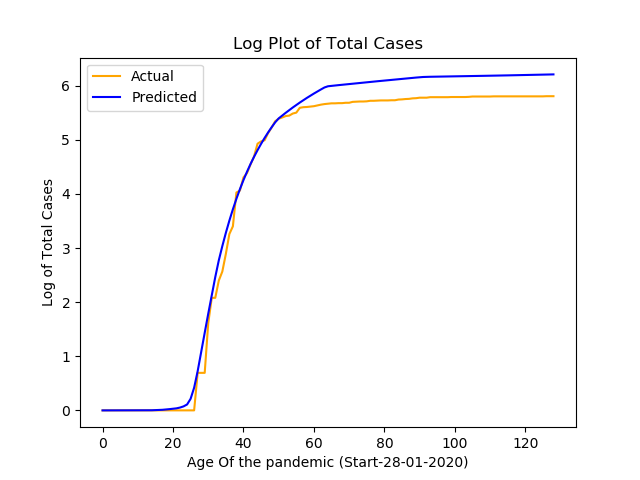} \\
    \small (c) Braunschweig (W) \\ Joint Network
    \end{tabular} \qquad
  \begin{tabular}[b]{c}
    \includegraphics[width=6cm, height=5cm]{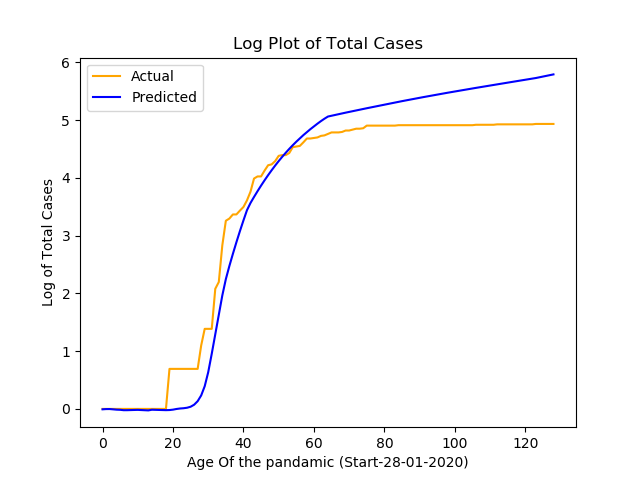} \\
    \small (d) Vorpommern-Greifswald (E) \\ West Network
    \end{tabular} \qquad
  \begin{tabular}[b]{c}
    \includegraphics[width=6cm, height=5cm]{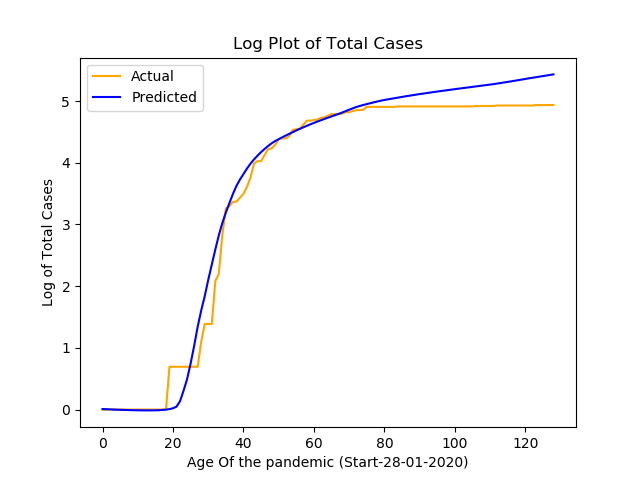} \\
    \small (e) Vorpommern-Greifswald (E) \\ East Network
    \end{tabular} \qquad
  \begin{tabular}[b]{c}
    \includegraphics[width=6cm, height=5cm]{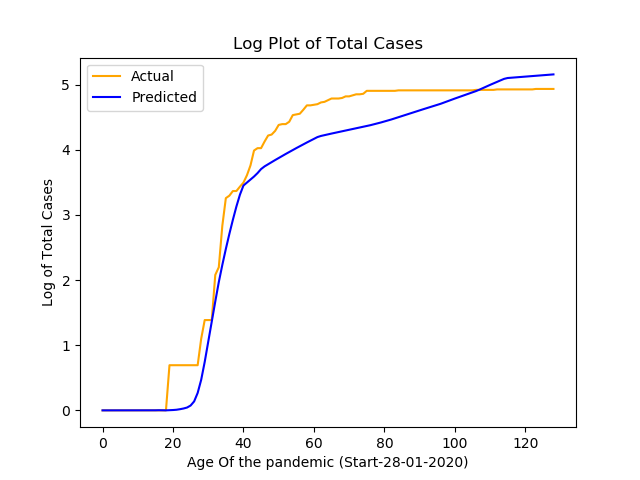} \\
    \small (f) Vorpommern-Greifswald (E) \\ Joint Network
    \end{tabular} \qquad    
  \caption{Log plots of actual and predicted cumulative cases of Braunschweig (Western district) and Vorpommern-Greifswald (Eastern district) on west, east and joint network respectively.}
  \label{fig:5.10}
\end{figure}
\par
Figure \ref{fig:5.10} shows the log plot of Braunschweig and Vorpommern Greifswald which are western and eastern districts of Germany respectively. The first plot represents the data predicted by the western network, second by the eastern network and finally the third by the joint network. Braunschweig showed no difference in the predictions when evaluated on the 3 districts, where it was over predicted by all the 3 networks. Over 60\% of the districts were correctly predicted by its own network and the rest 40\% were equally over/under predicted. Over 80\% of the western districts were underpredicted by the eastern network. Similar to section \ref{result : firstday} over 70\% of the western districts were accurately predicted in the joint network. The overall results were slightly better than that of the network without considering log of cases. For the eastern district Vorpommern-Greifswald, no huge difference was seen when evaluated on the 3 different networks. In certain cases, the eastern districts were better predicted in the joint network than its own or western network. In this strategy too, the eastern districts were not clearly generalised by its own network. The eastern districts gave slightly better predictions when evaluated on its own network than the western network but were both over or under predicted in both the cases. Very few eastern districts were accurately predicted in the two networks. To conclude the existence of two clusters, it is important that the eastern district is accurately predicted in its own network and over predicted in the western network which is clearly not the case here. Finally, the combined network gave better predictions that two and was comparable to that of western districts which again hinted towards non-existence of clusters.
\newline
\par
The experiment conducted on the mid-age group 35-79 exhibited a very similar characteristic as above. The absolute population in the age groups were replaced with the mid-age group as it forms a more relevant input feature and thus must exhibit the existence of classes if any. However, the plots showed no improvement from the previous case. In the final section, a more generic approach was used; the relative distribution of the population was used to finalize our study and confirm our results. Since there were inaccuracies in the prediction, this strategy motivated the use of the relative proportion of the population.

\subsection{Time series data considering relative proportions}
In the final strategy, as the relative proportion of the population is used to predict the relative cumulative cases per day, it is a very generalized approach. Here the network doesn't learn absolute numbers but learns the relative fractions which are a better choice of input features. Since the network learnt the relative numbers, the network can be further used for prediction of cases around the world.
\newline
\begin{figure}[!htb]
  \centering
  \begin{tabular}[b]{c}
    \includegraphics[width=6cm, height=5cm]{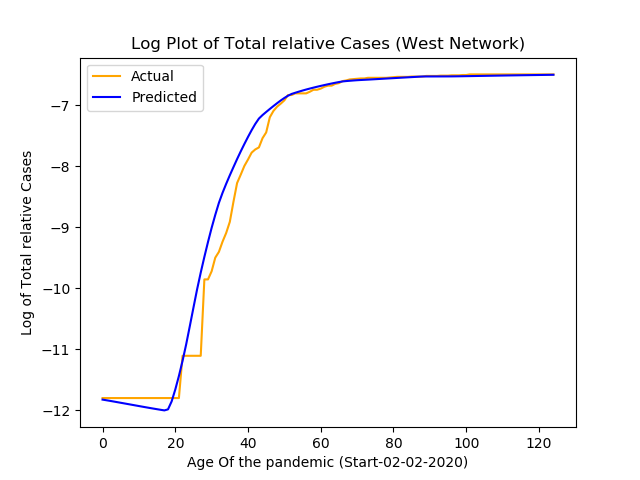} \\
    \small (a) Donau-Ries (W) \\ West Network
  \end{tabular} \qquad
  \begin{tabular}[b]{c}
    \includegraphics[width=6cm, height=5cm]{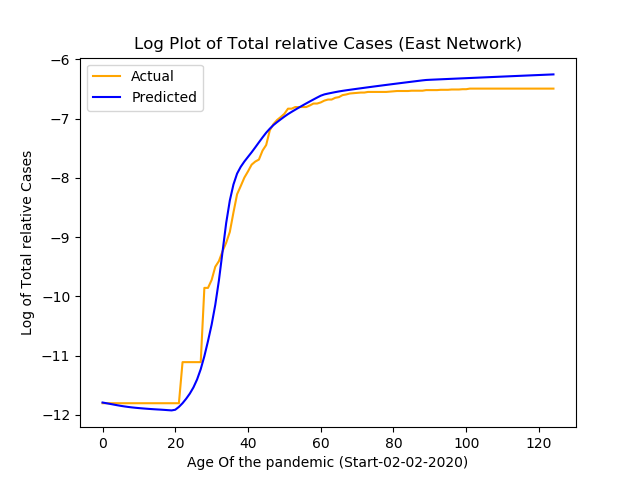} \\
    \small (b) Donau-Ries (W) \\ East Network
    \end{tabular} \qquad
  \begin{tabular}[b]{c}
    \includegraphics[width=6cm, height=5cm]{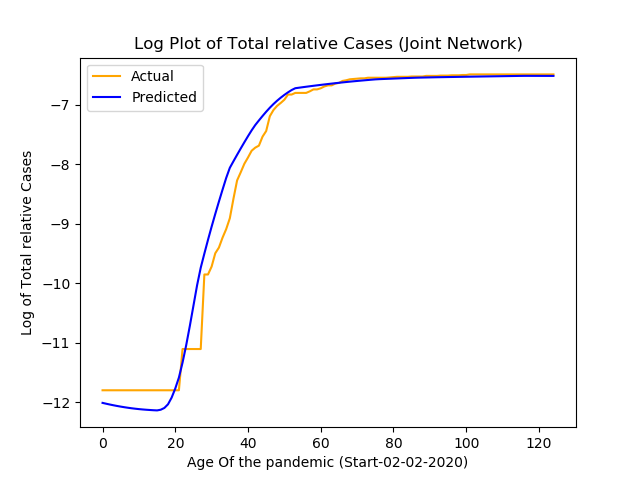} \\
    \small (c) Donau-Ries (W) \\ Joint Network
    \end{tabular} \qquad
  \begin{tabular}[b]{c}
    \includegraphics[width=6cm, height=5cm]{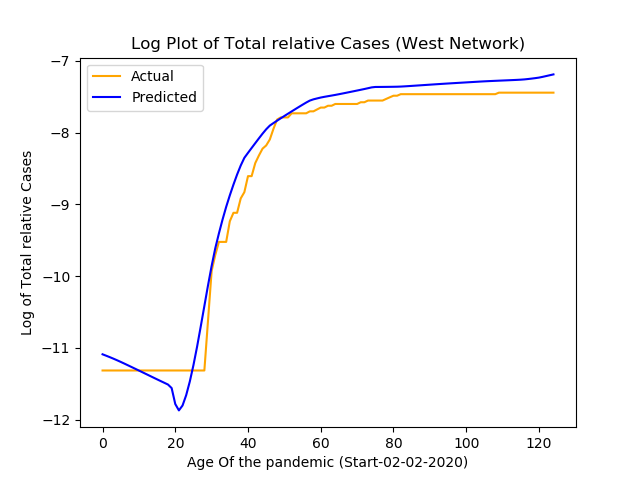} \\
    \small (d) Weimarer Land (E) \\ West Network
    \end{tabular} \qquad
  \begin{tabular}[b]{c}
    \includegraphics[width=6cm, height=5cm]{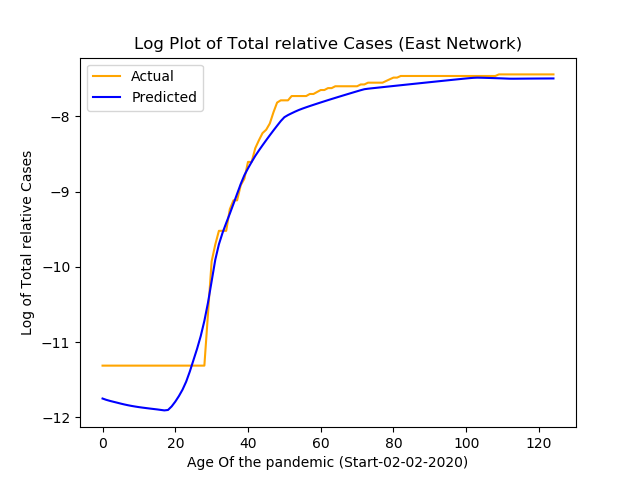} \\
    \small (e) Weimarer Land (E) \\ East Network
    \end{tabular} \qquad
  \begin{tabular}[b]{c}
    \includegraphics[width=6cm, height=5cm]{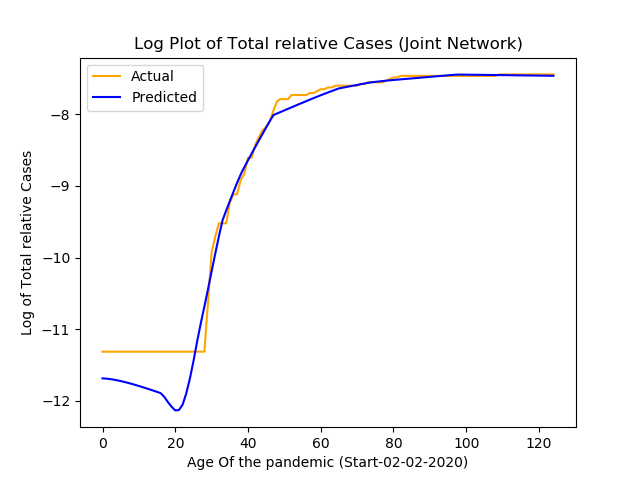} \\
    \small (f) Weimarer Land (E) \\ Joint Network
    \end{tabular} \qquad    
  \caption{Log plots of actual and predicted relative cases of Donau-Ries (Western district) and Weimarer Land (Eastern district) on west, east and joint network respectively}
  \label{fig:5.11}
\end{figure}

Similar to the previous case, since the log of the total relative cases were used, the network was able to provide good predictions. Since the log of fractions is negative, the graph is plotted in the negative y-axis. Among all the different approaches, this strategy was able to give the best prediction and was able to generalize the curve for most of the districts. Unlike the previous case, the network captured the upper part of the curve more precisely and had slight difficulties in learning the curve at the beginning of the pandemic for some districts. Overall the network exhibited good performance in predicting the total relative cases. As anticipated, the network was able to predict most of the districts in both east and west, and over or under predicted the rest, showing no evidence of the existence of class. In this section, the plots of variant using the relative population of the mid-age group of 35-79 used as an input feature to the network, is shown. Two districts, one from each west and east respectively, were chosen to validate our theory by evaluating the data on the 3 selected network. Figure \ref{fig:5.11} shows the log plot of relative cases of Donau-Reis and Weimarer Land evaluated on the west, east and finally the joint network. From the figure, it can be seen that there is no much difference in prediction for Donau-Ries when evaluated on the west and joint networks whereas it is slightly over predicted in the east network. Only around 30\% of the western districts were under-predicted in the eastern network and under 50\% were over predicted which is the quite opposite to the previous case using absolute population. This behaviour is opposite to the behaviour required to conclude that western and eastern districts form clusters. Furthermore, the west and the joint network performed excellently when evaluated on the western districts. The eastern district Weimarer Land was the precisely predicted in the east and joint network and was slightly over predicted in the East network. Getting into the statistics, around 60\% of eastern districts were overpredicted, 20\% were underpredicted and the rest were accurately predicted by the west network. Though the districts were inaccurately predicted in the counter network, the error between the actual and the predicted data was relatively lower compared to the other strategies. Overall, this network gave the best results and was able to generalize the curve efficiently. Despite being wrongly predicted in the counter network to some extent, the districts were accurately predicted in their own and combined network. This is a clear indication that two clusters of data don't exist and it further justifies that eastern and western districts together form a single cluster at least based on our speculated invisible feature. Based on our theory, this proves the non-existence of classes and hence can be concluded that the BCG vaccine doesn't form the hidden feature, inferring that the vaccine does not affect the reduction in the spread of the virus. The whole of Germany, therefore, showed a similar trend in the rise and control of the pandemic.

\subsection{Validation of the result}
To further prove that the vaccine information doesn't form the invisible feature, the network was finally trained using the vaccine information, i.e. the information regarding East and West Germany was included into the input feature to check if there was any difference in the predictions. Therefore, the network was trained using disposable income, density, relative proportion of population, first-day information, east and west information and day index to predict the log of relative cumulative cases.
\begin{figure}[!htb]
  \centering
  \begin{tabular}[b]{c}
    \includegraphics[width=6cm, height=5cm]{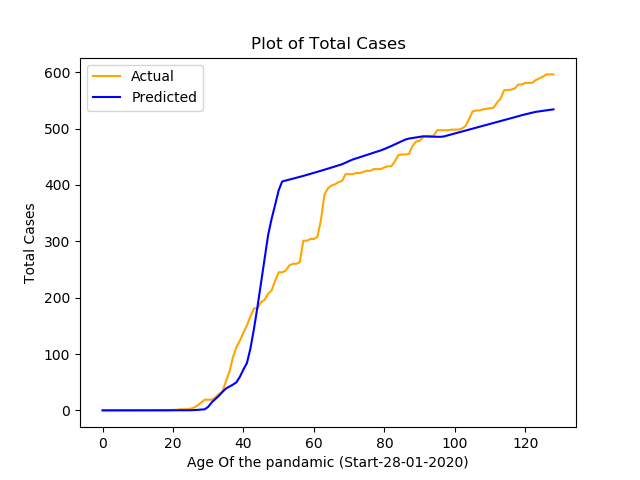} \\
    \small (a) Bochum (W) \\ considering vaccine \\ information
  \end{tabular} \qquad
  \begin{tabular}[b]{c}
    \includegraphics[width=6cm, height=5cm]{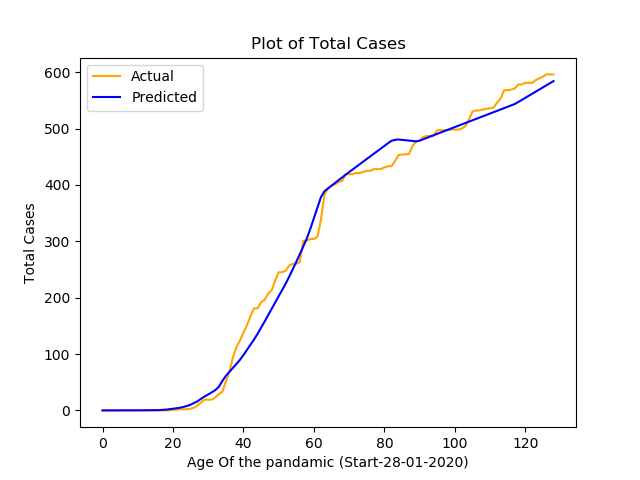} \\
    \small (b) Bochum (W) \\ without considering vaccine \\ information
    \end{tabular} \qquad
  \begin{tabular}[b]{c}
    \includegraphics[width=6cm, height=5cm]{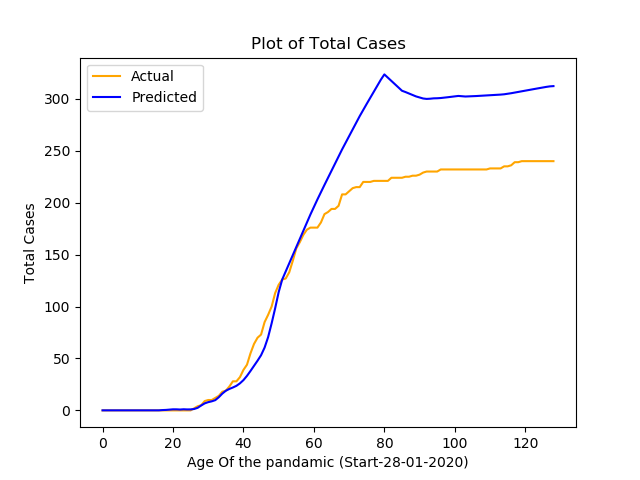} \\
    \small (c) Meißen (East) \\ considering vaccine \\ information
    \end{tabular} \qquad
  \begin{tabular}[b]{c}
    \includegraphics[width=6cm, height=5cm]{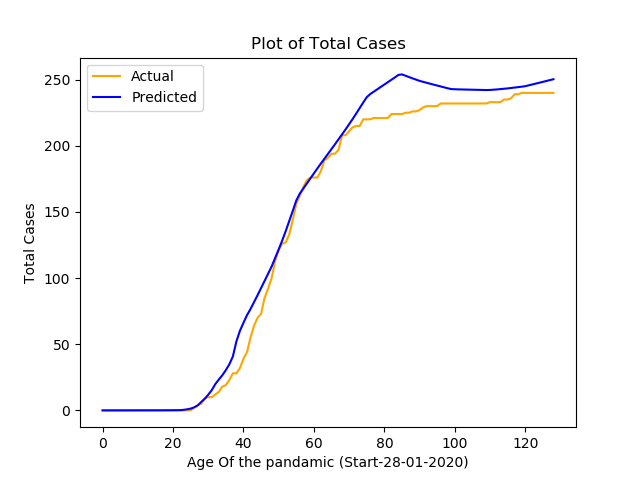} \\
    \small (d) Meißen (East) \\ without considering vaccine \\ information
    \end{tabular} \qquad
  \caption{plot of actual and predicted cases with (left) and without (right) considering the vaccine information}
  \label{fig:5.12}
\end{figure}
Figure \ref{fig:5.12} compares the plot of actual and predicted cases with and without considering the vaccine information as one of the input features. Since it is harder to compare log of cases, antilog is taken to plot total cumulative cases vs age of the pandemic. Here few plots are shown for which the vaccine information deteriorated the performance of the network. For most of the districts, there was no change in the performance and for very few, there was minor improvement when vaccine information was considered. However, there were no improvements for most of the over or under predicted districts. Therefore, based on our research, it confirms that BCG vaccine is not a candidate for an invisible feature and therefore is not a major factor in the reduction of the spread of the virus.

\chapter{Conclusion and future directions} 

In this work, we present a method to classify the data based on the invisible features and tried to classify the western and eastern part of Germany considering tuberculosis vaccine as an invisible feature. We successfully built an artificial neural network with logcosh loss function which learns the bigger cluster of data when clustered data is provided as an input. Our method, when used on set-values functions, will learn the majority branch of the set-valued function and this theory can be used to classify data. We introduced a neural network which not only helps to classify the data based on an invisible feature but also predicts the major cluster with a high degree of accuracy. We used this theory to check if the tuberculosis vaccine provides immunity towards the coronavirus.
\newline
\par
We presented a neural network to predict the total number of cases, logarithm of total cumulative cases, active cases, deaths and other parameters based on certain selected input features. As we know, acquiring relevant data to represent the model is one of the most important tasks in the field of deep learning, we built an efficient data set to precisely represent our model. We compiled different data sets to obtain information such as income, area, population, population distribution among the age groups, total cumulative coronavirus cases, deaths, active cases and other information regarding the virus for each German district. Through different strategies, we presented different networks for both accumulated and day series data.
\newline
\par
Based on the results of the different strategies, we concluded that the Eastern and the Western German districts did not form two separate clusters as speculated. The BCG vaccine did not form a hidden\textbackslash invisible feature and therefore inferring that the vaccine indeed did not affect in reducing the spread of the virus. The whole of Germany showed a similar trend in the spread of coronavirus though the number of cases in Eastern Germany was relatively lower. 
\newline
\par
Since only limited information was used to build our network, the data set representing the model can be improved by including more critical information regarding the virus. With the strategy which included past information, the model can be used to predict the future cases up to a certain day. The final model, which was built using the relative proportion of the population and cases is very generalized and can be further used for prediction of cases around the world. The data set and the model can be further developed to improve the overall performance of the network. The theory of classification of data sets based on the invisible feature can be extensively used to classify different real-world problems and can also be used to validate if a parameter is an invisible feature or not.


\appendix 



\chapter{Code and Data-set} 

\label{Code and Dataset} 

Source code, data sets and plots of the different strategies used are available in the GitHub link provided below: 
\begin{center}
    \url{https://github.com/nihalacharyaa/Studienarbeit.git}
\end{center}



\printbibliography[heading=bibintoc]

@article{classification,
author = {Moshagen, Thilo and Navilarekal Rajgopa, Ajay and Acharya Adde, Nihal},
year = {2020},
title = {Finding laws valid inside a Data Set and classifying it using Neural Networks [to be published: November 2020]},
url = {https://www.overleaf.com/read/xnvsrxjbrdgt},
}

@online{FeatureSelection,
    Author = {Jason Brownlee},
    Title = {How to Choose a Feature Selection Method For Machine Learning},
    Url = {https://machinelearningmastery.com/feature-selection-with-real-and-categorical-data/},
    Year = {2019}}

@online{FeatureCorrelation,
    Author = {Will Badr},
    Title = {Why Feature Correlation Matters},
    Url = {https://towardsdatascience.com/why-feature-correlation-matters-a-lot-847e8ba439c4},
    Year = {2019}}

@online{lossfuncations,
    Author = {Prince Grover},
    Title = {5 Regression Loss Functions All Machine Learners Should Know},
    Url = {https://heartbeat.fritz.ai/5-regression-loss-functions-all-machine-learners-should-know-4fb140e9d4b0},
    Year = {2018}}

@book{Goodfellow-et-al-2016,
    title={Deep Learning},
    author={Ian Goodfellow and Yoshua Bengio and Aaron Courville},
    publisher={MIT Press},
    note={\url{http://www.deeplearningbook.org}},
    year={2016}
}

@online{activation,
    Author = {Casper Hansen},
    Title = {Activation Functions Explained - GELU, SELU, ELU, ReLU and more},
    Url = {https://mlfromscratch.com/activation-functions-explained/#/},
    Year = {2019}}

@online{activationpics,
    Author = {Rinat Maksutov},
    Title = {Deep study of a not very deep neural network. Part 2: Activation functions},
    Url = {https://towardsdatascience.com/deep-study-of-a-not-very-deep-neural-network-part-2-activation-functions-fd9bd8d406fc/},
    Year = {2018}}

@article{LeCunBoserDenkerEtAl89,
  author = {LeCun, Y. and Boser, B. and Denker, J. S. and Henderson, D. and Howard, R. E. and Hubbard, W. and Jackel, L. D.},
  description = {CCNLab BibTeX},
  journal = {Neural Computation},
  pages = {541-551},
  title = {Backpropagation Applied to Handwritten Zip Code Recognition},
  volume = 1,
  year = 1989
}

@article{lecun2015deeplearning,
  author = {LeCun, Yann and Bengio, Yoshua and Hinton, Geoffrey},
  bibsource = {dblp computer science bibliography, https://dblp.org},
  description = {Deep Learning first paper.},
  doi = {10.1038/nature14539},
  journal = {Nature},
  number = 7553,
  pages = {436--444},
  title = {Deep Learning},
  url = {https://doi.org/10.1038/nature14539},
  volume = 521,
  year = 2015
  }

@article{MAL-006,
url = {http://dx.doi.org/10.1561/2200000006},
year = {2009},
volume = {2},
journal = {Foundations and Trends® in Machine Learning},
title = {Learning Deep Architectures for AI},
doi = {10.1561/2200000006},
issn = {1935-8237},
number = {1},
pages = {1-127},
author = {Yoshua Bengio}
}

@book{SVM,
editor = {Sch\"{o}lkopf, Bernhard and Burges, Christopher J. C. and Smola, Alexander J.},
title = {Advances in Kernel Methods: Support Vector Learning},
year = {1999},
isbn = {0262194163},
publisher = {MIT Press},
address = {Cambridge, MA, USA}
}

@article{NN,
  author = {Michael Nielsen},
  journal = {Determination press San Francisco, CA,},
  title = {Neural Networks and Deep Learning},
  volume = 2018,
  year = 2015
}

@article{journals/nn/Qian99,
  author = {Qian, Ning},
  ee = {https://www.wikidata.org/entity/Q52019658},
  journal = {Neural Networks},
  number = 1,
  pages = {145-151},
  title = {On the momentum term in gradient descent learning algorithms.},
  url = {http://dblp.uni-trier.de/db/journals/nn/nn12.html#Qian99},
  volume = 12,
  year = 1999
}

@article{loss,
author = {Nie, Feiping and Zhanxuan, Hu and Li, Xuelong},
year = {2018},
month = {01},
pages = {37-52},
title = {An investigation for loss functions widely used in machine learning},
volume = {18},
journal = {Communications in Information and Systems},
doi = {10.4310/CIS.2018.v18.n1.a2}
}

@online{WHOTB,
    Author = {WHO (World Health Organization)},
    Title = {Tuberculosis},
    Url = {https://www.who.int/news-room/fact-sheets/detail/tuberculosis},
    Year = {2020}}

@article{WHOBCG,
	author = {World Health Organization = Organisation mondiale de la Santé},
	journal = {Weekly Epidemiological Record = Relevé épidémiologique hebdomadaire},
	title = {BCG vaccines: WHO position paper – February 2018 – Vaccins BCG: Note de synthèse de l’OMS – Février 2018},
	volume = {93},
	year = {2018},
	pages = {73 - 96},
	number = {08},
	publisher = {World Health Organization = Organisation mondiale de la Santé},
	type = {Journal / periodical articles}
}

@article {Neteaaaf1098,
	author = {Netea, Mihai G. and Joosten, Leo A. B. and Latz, Eicke and Mills, Kingston H. G. and Natoli, Gioacchino and Stunnenberg, Hendrik G. and O{\textquoteright}Neill, Luke A. J. and Xavier, Ramnik J.},
	title = {Trained immunity: A program of innate immune memory in health and disease},
	volume = {352},
	number = {6284},
	elocation-id = {aaf1098},
	year = {2016},
	doi = {10.1126/science.aaf1098},
	publisher = {American Association for the Advancement of Science},
	issn = {0036-8075},
	URL = {https://science.sciencemag.org/content/352/6284/aaf1098},
	eprint = {https://science.sciencemag.org/content/352/6284/aaf1098.full.pdf},
	journal = {Science}
}

@article {covidBCG,
	author = {O'Neill and Luke A. J. and Netea and Mihai G.},
	title = {BCG-induced trained immunity: can it offer protection against COVID-19?},
	volume = {20},
	number = {335},
	elocation-id = {O'Neill2020},
	year = {2020},
	doi = {10.1038/s41577-020-0337-y},
	issn = {1474-1741},
	URL = {https://doi.org/10.1038/s41577-020-0337-y},
	eprint = {https://science.sciencemag.org/content/352/6284/aaf1098.full.pdf},
	journal = {Nature Reviews Immunology}
}

@article {Escobar202008410,
	author = {Escobar, Luis E. and Molina-Cruz, Alvaro and Barillas-Mury, Carolina},
	title = {BCG vaccine protection from severe coronavirus disease 2019 (COVID-19)},
	elocation-id = {202008410},
	year = {2020},
	doi = {10.1073/pnas.2008410117},
	publisher = {National Academy of Sciences},
	issn = {0027-8424},
	URL = {https://www.pnas.org/content/early/2020/07/07/2008410117},
	eprint = {https://www.pnas.org/content/early/2020/07/07/2008410117.full.pdf},
	journal = {Proceedings of the National Academy of Sciences}
}

@article {covidEASTWEST,
	author = {Hauer, Julia and Fischer, Ute and Auer, Franziska and Borkhardt, Arndt.},
	title = {Regional BCG vaccination policy in former East- and West Germany may impact on both severity of SARS-CoV-2 and incidence of childhood leukemia},
	volume = {34},
	number = {2217},
	elocation-id = {Hauer2020},
	year = {2020},
	doi = {10.1038/s41375-020-0871-4},
	issn = {1476-5551},
	URL = {https://doi.org/10.1038/s41375-020-0871-4},
	eprint = {https://science.sciencemag.org/content/352/6284/aaf1098.full.pdf},
	journal = {Leukemia}
}

@article {Miller2020,
	author = {Miller, Aaron and Reandelar, Mac Josh and Fasciglione, Kimberly and Roumenova, Violeta and Li, Yan and Otazu, Gonzalo H},
	title = {Correlation between universal BCG vaccination policy and reduced morbidity and mortality for COVID-19: an epidemiological study},
	elocation-id = {2020.03.24.20042937},
	year = {2020},
	doi = {10.1101/2020.03.24.20042937},
	publisher = {Cold Spring Harbor Laboratory Press},
	URL = {https://www.medrxiv.org/content/early/2020/03/28/2020.03.24.20042937},
	eprint = {https://www.medrxiv.org/content/early/2020/03/28/2020.03.24.20042937.full.pdf},
	journal = {medRxiv}
}

@article {Dowd9696,
	author = {Dowd, Jennifer Beam and Andriano, Liliana and Brazel, David M. and Rotondi, Valentina and Block, Per and Ding, Xuejie and Liu, Yan and Mills, Melinda C.},
	title = {Demographic science aids in understanding the spread and fatality rates of COVID-19},
	volume = {117},
	number = {18},
	pages = {9696--9698},
	year = {2020},
	doi = {10.1073/pnas.2004911117},
	publisher = {National Academy of Sciences},
	issn = {0027-8424},
	URL = {https://www.pnas.org/content/117/18/9696},
	eprint = {https://www.pnas.org/content/117/18/9696.full.pdf},
	journal = {Proceedings of the National Academy of Sciences}
}

@article {AgegroupGermany,
	author = { Yaylali, Ayse},
	title = {Factors Affecting the Number of COVID-19 Cases and the Death Rate: Empirical Evidence from the German States },
	year = {2020},
	doi = {10.2139/ssrn.3617986},
	URL = {http://dx.doi.org/10.2139/ssrn.3617986 },
}

@misc{coviddata,
    author       = {Torben},
    title        = {COVID-19-Coronavirus-German-Regions },
    year         = {2020},
    note        = {data retrieved from github, 
                    \url{https://github.com/entorb/COVID-19-Coronavirus-German-Regions}}
    }

@misc{area,
    author       = {Statistisches bundesamt},
    title        = {Gebietsfläche: Kreise, Stichtag},
    year         = {2016},
    note        = {data retrieved from statistisches bundesamt - Genesis online - 11111-0002, 
                    \url{https://www-genesis.destatis.de/genesis/online}}
    }

@misc{agestructure,
    author       = {Statistisches bundesamt},
    title        = {Bevölkerung: Kreise, Stichtag, Altersgruppen},
    year         = {2019},
    note        = {data retrieved from statistisches bundesamt - Genesis online - 12411-0017, 
                    \url{https://www-genesis.destatis.de/genesis/online}}
    }

@misc{income,
    author       = {Eric, Seils and Helge, Baumann},
    title        = {BVERFÜGBARE HAUSHALTSEINKOMMEN IM REGIONALEN VERGLEICH},
    year         = {2018},
    url         =   {https://www.boeckler.de/pdf/wsi_vm_verfuegbare_einkommen.pdf},
    journal = {WSI verteilungsmonitor}
    }

@misc{timeseries,
    author       = {RKI (Robert Koch-Institut)},
    title        = {RKI COVID19},
    year         = {2020},
    url         =   {https://npgeo-corona-npgeo-de.hub.arcgis.com/datasets/dd4580c810204019a7b8eb3e0b329dd6_0?page=4&selectedAttribute=Altersgruppe},
    journal = {NPGEO Corona}
    }


\end{document}